\documentclass[letterpaper, 10 pt, conference]{ieeeconf}  
\IEEEoverridecommandlockouts                              

\usepackage{bm}
\usepackage{url}
\usepackage{graphicx} 
\usepackage{mathptmx} 
\usepackage[mathscr]{euscript}  
\usepackage{amsmath} 
\usepackage{amssymb}  
\usepackage{balance}
\usepackage{multirow}
\usepackage{subfigure}
\usepackage{siunitx}
\usepackage{xcolor}
\usepackage{adjustbox}
\usepackage{array}
\usepackage{booktabs}
\usepackage{multirow}
\usepackage{xcolor}
\usepackage{siunitx}
\usepackage{diagbox}
\usepackage{boldline}
\usepackage{pifont}
\usepackage{todonotes}
\usepackage{yfonts}
\usepackage[percent]{overpic}
\usepackage{tcolorbox}
\usepackage{gensymb}

\newcommand{\cmark}{\ding{51}}%
\newcommand{\xmark}{\ding{55}}%
\newcommand{\etal}{\emph{et~al.}}

\definecolor{sky}{RGB}{70, 130, 180}
\definecolor{building}{RGB}{70, 70, 70}
\definecolor{road}{RGB}{128, 64, 128}
\definecolor{sidewalk}{RGB}{244, 35, 232}
\definecolor{cyclist}{RGB}{190, 153, 153}
\definecolor{vegetation}{RGB}{107, 142, 35}
\definecolor{pole}{RGB}{153, 153, 153}
\definecolor{sign}{RGB}{220, 220, 0}
\definecolor{pedestrian}{RGB}{220, 20, 60}
\definecolor{mcar}{RGB}{0, 255, 0}
\definecolor{scar}{RGB}{0, 0, 255}

\newcolumntype{P}[1]{>{\centering\arraybackslash}p{#1}}

\renewcommand{\baselinestretch}{0.985}

\newcolumntype{R}[2]{%
    >{\adjustbox{angle=#1,lap=\width-(#2)}\bgroup}%
    l%
    <{\egroup}%
}

\title{\LARGE \bf
USegScene: Unsupervised Learning of Depth, Optical Flow and Ego-Motion with Semantic Guidance and Coupled Networks
}

\author{Johan Vertens \and Wolfram Burgard
\thanks{Johan Vertens is with the University of Freiburg, Germany. Wolfram
Burgard is with the University of Technology, Nuremberg, Germany. Corresponding author: {\tt\small vertensj@informatik.uni-freiburg.de}}
}

\begin{document} 

\maketitle
\thispagestyle{empty}
\pagestyle{empty}

\begin{abstract}
  In this paper we propose USegScene, a framework for semantically guided unsupervised learning of depth, optical flow and ego-motion estimation for stereo camera images using convolutional neural networks. Our framework leverages semantic information for improved regularization of depth and optical flow maps, multimodal fusion and occlusion filling considering dynamic rigid object motions as independent SE(3) transformations. Furthermore, complementary to pure photometric matching, we propose matching of semantic features, pixel-wise classes and object instance borders between the consecutive images. In contrast to previous methods, we propose a network architecture that jointly predicts all outputs using shared encoders and allows passing information across the task-domains, e.g the prediction of optical flow can benefit from the prediction of the depth. Furthermore, we explicitly learn the depth and optical flow occlusion maps inside the network, which are leveraged in order to improve the predictions in the respective regions. 
  We present results on the popular KITTI dataset and show that our approach outperforms previous methods by a large margin.
\end{abstract}


\section{Introduction}
\label{sec:introduction}

The estimation of optical flow, depth and ego-motion from stereo-images are fundamental computer vision problems with an enormous amount of robotic applications.
In the field of autonomous driving it is used for solving problems like mapping \cite{mur2015orb}, object tracking \cite{dang2002fusing, choi2015near}, or localization \cite{cvivsic2015stereo}, enabling the systems to reason about their surrounded environment.

In previous years researchers mainly approached the tasks with supervised learning, leveraging datasets like KITTI \cite{Geiger2013IJRR} or synthetic datasets such as Synthia \cite{ros2016synthia} or FlyingChairs \cite{mayer2016}. However, real world datasets with a high amount of ground-truth data are still rare, since expensive laserscanners are required and complicated procedures need to be carried out to gather such data. Furthermore, networks trained for depth, optical flow, and visual odometry on these datasets tend to overfit to the datasets specific camera-setup and thus have limited flexibility. 

Recent works propose to tackle the individual problems with unsupervised methods using photometric consistency to trigger the learning \cite{jason2016back, meister2018unflow, godard2017unsupervised, li2018undeepvo}. Due to the pure usage of photometric cues these approaches have problems with largely untextured regions or significant lighting changes between the consecutive frames.

\begin{figure}[h]
\scriptsize 
\centering 
\setlength{\tabcolsep}{0.3em}
\renewcommand{\arraystretch}{1}
\begin{tabular}{P{4cm} P{4cm}}
\multicolumn{2}{c}{\includegraphics[width=8.025cm]{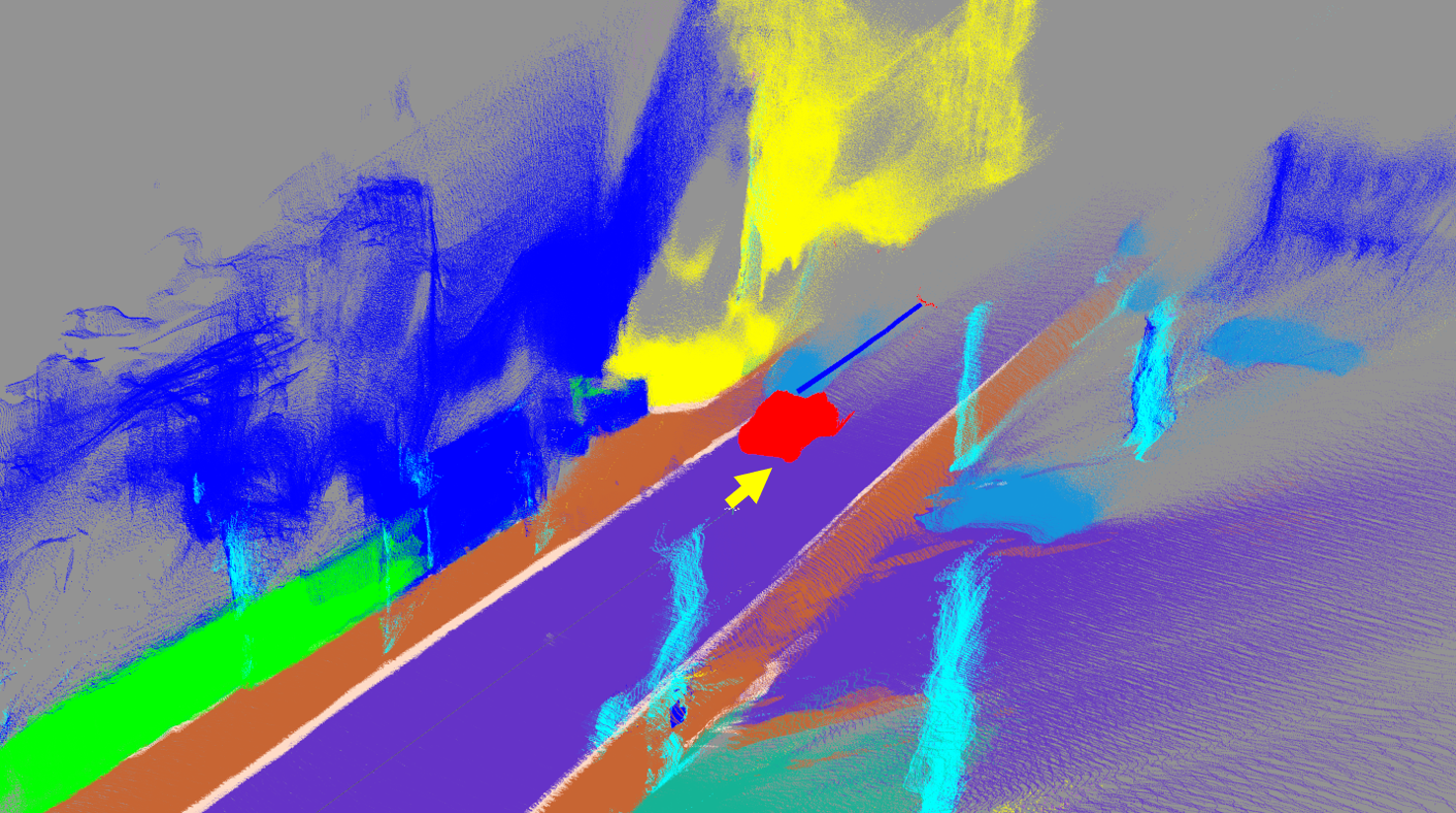}}\\
\includegraphics[width=.97\linewidth]{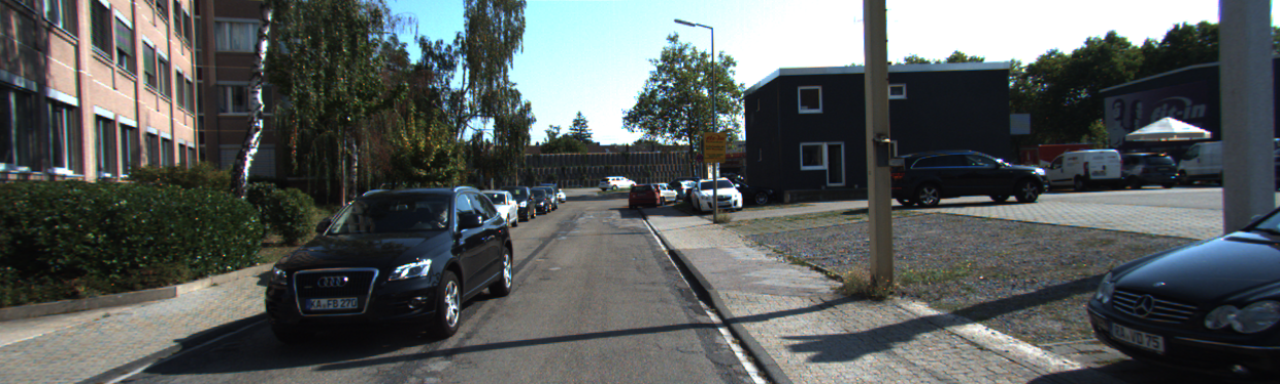} 
& \includegraphics[width=.97\linewidth]{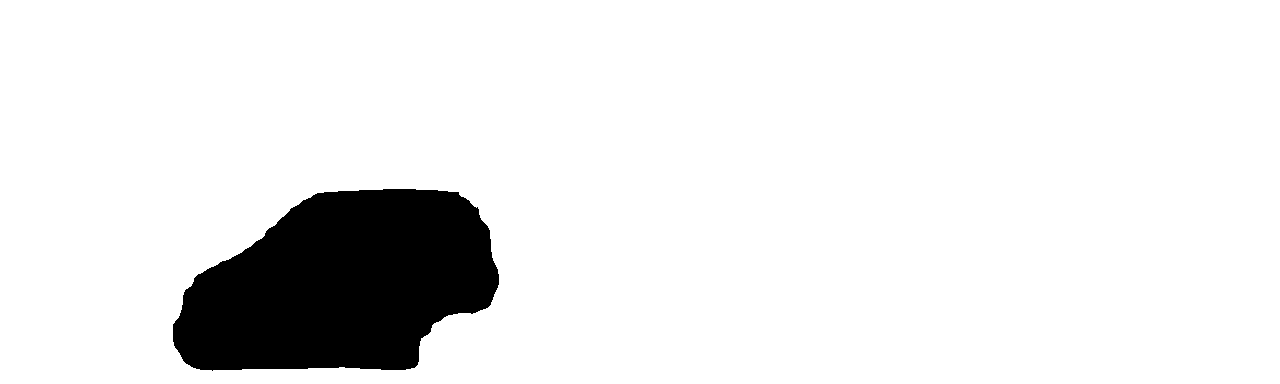} \\
 \includegraphics[width=.97\linewidth]{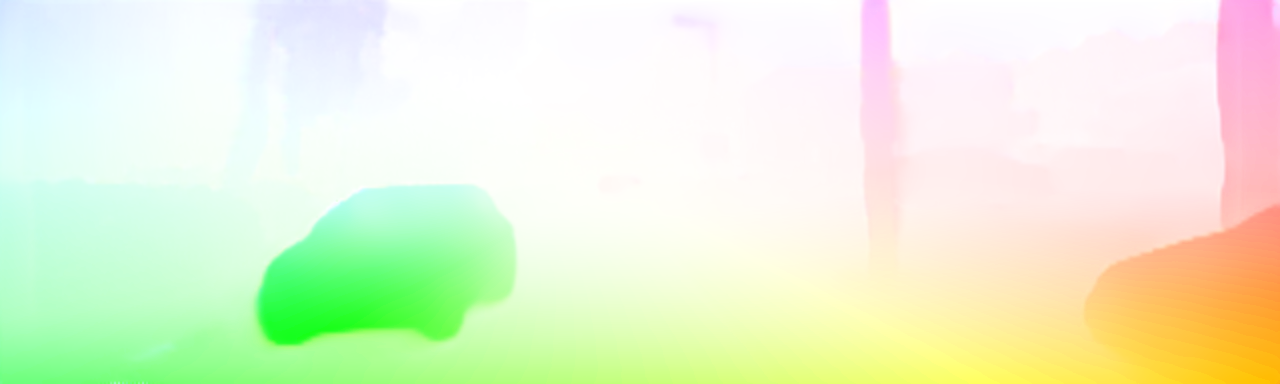} &
\includegraphics[width=.97\linewidth]{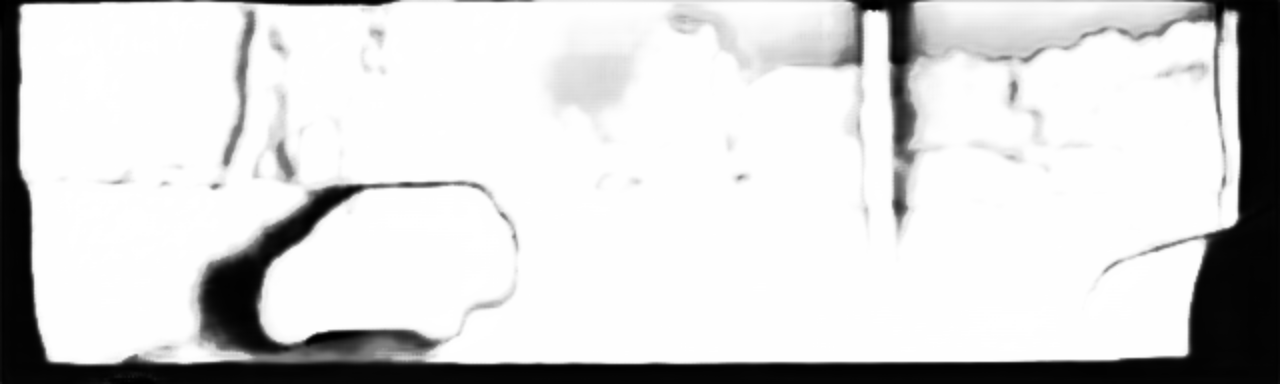} \\
\includegraphics[width=.97\linewidth]{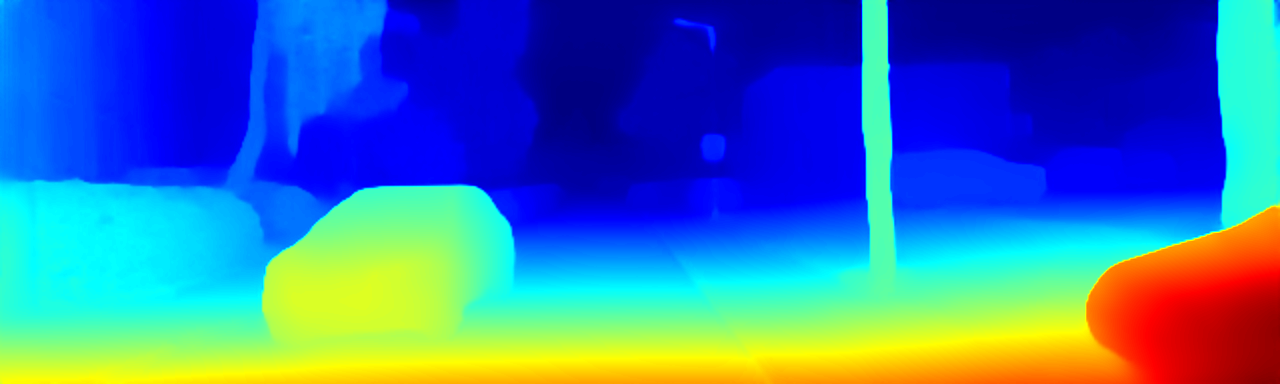} & \includegraphics[width=.97\linewidth]{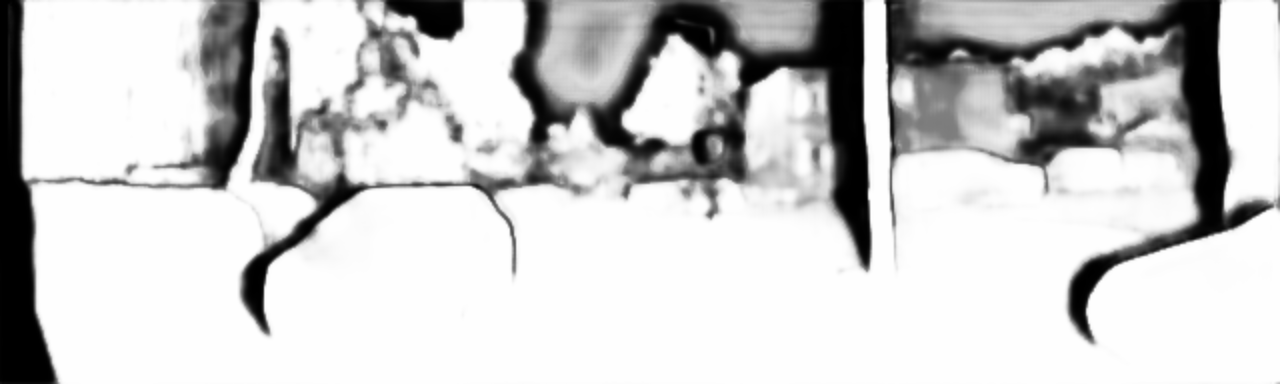}
\end{tabular} 
\caption{Example output of our method. From top to bottom: Aggregated (20 frames) semantically segmented point-cloud, color image of left camera (left), motion segmentation map (right), optical flow (left), predicted occlusion between consecutive images (right), predicted disparity (left), predicted occlusion between left-right image pair (right). The yellow arrow illustrates the camera-pose related to all predictions.}
\label{fig:covergirl}
\vspace{-0.3cm}
\end{figure}

Furthermore, most of the previous works struggle with occlusions between the consecutive images and use simple edge-based smoothing techniques that do not involve context information.

Other works \cite{sfmlearner, yin2018geonet, wang2019unos, epc++, zou2018df} couple the tasks of estimating depth, optical flow and ego-motion by exploiting problem-dependencies within the loss function. This way the tasks are not longer learned independently but are constrained by each other in the learning process. However, these proposed approaches do not leverage context information, such as semantic segmentation that we show can help the network to improve the predictions. 
Additionally, these approaches tackle the individual tasks with independent and largely decoupled networks ,making the complete system less efficient, and do not handle occlusions that occur due to dynamic objects.

In this paper, we propose an unsupervised learning framework for semantic aware estimation of depth, optical flow and ego-motion using stereo cameras. We jointly infer all predictions using two consecutive stereo-pairs and their corresponding semantic features in a coupled network with shared encoders. The unified nature of the network enables leveraging complementary information across the output-domains, e.g the ego-motion estimation can profit from the estimation of depth and optical flow.

Complementary to direct photometric losses we propose semantic feature matching that matches feature maps from a semantic segmentation network, improving predictions on textureless regions.
This pre-trained segmentation network is trained to predict the class-wise semantic segmentation and additionally regresses the borders of distinct rigid instances of objects in the scene.

For flow and disparity regularization we propose a novel weighted semantic patch smoothing loss (WSPS) which smooths the optical flow or disparity map only within certain semantic classes, considering the output of the semantic network.

To cope with occlusions we further propose a method that fills these regions with information from estimated SE(3) transformations of the rigid ego-motion and independent moving rigid objects. In contrast to previous works, our network also predicts the occluded regions directly without the need of multiple inference passes and predicts correction for these regions.

We evaluate our method on the popular real world dataset KITTI \cite{Geiger2012CVPR}, achieving state-of-the-art accuracies.

\section{Related Work}
\label{sec:relatedWork}

As our method comprises of the estimation of disparity, ego-motion and optical flow we list related work for each field of research individually. Finally we give an overview on previous works that focus on the combination of these different tasks. We mainly focus on works that leverage convolutional neural networks.

\subsection{Disparity Estimation}
Traditionally, disparity estimation algorithms follow the paradigm of computing costs, cost aggregation, global optimization and depth refinement \cite{scharstein2002taxonomy}. Published works propose to learn parts of this pipeline with neural networks. LeCun \etal \cite{zbontar2016stereo} propose to learn the matching costs by training a siamese-network to differentiate between small similar and dissimilar image patches. This training is done in a supervised fashion where a binary classification dataset yields the basis as training data. Similarly Shaked et. al \cite{shaked2017improved} learn the matching costs with a new highway network architecture that leverages multilevel residual shortcut connections. Additionally, they propose a global disparity network to estimate the disparity confidence.
As more real world \cite{Geiger2012CVPR, scharstein2014high} and synthetic datasets \cite{mayer2016, gaidon2016virtual, ros2016synthia} for depth estimation became available, many approaches have been proposed to tackle the problem in an end-to-end way. DispNet \cite{mayer2016large} learns to predict the disparity directly using a two-stream network with explicit correlation layers.

In the work from Kendall \etal \cite{kendall2017end} a new way to construct the cost volume, where all pixel-wise costs according to a specific disparity value are aggregated and then convolved with 3D convolutions.

Chang \etal \cite{psmnet} present a network architecture that aggregates global context information via a spatial pyramid pooling module for improving the disparity accuracy in regions with difficult appearance or lighting.

Other works \cite{godard2017unsupervised, zhong2017self} learn to predict depth in an unsupervised fashion. They use photometric cues and forward-backward warping in order to train the network without the need of any labeled data.

Additionally, some methods \cite{yang2018segstereo, zhang2019dispsegnet} have been proposed that solve disparity estimation and semantic segmentation within a single network. The authors show that the learned semantic embedding helps to improve the accuracy.

\subsection{Optical Flow Estimation}

In recent years CNNs became increasingly popular in the field of optical flow estimation, superseding previous methods. The authors of FlowNet \cite{dosovitskiy2015flownet} propose to learn pixel-wise optical flow using a convolutional neural network and synthetic data. Further works present cascaded \cite{IMKDB16} and pyramid network \cite{pwcnet} architectures that make the predictions more accurate and the system more efficient.

Hui \etal \cite{hui2019lightweight} propose a fast network and uses flow regularization for further speeding up inference time while maintaining accurate and sharp predictions.

In order to get rid of the dependence of ground-truth data, Yu \etal \cite{jason2016back} propose to learn optical flow in an unsupervised fashion using photometric losses and smoothing terms. The authors of UnFlow \cite{meister2018unflow} further propose a loss that is robust to illumination changes across the consecutive images.

Recent advances \cite{wang2018occlusion, liu2019selflow, liu2019ddflow} in unsupervised optical flow estimation cope with the problem of occlusion. They mask out faulty photometric losses \cite{wang2018occlusion}, or handle occlusions by training a student network from a teacher network that has more information about occluded regions \cite{liu2019selflow}.

\subsection{Odometry Estimation}

Visual odometry is a well studied field of computer vision where many efficient and robust methods exist \cite{mur2015orb, forster2014svo, engel2017direct}. However, researchers \cite{wang2017deepvo,wang2018end} have been investigating learning visual odometry with neural networks in order to overcome problems with texture-less environments or changing lighting conditions. Additionally these methods do not rely on apriori handcrafted feature descriptors but learn them themselves.

In recent years many works \cite{zhou2017unsupervised, li2018undeepvo, iyer2018geometric} learn visual odometry in an unsupervised way. Beside an odometry network, a depth estimator is trained that predicts the scene geometry. The predicted geometry and the ego-motion estimate can then be used to synthesize adjacent images and a photometric loss is applied to train the networks jointly. 

\subsection{Combined Approaches}

Leveraging all advances in the individual fields of disparity, optical flow and ego-motion estimation, many works emerged that combine all tasks in a single unsupervised framework \cite{yin2018geonet, zou2018df, epc++, cao2019learning, wang2019unos}. In GeoNet \cite{yin2018geonet} firstly monocular depth and the ego-motion is learned. A rigid flow field is then computed by shifting the 3D geometry by the predicted ego-motion and projecting the motion vectors on the image plane. Since this flow prediction is only valid for static parts of the scene, they also propose a residual optical flow module that corrects for the dynamic regions. 
In the work of Zou \etal \cite{zou2018df} the rigid optical flow constraints the learning of a pixel-wise optical flow module. The constraint is not applied on dynamic regions, which they detect with forward-backward flow consistency checking. The authors of EPC++ \cite{epc++} additionally add consistency constraints between consecutive depth maps and segment dynamic regions using scene flow vectors. Wang \etal \cite{wang2019unos} present a unified system for learning optical flow, visual odometry and stereo depth estimation. Instead of directly using the ego-motion estimate from a network they refine the ego-motion using the flow and depth cues.

All combined approaches exploit the relations between the individual tasks and show that joint training of all depth, optical flow and visual odometry yields overall significantly better results in each domain. However, previous works do not couple the tasks for dynamic regions while exploiting semantic cues. Additionally, none of the previous works couple all problems in the network structure directly but rather use stand-alone networks for each individual task independently. Furthermore, all previous methods do not handle regions that are occluded and dynamic. We, instead propose a system which improves the prediction quality in these regions using rigid object-instance level SE(3) transforms.

\section{Technical Approach}
\label{sec:technicalApproach}

In the following we describe our approach for unsupervised learning of depth, optical flow and ego-motion. We first present our unified network architectures that solves all tasks within a single network which gets additional semantic features as input. We then present semantic feature matching which enforces semantic consistency across the consecutive frames. Furthermore, we present a loss for semantic aware regularization which guides the network to predict smooth output maps within equal classes, while being sharp at class boundaries. 
Using a similar technique as for the semantic aware regularization we propose to constrain occluded depth regions by neighboring non-occluded regions with the same semantic class.
We also present a method to distinguish between static and dynamic objects and propose novel 3D consistency losses which couples optical flow, disparity and ego-motion directly in the 3D space. Finally we propose a loss that enforces the network to fill occluded regions in optical flow maps using ego-motion and independent motions of rigid dynamic objects.

\subsection{Unified 3D Geometry and Motion Network}

\begin{figure*}[h!]
\centering
\includegraphics[width=\linewidth]{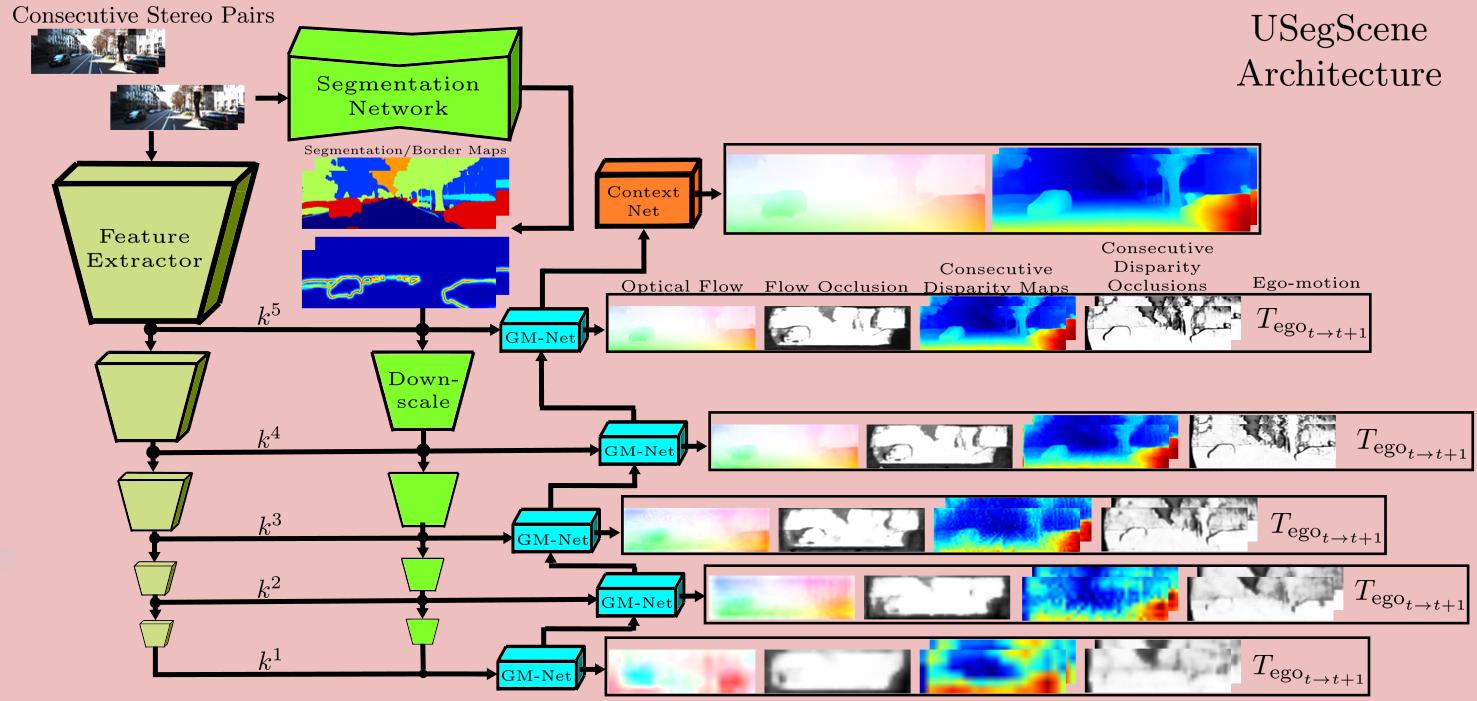}
\caption{Overview of our USegScene architecture.
Using two consecutive stereo image-pairs we predict optical flow $F$, two consecutive disparity maps $D$, ego-motion $T_{\mathrm{ego}}$ and occlusions $O$ in respective estimators at multiple levels $k^1, k^2, k^3, k^4, k^5$ with different resolutions. Each estimator additionally leverages complementary information from a semantic segmentation network, which predicts pixel-wise class labels $S$ and a regression of instance-borders $B$.}
\label{fig:scheme}
\vspace{-0.3cm}
\end{figure*}

While previous methods predict optical flow, disparity and ego-motion with independent networks we present a neural network architecture that predict these outputs jointly. In general, our architecture can be divided into seven different modules namely a semantic segmentation network, a shared feature encoder, a disparity estimator, an optical flow estimator, an occlusion refinement module, an ego-motion estimator and a context-module.

Our network takes advantage from the fact that the predictions of optical flow, disparity and ego-motion, leverage similar feature representations that are well suited for dense matching. Thus, our structure does not predict this redundant information for optical flow, disparity and ego-motion individually but computes them once and feeds it to the modules in a shared manner. In contrast to other works, the output of the disparity module gets passed to the optical flow and the ego-motion module, while the ego-motion module also benefits from the output of the optical flow module.

To cope with texture-less regions and occlusions, we additionally feed semantic cues from a semantic segmentation network into each of the sub-modules, making our modules context aware. The high-level semantic information is then used in the matching process, making the predictions more robust.

Our network is structured as a multi-scale pyramid neural network similar to PWCNet \cite{pwcnet}, following a course-to-fine scheme. In total we use 5 different levels $k_{1}...k_{5} \in K$ where in each level a optical flow map, a disparity map and the ego-motion gets estimated. After the last level $k_{5}$ we further pass the disparity and optical flow predictions through a context-network.

\subsubsection{Feature Encoder}
We adapt the feature encoder from PWCNet \cite{pwcnet} and add a SPP module \cite{psmnet} to it, which showed significant improvements in disparity estimation. Our feature encoder extracts features $f_{l_{t}}$, $f_{l_{t+1}}$, $f_{r_{t}}$ and $f_{r_{t+1}}$ corresponding to the consecutive input stereo image pairs $(I_{l_{t}}$, $I_{r_{t}})$, $(I_{l_{t+1}}$, $I_{r_{t+1}})$. The feature maps are predicted for the different scales $\frac{1}{2}$, $\frac{1}{4}$, $\frac{1}{8}$, $\frac{1}{16}$ and $\frac{1}{32}$ and then fed into the 5 levels of the depth, optical flow and ego-motion estimators.

\subsubsection{Semantic Segmentation Network}

\begin{figure}
\scriptsize 
\centering 
\setlength{\tabcolsep}{0.3em}
\renewcommand{\arraystretch}{1}
\begin{tabular}{P{6cm}}
\includegraphics[width=.97\linewidth]{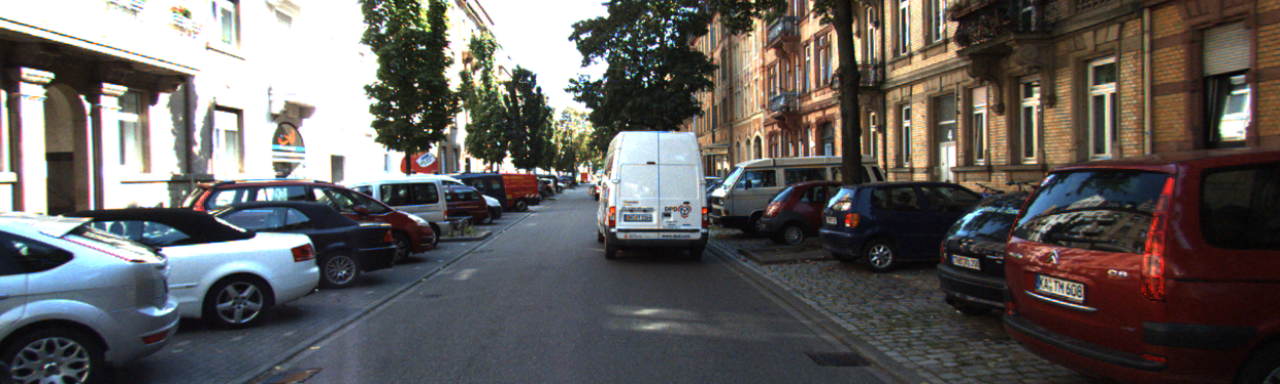} \\
\includegraphics[width=.97\linewidth]{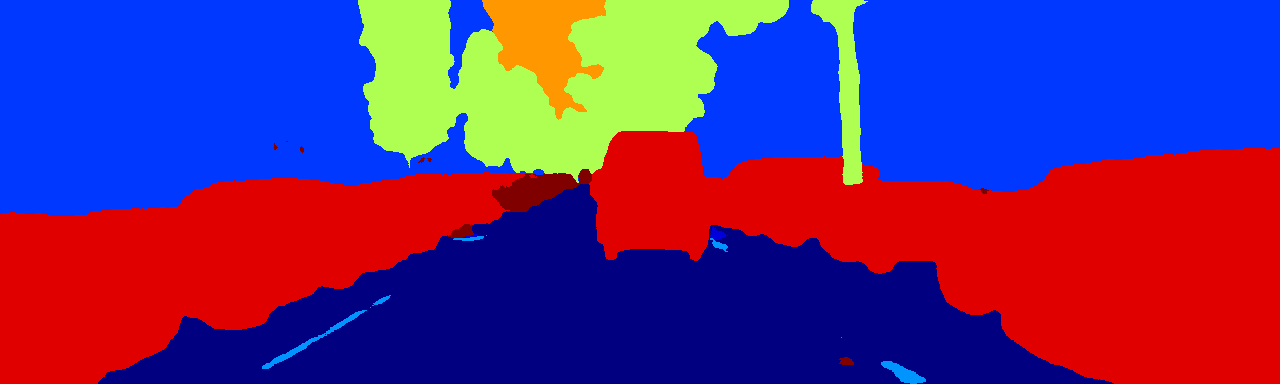} \\
\includegraphics[width=.97\linewidth]{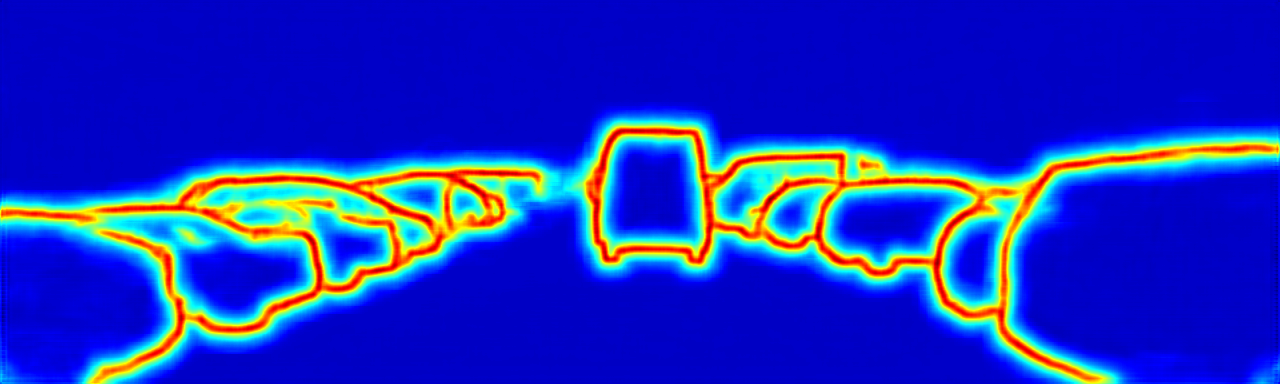} \\
\includegraphics[width=.97\linewidth]{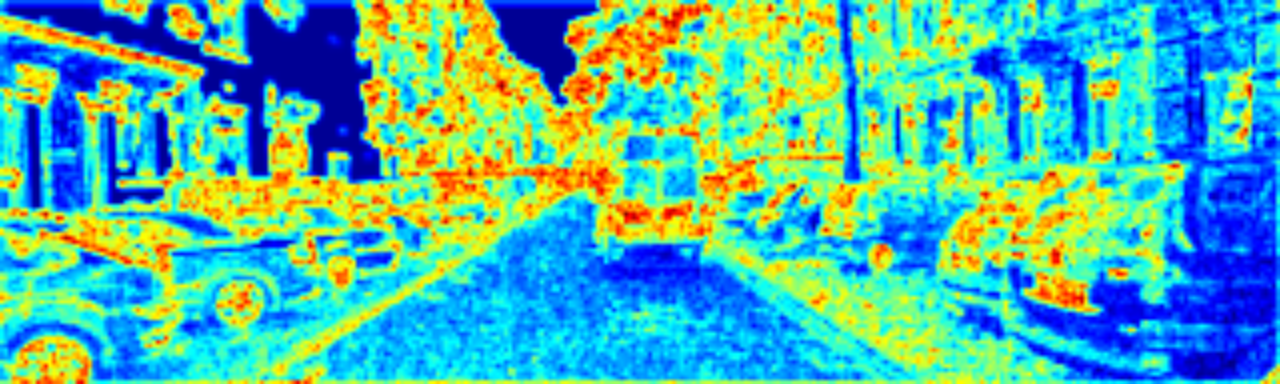}
\end{tabular} 
\caption{Output prediction examples of our semantic segmentation network. From top to bottom: Color image of left camera, pixel-wise semantic segmentation, regression of instance-wise object borders, features from the second block of the ResNext architecture of the segmentation network. For better visualization we sum the features along the channel-dimension and apply a jet-colorcoding.}
\label{fig:semanticExamples}
\vspace{-0.3cm}
\end{figure}

We use a ResNext50 \cite{xie2017aggregated} architecture with ASPP \cite{chen2017rethinking} and skip connections in order to predict pixel-wise semantic segmentation maps. We choose the classes in a way that they represent classes with distinct geometric complexity and rigidity characteristics. Our classes are: \textit{Road, Sidewalk, Building, Construction, Fence, Poles, Vegetation, Various Terrain, Sky, Pedestrian/Rider, Vehicle} and \textit{Background}. Beside the segmentation map $S$ our semantic network also predicts a regression of borders of rigid object instances $B=e^{-\frac{d}{\sigma}}$ where $d$ is the distance to the instance borders in pixels and $\sigma$ a shaping constant for the exponential, which we set to $15$ during all experiments. Fig. \ref{fig:semanticExamples} shows an example output of our segmentation network. We feed both, the segmentation maps $S_{l_{t}}$, $S_{l_{t+1}}$ and the border regression maps $B_{l_{t}}$, $B_{l_{t+1}}$, corresponding to the consecutive left images $I_{l_{t}}$, $I_{l_{t+1}}$, into every level of our disparity, optical flow and ego-motion modules. In order to be compatible to the corresponding scales for each level we simply re-size the segmentation and border maps bi-linearly to match with the input resolution of the respective modules.

\subsubsection{Depth Estimator}

\begin{figure*}[h]
\centering
\includegraphics[width=\linewidth]{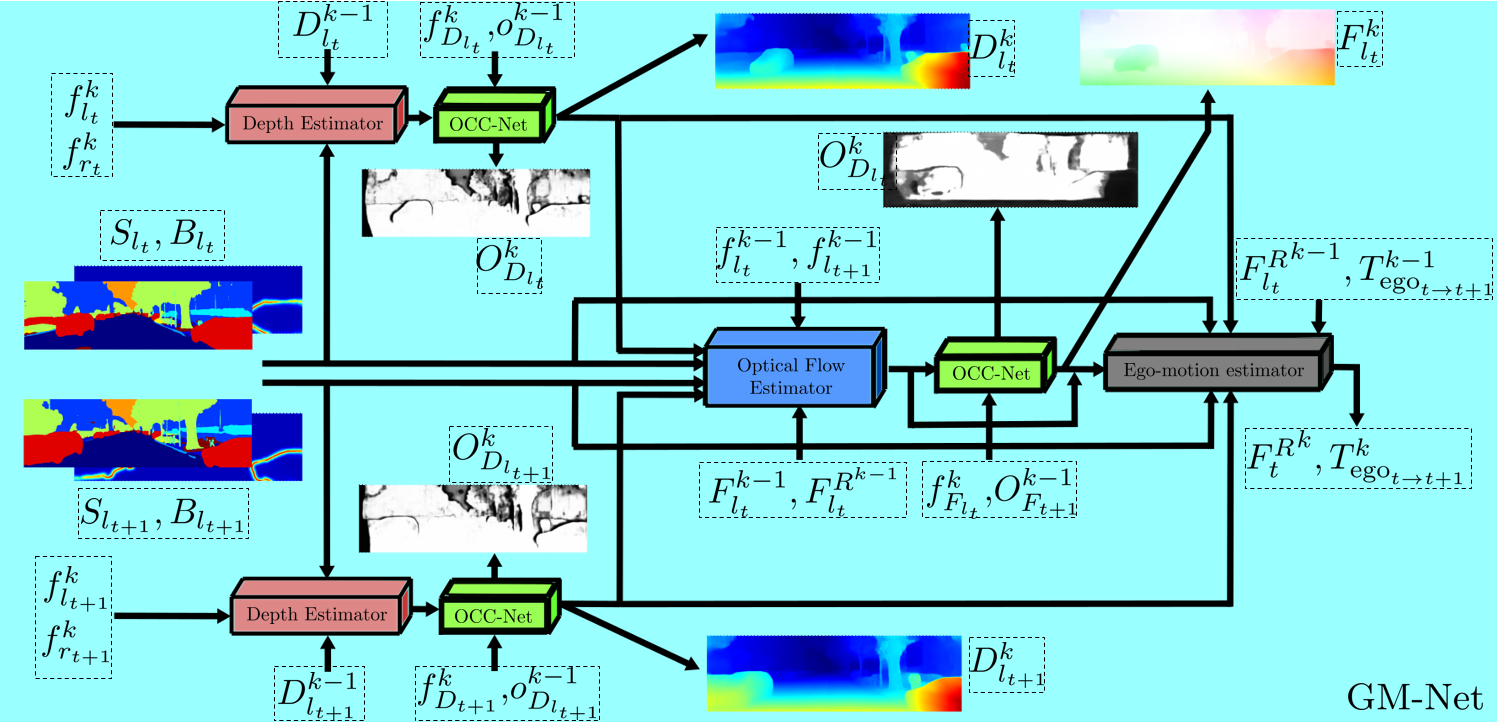}
\caption{This figure illustrates the module GM-Net, which comprises of the depth estimator, optical flow estimator and ego-motion estimator.  The GM-Net is applied to every level within the USegScene architecture predicting the disparities $D_{l_{t}}^k$, $D_{l_{t+1}}^k$, optical flow $F_{l_{t}}^k$ and ego-motion $T_{\mathrm{ego}}^k$ for the current level, given the previous predictions $D_{l_{t}}^{k-1}$, $D_{l_{t+1}}^{k-1}$, $F_{l_{t}}^{k-1}$, $T_{\mathrm{ego}}^{k-1}$ and the features from the shared encoder $f_{l_{t}}^k$, $f_{l_{t+1}}^k$, $f_{r_{t}}^k$, $f_{r_{t+1}}^k$. In the GM-Net all sub-modules leverage the predicted consecutive semantic class and instance-border information $S$, $B$.
Additionally we apply our OCC-Net to each disparity and optical flow estimator, which estimates the respective occlusion maps $O_{D_{l_{t}}}$,$O_{D_{l_{t+1}}}$,$O_{F_{l_{t}}}$ and applies corrections in those regions.
In our architecture the optical flow estimator can benefit from the consecutive disparity maps and the rigid flow.
Complementary to the features from the shared encoder the ego-motion estimator also takes the consecutive disparity maps $D_{l_{t}}^k$, $D_{l_{t+1}}^k$ and the predicted optical flow $F_{l_{t}}$ as input.}
\label{fig:gmnetscheme}
\vspace{-0.3cm}
\end{figure*}

In order to predict the 3D geometry as two consecutive disparity maps $D_{l_{t}}^k$ and $D_{l_{t+1}}^k$, we adapt the estimator of PWCNet \cite{pwcnet} with dense blocks that was originally designed to predict optical flow. In contrast to previous works such as \cite{psmnet}, which builds a memory expensive cost volume and use 3D convolutions, this estimator only uses 2D convolutions and fast correlation modules, resulting in a faster network with smaller memory footprint. In addition to the features of the left and right image $f_{l_{t}}^k$ and $f_{r_{t}}^k$ from the shared feature encoder of the respective level and the previous disparity estimate $D_{l_{t}}^{k-1}$ from the previous level, we concatenate the re-sized segmentation maps $S_{l_{t}}^k$, $S_{r_{t}}^k$ and border maps $B_{l_{t}}^k$, $B_{r_{t}}^k$ to the features after the correlation layer, thus making our disparity estimator semantic aware.

\subsubsection{Optical Flow Estimator}
Similar to the depth estimator our optical flow estimator follows the estimator architecture of PWCNet \cite{pwcnet}, but further leverages complementary geometric and semantic cues from the depth estimators, the semantic segmentation network and the ego-motion estimators. More precisely, our optical flow estimator at level $k \in K$ takes the feature maps $f_{l_{t}}^k$ and $f_{l_{t+1}}^k$ from the shared feature encoder, the disparity maps $D_{l_{t}}^k$, $D_{l_{t+1}}^k$, the semantic and border maps $S_{l_{t}}^k$, $S_{l_{t+1}}^k$, $B_{l_{t}}^k$, $B_{l_{t+1}}^k$ and the optical flow estimate from the previous level $F_{l_{t}}^{k-1}$ in order to predict the next optical flow estimate $F_{l_{t}}^{k}$. 
Furthermore, using the disparity $D_{l_{t}}^k$ of the current level and the ego-motion estimate $T_{\mathrm{ego}_{t \rightarrow t+1}}^{k-1}$ of the previous level, the rigid flow ${F_{l_{t}}^R}^{k}$, which refers to an optical flow map that is valid only for the static regions, is calculated and additionally concatenated to the inputs of the optical flow estimator. For more information on how to render the rigid flow please refer to chapter \ref{chap:unsupervisedTraining}.
Thus, our multi-modal optical flow branch is able to learn to match consecutive features of the consecutive frames but also 3D geometry and semantic information. Since we pass the rigid flow to the estimator as prior information about the optical flow in static regions, the full prediction of the optical flow is eased, which we show in our experiments in section \ref{sec::opticalFLowEval}.

\subsubsection{Occlusion Refinement Network}

\begin{figure}[h]
\centering
\includegraphics[width=\linewidth]{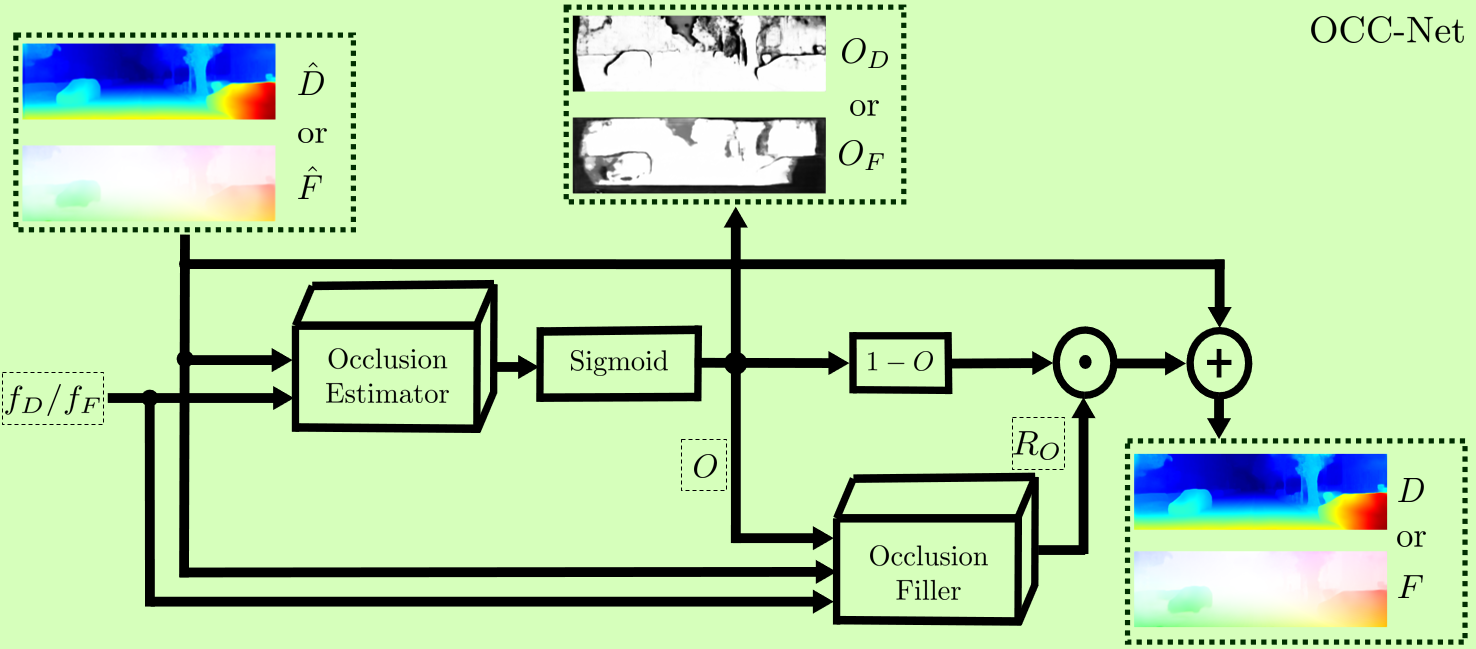}
\caption{This figure illustrates the architecture of our OCC-Net module. Given an initial disparity or optical flow map and high-level features from the respective estimator,  we first estimate the occluded regions. Following, we predict corrections that are gated by the occlusion and added to the initial disparity/optical flow map.}
\label{fig:occscheme}
\vspace{-0.3cm}
\end{figure}
Since pixel-wise disparity and optical flow estimation is dependent on consistently observable image features, handling the occluded pixels in the consecutive stereo image pairs plays an important role. We therefore propose a novel module in our network that is capable to learn to detect and correct occluded regions in the disparity and optical flow maps. First, features $f_F^k$ or $f_D^k$ and the corresponding initial optical flow $\bar{D}^{k}$ or disparity map $\bar{F}^{k}$ are passed into a 4-layer convolutional occlusion detection module that predicts the likelihood of a pixel being occluded or not. Here, $f_F^k$ and $f_D^k$ are the features taken from before the last layer of the respective optical flow or disparity estimator modules at level $k$. We use a sigmoid function at the end of the occlusion detection module in order to have a output range between 0 and 1, where 0 means that a pixel is occluded and 1 means that the pixel is not occluded. Second, an occlusion correction network takes $f_F^k$ or $f_D^k$, the estimated occlusion map $O_F^k$ or $O_D^k$, and the optical flow/disparity map $\bar{D}^{k}$ or $\bar{F}^{k}$ in order to predict corrections $R_O$ for the occluded parts of the input maps. Finally, the corrections are gated by the estimated occlusion map and added to the input optical flow/disparity map resulting in the final maps  $F^k$ or $D^k$. Fig. \ref{fig:occscheme} shows the schematic of the module, which we refer to as OCC-Net. We apply OCC-Net after each disparity and optical flow estimator.

\subsubsection{Ego-Motion Estimator}

\begin{figure}[h]
\centering
\includegraphics[width=\linewidth]{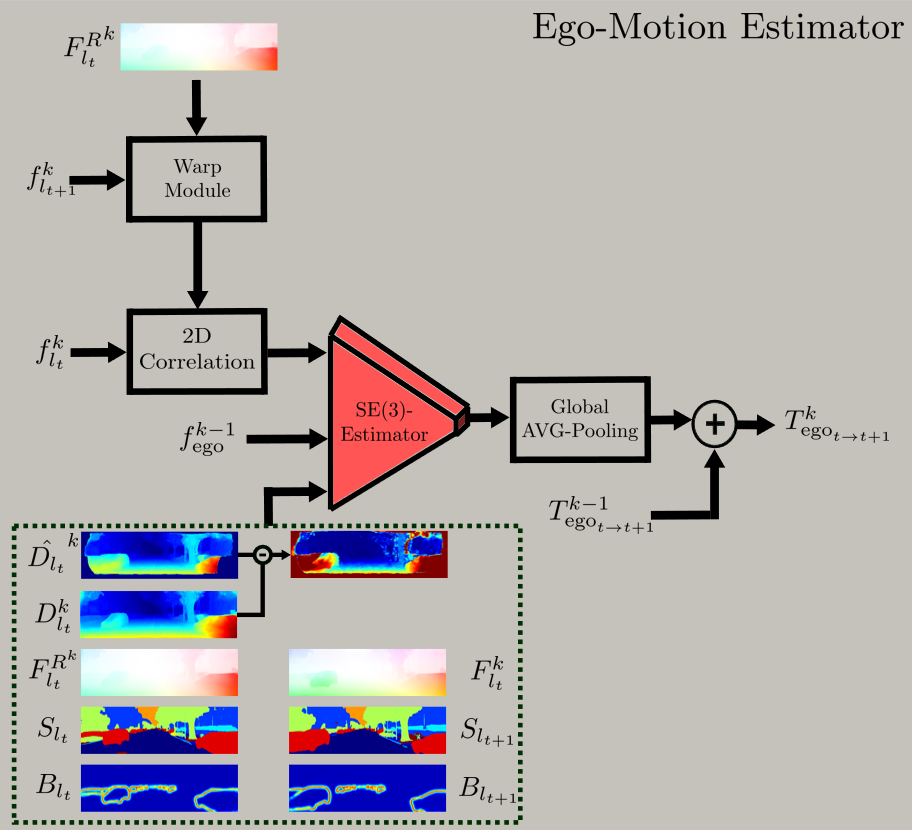}
\caption{This figure illustrates our ego-motion estimator. First, we compile the rigid flow using the disparity and the previous ego-motion estimate. 
We then use the predicted rigid flow to warp features related to the next left image to the view of the previous left image. Following, the warped features are correlated with the features of the previous left image. The correlation result is fed together with the disparities, disparity change, optical flow, rigid optical flow, semantic information and learned high-level features to the SE(3) estimator.
The final ego-motion is obtained by applying global average pooling and adding the ego-motion of the previous level.
 }
\label{fig:odoscheme}
\vspace{-0.3cm}
\end{figure}
In previous works \cite{sfmlearner, yin2018geonet, li2018undeepvo} for unsupervised learning of ego-motion the problem is approached with dense photometric matching while the respective networks do not use 3D information in their inputs.
We propose a CNN architecture that additionally leverages 3D geometry given by the consecutive disparity maps $D_{l_{t}}^k$ and $D_{l_{t+1}}^k$. In each level $k$ we use $D_{l_{t}}^{k}$ and the ego-motion of the previous level $T_{\mathrm{ego}_{t \rightarrow t+1}}^{k-1}$ to warp $f_{l_{t+1}}^{k}$ to the reference of the first frame $I_{l_{t}}$. We then correlate the warped features $\hat{f_{l_{t}}^k}$ with $f_{l_{t}}^k$ and feed the correlation result together with the semantic maps $S_{l_{t}}^k$, $S_{l_{t+1}}^k$, border maps $B_{l_{t}}^k$, $B_{l_{t+1}}^k$, the optical flow $F_{l_{t}}^k$ and the consecutive disparity maps $D_{l_{t}}^{k}, D_{l_{t+1}}^{k}$ in a convolutional network block.
Considering a ego-motion estimator at level $k$ our estimator consists of $5+(k-1)$ layers, where each additional layer has a stride of two, so that its output of every level has the same dimension as the lowest level $k_1$.Similar to previous works \cite{sfmlearner} we apply global average pooling at the output of each ego-motion estimator that predicts the rigid body transformation $T_{\mathrm{ego}_{t \rightarrow t+1}} = \begin{pmatrix}R & t\\ 0 & 1\end{pmatrix}$ where $R \in \mathit{SO}(3)$ and $t \in \mathbb{R}^3$. We predict our transformation in lie-algebraic exponential coordinates $\xi = (v^T \: \omega^T) \in \mathfrak{se}(3)$ and use the exponential map with small-angle approximations \cite{eade2013lie} to map from $\mathfrak{se}(3)$ to $\mathit{SE}(3)$. In every level we add the previous lie-algebraic exponential coordinates from the previous level to the new result. Thus, higher levels only predict residuals that further refine the ego-motion prediction. We illustrate the ego-motion estimator architecture on Fig. \ref{fig:odoscheme}.  \\

The complete network architecture is illustrated in Fig. \ref{fig:scheme}, where GM-Net denotes the network branches for optical flow, disparity and odometry in each level. For brevity, we do not present all architecture parameters, but depict the general scheme of our network.
Following PWCNet \cite{pwcnet}, as there are no previous predictions for optical flow, disparity or ego-motion in the lowest level $k_1$ we directly feed the feature representations into the network without having a correlation before.

\subsection{Unsupervised Training}
\label{chap:unsupervisedTraining}

For training our network architecture we formulate the optical flow, disparity and ego-motion estimation as a photometric optimization problem similar to previous works \cite{wang2019unos, epc++, yin2018geonet}.
Given the optical flow $F_\mathscr{T}$ or disparity $D_\mathscr{T}$ corresponding to a target image $I_\mathscr{T}$, we can warp a source Image $I_\mathscr{S}$ to $I_\mathscr{T}$ by using spatial transformer networks \cite{spatialtransformer}. 

For synthesizing $I_\mathscr{T}$ from $I_\mathscr{S}$ given the ego-motion $T_{ego_{\mathscr{T} \rightarrow \mathscr{S}}}$, we use the depth $Z_\mathscr{T}=\frac{fb}{D_\mathscr{T}}$, where $f$ is the camera focal-length and $b$ is the baseline of the stereo rig, and back-project the pixel locations of $I_\mathscr{T}$ to 3D. In order to get the pixel correspondences in $I_\mathscr{S}$ we then transform the points by $T_{ego_{\mathscr{T} \rightarrow \mathscr{S}}}$ and project them back on the image plane using the camera intrinsic matrix $K$ , resulting in the corresponding pixel coordinates of $I_\mathscr{T}$ in $I_\mathscr{S}$.

The pixel correspondences for the synthesizing operations can be found with:

\begin{equation}
\begin{aligned}
  p_\mathscr{S}^{F_\mathscr{T}} = p_\mathscr{T} + F_{\mathscr{T}}(p_t) \\
  p_\mathscr{S}^{D_\mathscr{T}} = p_\mathscr{T} - D_\mathscr{T}(p_\mathscr{T}) \\
  p_\mathscr{S}^{F_{\mathscr{T}}^R} = \phi(K \: T_{ego_{\mathscr{T} \rightarrow \mathscr{S}}} \gamma (p_\mathscr{T} \mid K,D_{\mathscr{T}})) \\
\end{aligned}
\end{equation}

where $p_\mathscr{S}^{F_\mathscr{T}}$ denote the pixel-locations in $I_\mathscr{S}$ corresponding to $p_\mathscr{T}$ mapped by the optical flow $F_\mathscr{T}$, $p_\mathscr{S}^{D_\mathscr{T}}$ denote the pixel-locations in $I_\mathscr{S}$ corresponding to $p_\mathscr{T}$ mapped by the disparity $D_\mathscr{T}$, and $p_\mathscr{S}^{F_{\mathscr{T}}^R}$ denote the pixel-locations in $I_\mathscr{S}$ corresponding to $p_\mathscr{T}$ mapped by the disparity $D$ and the ego-motion $T_{ego_{\mathscr{T} \rightarrow \mathscr{S}}}$. Here, $\gamma (p_\mathscr{T} \mid K,D_{\mathscr{T}}) = D_{\mathscr{T}}(p_\mathscr{T}) K^{-1} h(p_\mathscr{T})$ means the back-projection of pixel coordinate $p_\mathscr{T}$ to the 3D space, where $h(p_{\mathscr{T}})$ converts $p_\mathscr{T}$ into the homogeneous vector form. The function $\phi(x)$ divides a vector x by it's last element. 

Leveraging the calculated target pixel-locations we denote a synthesized target image as $\hat{I_\mathscr{T}}  = W (I_\mathscr{S} ,M_\mathscr{T})$, where $W$ is the warping function that synthesizes the target bi-linearly using a displacement map $M$, such as the optical flow or disparity map.

We further denote the pixel-displacement vectors between $p_\mathscr{S}^{F_{\mathscr{T}}^R}$ and $p_\mathscr{T}$ as the rigid optical flow map $F_{\mathscr{T}}^R = p_\mathscr{S}^{F_{\mathscr{T}}^R} - p_\mathscr{T}$, which is valid in completely static scenes.

Given the pixel correspondences, we define a loss function that calculates direct photometric differences and the structure similarity \cite{wang2004image} between the synthesized target image $\hat{I_{\mathscr{T}}}$ and the observed target image $I_{\mathscr{T}}$:

\begin{equation}
\begin{aligned}
  \mathcal{L}_{p}^{M_{\mathscr{T}}} = O_{M_{\mathscr{T}}} [\alpha_p \mid I_{\mathscr{T}} - W(I_{\mathscr{S}}, M_{\mathscr{T}}) \mid + \beta_p \mathrm{SSIM}(I_{\mathscr{T}}, W(I_{\mathscr{S}}, M_{\mathscr{T}}))]
\end{aligned}
\end{equation}

Here, $O_{M_{\mathscr{T}}}$ denotes an occlusion mask that indicates which pixels are matchable between the corresponding frames, the weights $\alpha_p$ and $\beta_p$ are empirically set to $0.6$ and $0.4$ respectively. 




\subsubsection{Estimating Occlusions}

We follow \cite{yin2018geonet} to estimate occluded regions $O_{M}$, corresponding to any flow or disparity map $M$, using forward-backward consistency checks on the disparity and optical flow maps. Pixels where forward and backward optical flow/disparity contradict are assumed to be occluded.

Specifically, while training, the optical flow and disparity modules infer the output in both directions, meaning a flow or disparity map gets calculated that maps from a source image $I_{\mathscr{S}}$ to a target image $I_{\mathscr{T}}$ and also from $I_{\mathscr{T}}$ to $I_{\mathscr{S}}$. 

We define the binary occlusion map as:

\begin{equation}
\begin{aligned}
O_{M_{\mathrm{fb}}}=
  (\mid \mid \Delta M  \mid \mid_2 < \max\{\alpha,\beta \mid \mid \Delta M  \mid \mid_2\})
\end{aligned}
\end{equation}
where $\Delta M$ is the difference between the aligned forward and backward inference of $M$: $\Delta M =  M_\mathscr{T} - W(M_\mathscr{S}, M_\mathscr{T})$.

We use the calculated binary occlusion maps in turn for self-supervised training of the occlusion detectors within the OCC-Nets. Since the forward-backward consistency checks are only needed for training, at testing time our network is able to predict occlusions without the need of running the optical flow or disparity estimators in forward and backward mode, saving computation time. While training the occlusion detectors the loss function is simply:

\begin{equation}
\begin{aligned}
  \mathcal{L}_{\mathrm{occ}}^{M} = \mid \mid O_M - O_{M_{\mathrm{fb}}} \mid \mid_{2}
\end{aligned}
\end{equation}

Here, $O_M$ is the predicted occlusion map generated by the respective OCC-Net.
We set $\alpha=0.5$ and $\beta=2 e^{-3}$ for inferring optical flow occlusions and $\alpha=1.0$ and $\beta=4 e^{-2}$ for inferring disparity occlusions respectively.

\subsubsection{Semantic Feature Matching}

Since pure photometric losses do not deliver meaningful information in ill posed conditions, such as low-lighting, textureless regions, transparent areas or reflections, we propose a loss function that matches complementary semantic feature representations from our semantic segmentation network. Specifically, we match source softmax segmentation values  $f_s$, the border regression map $B$, as well as features $f_{\mathrm{seg}}$ from the encoder of the segmentation network, to the target view $\mathscr{T}$. In our case, $f_s$ correspond to features that are inferred from consecutive images $I_{\mathscr{S}}$ and $I_{\mathscr{T}}$ and are matched by $M_{\mathscr{T}}$.
As semantic features $f_{\mathrm{seg}}$ we take the feature maps after the second block of the ResNext-architecture of our semantic segmentation network and apply a sigmoid activation function to squeeze the values to the range between $0$ and $1$.

We formulate our semantic feature matching loss as:

\begin{equation}
\mathcal{L}_{sm}^{M_{\mathscr{T}}} = \sum_{f \in {(f_{s}, B, f_{\mathrm{seg}})}} O_{M_{\mathscr{T}}} (\mid f_{\mathscr{T}} - W(f_{\mathscr{S}}, M_{\mathscr{T}}) \mid)
\end{equation}

An example of each semantic modality $S$, $B$, $f_{\mathrm{seg}}$ is shown in Fig. \ref{fig:semanticExamples}.



\subsubsection{Weighted Semantic Patch Smoothing (WSPS)}

In image regions where neither photometric nor semantic cues exist, we guide the optical flow and disparity estimation by information from neighboring pixels. Many works \cite{godard2017unsupervised, epc++, ranjan2019competitive} use smoothing techniques, which minimize the local second derivative while down-weighting the smoothness effect at positions with high gradient in the corresponding color image.
Instead, we propose a smoothing method that considers larger neighborhoods and additional semantic cues.

For every pixel location $p_m \in P_M$ of a displacement map $M$ we define $\mathcal{N}(p_m)$ as the set of pixel locations within a quadratic neighborhood around the location $p_m$ where the semantic class at location $p_m$ is equal to $p_n \in \mathcal{N}(p_m)$, e.g $S(p_n) = S(p_m)$.
Within this neighborhood, we minimize the difference of the gradient of the flow and disparity map at the center pixel $\nabla M(p_m)$ with all other gradients of other pixels $\nabla M(p_n)$ of the same class.

 We formulate our weighted semantic patch smoothing WSPS loss as follows:

\begin{equation}
\begin{aligned}
    \mathcal{L}_{\mathrm{WSPS}}^{M} =  \sum_{p_m \in P_M} \frac{1}{N} \alpha_c \sum_{p_n \in  \mathcal{N}(p_m)} \omega(p_n, p_m, Z(p_m)) \cdot \\ \rho(  \mid \nabla M(p_n) - \nabla M(p_m) \mid )\\
\end{aligned}
\end{equation}

where: \\

\begin{equation}
    \omega(u_n,u_m,z)=e^{-\frac{\mid \mid p_m - p_n \mid \mid_2 }{\eta \theta(z)}}
\end{equation}

\begin{equation}
    \theta(z) = \frac{1}{1 + \frac{z^2}{bf}}
\label{eq:dispError}
\end{equation}

The function $\omega$ weights the differences inversely proportional to the euclidean distance between the locations $p_n$ and $p_m$ and the depth of the center pixel. Thus, differences of gradients that are far away from each other are less penalized.
Further, if the center pixel has far depth $z$ this effect gets strengthened since the covered real area of the neighborhood becomes bigger with high distances to the camera, due to the projection-characteristics. The regularization is weighted class-dependently by $\alpha_c$, which we empirically set to $0.1$ for classes with high structural variance: \textit{Construction, Fence, Poles, Vegetation, Various Terrain,  Pedestrian/Rider}, $1.0$ for classes : \textit{Building, Various Terrain, Sky} and $8.0$ for occluded pixels or classes with low structural variance or less textural cues: \textit{Vehicle}, \textit{Road}, \textit{Sidewalk}. The parameter $\eta$ defines how much pixels $p_n$ that are far away from $p_m$  are influenced by the loss. We set $\eta$ empirically to $10$ for all experiments.

To avoid oversmoothing between distinct object instances, we enhance our semantic segmentation map with instance information of the types \textit{Vehicle} and \textit{Pedestrian/Rider} using Mask-RCNN \cite{he2017mask}. Note that Mask-RCNN is however not needed at testing time. Consequently, we consider every object instance as a individual class focusing the smoothing inside the instance. Additionally, by using a robust loss function $\rho$ \cite{barron2019general} (with parameters: $\alpha=0.5, c=0.1$) we allow outliers within a class patch up to a certain level, accounting for intra-class breaks.

In comparison to standard local smoothing \cite{godard2017unsupervised}, our loss improves predictions in regions with challenging conditions, such as difficult lighting or overexposure. Pixels that lie in ill-posed regions are not just coupled to their direct neighbors, but over a large patch, regularizing over longer distances. Instead of just down-weighting the loss at pixels with high color gradient, our loss explicitly avoids oversmoothing between different semantic objects. Furthermore, our loss offers the possibility to adjust the smoothing strength over the different classes, e.g allowing less smoothing on classes with high structural variance, such as vegetation.

\begin{figure}
\centering
\includegraphics[width=\linewidth/2]{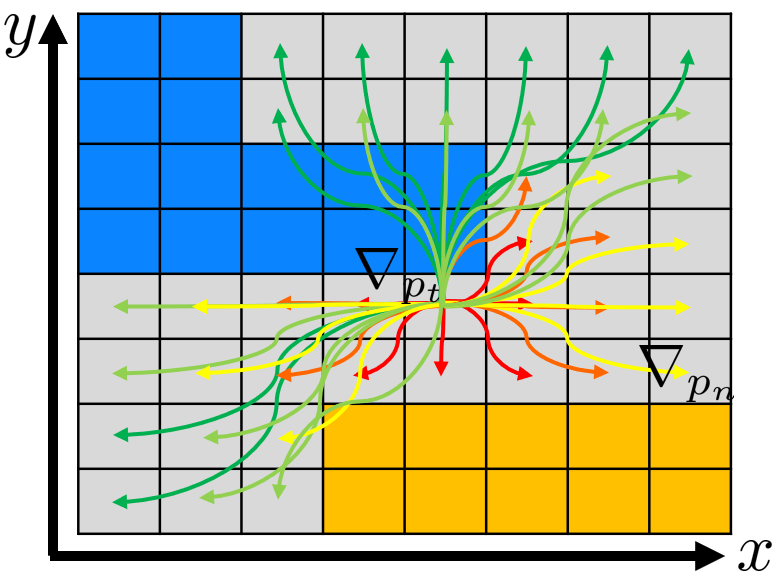}
\caption{
Weighted Semantic Patch Smoothing (WSPS) regularizes either disparity or optical flow maps. We enforce consistency between a disparity/flow gradient $\nabla p_m$ with a fixed position and all other gradients $\nabla p_n$ that lie within a certain neighborhood and have equal semantic class/instance (visualized with distinct colors). Each gradient difference is weighted by a function, which is dependent of pixel distance and depth. Red means high weight, while green means low weight.}
\label{fig:WSPSscheme}
\vspace{-0.3cm}
\end{figure}

We illustrate WSPS on Fig. \ref{fig:WSPSscheme}.

\subsubsection{Semantic Depth Occlusion Filling}

\begin{figure}
\centering
\includegraphics[width=\linewidth/2]{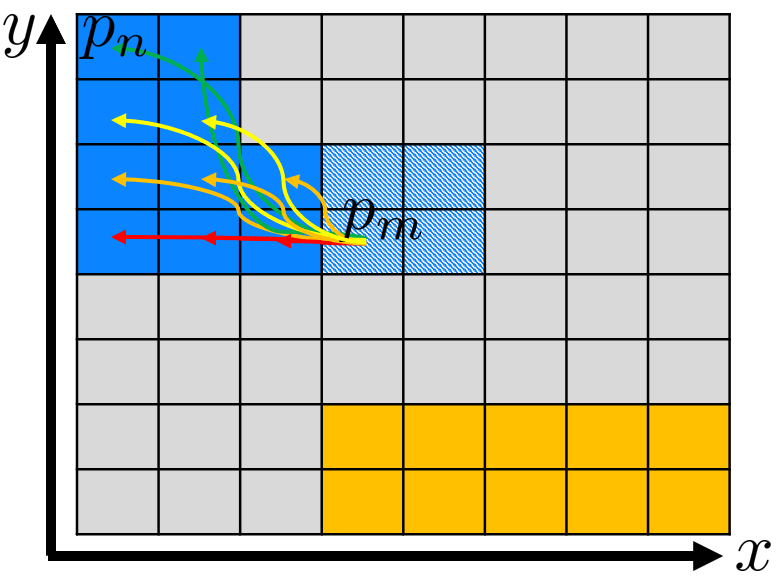}
\caption{We guide the learning of occluded disparity regions (dotted regions) by values of non-occluded regions of the same semantic class (distinct colors). A disparity/flow vector $p_m$ with a fixed position gets constrained by all other vectors $p_n$ within a certain neighborhood. Constrains between $p_m$ and $p_n$ that have a larger y-distance are weighted less (red means high weight, green means low weight), since corresponding disparity values should lie on the x-axis due to image rectification.}
\label{fig:occ_guide_scheme}
\vspace{-0.3cm}
\end{figure}

In previous works \cite{zhang2019dispsegnet}, occluded disparity values get post-processed by assigning the nearest not occluded value to the occluded ones. This method, however does not consider the context and thus fails if the nearest valid values are not belonging to the same object that is further or nearer. 

Instead, we guide the disparity predictions inside occluded areas using the semantic information at training time, similar to WSPS. Thus, we do not rely on any post-processing at testing time.
We formulate our loss for guiding occluded depth pixels as:

\begin{multline*}
    \mathcal{L}_{D_{\mathrm{occ}}}^D =\\
    \sum_{p_m \in P_{O_D}} \alpha_c \frac{\sum_{p_n \in \mathcal{N}(p_m)} \omega(p_n, p_m, Z(p_m)) \rho( \mid D(p_n) - D(p_m) \mid)}{\sum_{p_n \in \mathcal{N}(p_m) } \omega(p_n, p_m, Z(p_m))}\\
\end{multline*}

where:

\begin{equation}
\omega(p_n,p_m,z)=e^{-\frac{\mid p_{m_y} - p_{n_y} \mid}{\eta \theta(z)}}
\end{equation}

For every pixel location $p_m \in P_{O_D}$ that is occluded by $O_D$ we minimize the absolute difference between $D(p_m)$ and the value $D(p_n)$, where $p_n$ is a pixel location that is in $\mathcal{N}(p_m)$, which is the set of pixel locations within a square neighborhood around $p_m$ that have the same semantic class as $p_n$ and that are not occluded. Since all stereo pairs are rectified, the stereo matches are located along the horizontal axes. We therefore down-weight the differences between $p_m$ and $p_n$ by the distance of their y-coordinates $p_{m_{y}}$ and $p_{n_{y}}$. Further, we down-weight the influence of the loss proportional to the depth $z$ at $p_m$ and a constant factor $\eta$ similar to WSPS.
As in WSPS, we leverage the robust loss function $\rho$ \cite{barron2019general},  where we set $\alpha$ to $0.5$ and $c$ to $0.1$. We illustrate the method in Fig. \ref{fig:occ_guide_scheme}

\subsubsection{Sky Regularization}

Since sky has infinite depth and weak visual features, previous methods often predict unreasonable optical flow and disparity values in these regions or foreground-predictions 'bleed' into the sky regions. To overcome this problem, we propose to minimize the optical flow or disparity in regions that belong to the class \textit{sky} using the segmentation map from our semantic network. Specifically we formulate our loss as:

\begin{equation}
    \mathcal{L}_{\mathrm{sky}}^{M} = \frac{1}{N} \sum_{p_m \mid S_{M}(p_m)=\mathrm{sky}} \mid\mid M(p_m) \mid\mid_2)
\end{equation}

where $N$ is the number of pixels belonging to the class \textit{sky}.

\subsubsection{3D Coupling}

While other works \cite{wang2019unos, zou2018df, epc++} propose a minimization of the difference between the rigid flow and the optical flow within rigid parts of the image, we instead constrain the network directly in 3D space.

First, a source and a target depth map $Z_{\mathscr{S}}$ and $Z_{\mathscr{T}}$ are back-projected to 3D, resulting in the point-clouds $P_{\mathscr{S}}$ and $P_{\mathscr{T}}$. Second, we synthesize $\hat{P_{\mathscr{T}}}$ using the predicted optical flow $F_{{\mathscr{T}}}$ and $P_{\mathscr{S}}$. Third, we transform the synthesized pointcloud  $\hat{P_{\mathscr{T}}}$ by the ego-motion $T_{ego_{{\mathscr{S}} \rightarrow {\mathscr{T}}}}$ in order to align it with the target pointcloud $P_{\mathscr{T}}$. We then formulate a loss function that minimizes the euclidean distance between the two pointclouds for the static parts of the scene, which are not occluded by $O_{F_{\mathscr{T}}}$.

 We define our loss as:

\begin{equation}
\begin{aligned}
    \mathcal{L}_{\mathrm{3D_{\mathrm{rigid}}}}^{{\mathscr{S}} \rightarrow {\mathscr{T}}} = \theta(Z_{\mathscr{T}}) O_{F_{\mathscr{T}}} (1-C_{\mathscr{T}}) (\mid\mid P_{\mathscr{T}} - T_{ego_{{\mathscr{S}} \rightarrow {\mathscr{T}}}}\hat{P_{\mathscr{T}}} \mid\mid_2)
\end{aligned}
\end{equation}

Here, $C_{\mathscr{T}}$ denotes a map which is 1 at locations that are dynamic and 0 otherwise.
For far points the depth tends to be inaccurate due to the limited resolution of the cameras.
Therefore, with $\theta(Z) = \frac{z^2}{bf}$ we weight the euclidean loss inversely proportional to an approximated quadratic depth error as in Eq. \ref{eq:dispError}, making our loss function numerically stable.

\subsubsection{Dynamic Detection and Instance 3D Loss}

We segment dynamic objects using the predicted disparities $D_{l_{t}}, D_{l_{t+1}}$, optical flow $F_{l_t}$ and semantic instances $\mathscr{I}_{l_{t}}$, $\mathscr{I}_{l_{t+1}}$ corresponding to the left consecutive images. As in previous sections, we define a source or target map as either being related to time-step $t$ or the consecutive $t+1$.  As for the WSPS term, we infer the instances with Mask-RCNN \cite{he2017mask}. For every instance $i$ of the \textit{vehicle} class we calculate the SE(3) transformation $T_{{\mathscr{S}} \rightarrow {\mathscr{T}}}^i$, which represents the rigid motion of the corresponding dynamic object.
Therefore, we compose the total motion that is needed to transform the aligned instance pointcloud $ \hat{P_{{\mathscr{T}}}^i} = W(P_{{\mathscr{S}}}^i, F_{{\mathscr{T}}})$ to the corresponding target instance pointcloud $P_{{\mathscr{T}}}^i$, by combining the predicted ego-motion $T_{\mathrm{ego}_{\mathscr{S}} \rightarrow {\mathscr{T}}}$ with the dynamic motion of the individual object $T_{{\mathscr{S}} \rightarrow {\mathscr{T}}}^i$.

We define the instance-object motions as a optimization problem as follows:

\begin{equation}
\begin{aligned}
T_{{\mathscr{S}} \rightarrow {\mathscr{T}}}^i = \underset{T}{\mathrm{argmin}}[ Q \mid\mid T T_{\mathrm{ego}_{{\mathscr{S}} \rightarrow {\mathscr{T}}}} \hat{P_{{\mathscr{T}}}^i} - P_{{\mathscr{T}}}^i \mid \mid_2 ]
\end{aligned}
\end{equation}

where:\\
\begin{equation}
Q = e ^ {-( \mid\mid \hat{P_{{\mathscr{T}}}^i} - P_{{\mathscr{T}}} \mid \mid_2 - \mathrm{median}(\mid\mid \hat{P_{{\mathscr{T}}}^i} - P_{\mathscr{T}} \mid \mid_2) )}
\end{equation}

Here, $Q$ is a weighting term reduces the influence of outliers. We solve the optimization problem with a differentiable SVD \cite{giles2008extended}.

Using the estimated motion of the rigid instances we then present a loss function that complements our 3D coupling loss for rigid regions with additional constraints for dynamic rigid instances:

\begin{equation}
    \mathcal{L}_{3D_{\mathrm{dyn}}}^{{\mathscr{S}} \rightarrow {\mathscr{T}}} = \sum_{i \in \mathscr{I}} \theta(D_{\mathscr{T}}^i)  \mid\mid T_{a}^i \hat{P_{\mathscr{T}}^i} - P_{\mathscr{T}}^i \mid \mid_2
\end{equation}

where $D_{\mathscr{T}}^i$ refers to the average disparity of object instance $i$ and $T_{a}^i = T_{{\mathscr{S}} \rightarrow {\mathscr{T}}}^i T_{\mathrm{ego}_{{\mathscr{S}} \rightarrow {\mathscr{T}}}}$ is the apparent instance motion corresponding to instance $i$.
For solving the data association of the target and source instance masks, we use the optical flow predictions to warp the source instance masks to the target instance masks and calculate the IoU score for all combinations between source and target instances. Finally we use the hungarian algorithm \cite{kuhn1955hungarian} to assign the best association. Associations with an IoU score under 0.5 are not considered for the loss function.

Given all instance transformations $T_{{\mathscr{S}} \rightarrow {\mathscr{T}}}^\mathscr{I}$, we compile the dynamic object segmentation map $C_\mathscr{T}$ where all pixels are set to 1 that belong to instances that have a speed over $15$ $km/h$. All other pixels are set to 0. Objects that belong to a non-rigid class such as \textit{person} are set always to dynamic during training.

Combining both 3D losses for rigid and dynamic regions we define the overall 3D loss as:
\begin{equation}
    \mathcal{L}_{3D}^{{\mathscr{S}} \rightarrow {\mathscr{T}}} = \mathcal{L}_{\mathrm{3D_{\mathrm{rigid}}}}^{{\mathscr{S}} \rightarrow {\mathscr{T}}} +   \mathcal{L}_{3D_{\mathrm{dyn}}}^{{\mathscr{S}} \rightarrow {\mathscr{T}}}
\end{equation}

\subsubsection{Occlusion Filling within Optical Flow Maps using 3D Geometry and Motion}

\begin{figure}
\centering
\includegraphics[width=\linewidth]{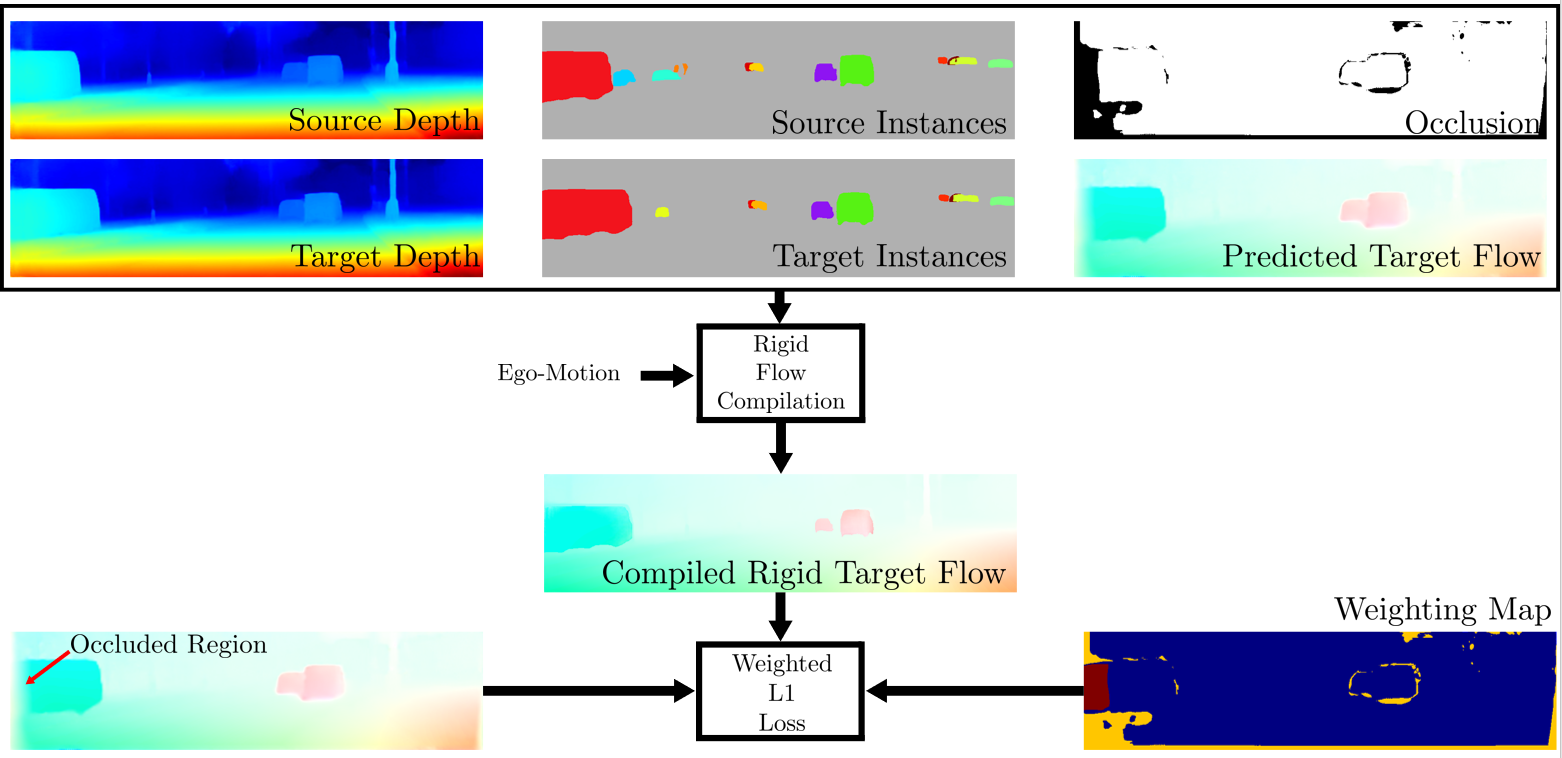}
\caption{This figure illustrates our proposed rigid optical flow occlusion filling loss. Leveraging the depth, optical flow, occlusion, ego-motion and object-instance estimations we compute SE(3) transformations for all individual instances. We then compile a optical flow from all corresponding rigid transformations and use it as guidance for the predicted flow in occluded regions. We employ a L1 loss in which the occluded object instance pixels are weighted by a quality measure that is based on the 3D reconstruction errors.}
\label{fig:foccscheme}
\vspace{-0.3cm}
\end{figure}

For filling occluded areas in the optical flow maps previous works \cite{epc++} propose to fill occluded pixels with the information of the rigid flow $F^R$. This method however, does handle partially occluded dynamic moving rigid objects.
We propose an enhanced version that leverages all predicted instance motions $T^\mathscr{I}$ to create rigid flow maps for all pixels.  We formulate our occlusion loss as:

\begin{equation}
\begin{aligned}
\mathcal{L}_{\mathrm{F_{occ}}}^{{\mathscr{S}} \rightarrow {\mathscr{T}}} =  \gamma(1-O_{F_{{\mathscr{T}}}}) (1-C_{\mathscr{T}}) \mid F_{{\mathscr{T}}} - F_{{\mathscr{T}}}^{R} \mid + (1-O_{F_{{\mathscr{T}}}}) C_{\mathscr{T}} \\
\cdot  \sum_{i \in \mathscr{I}} \sigma (T_{a}^i \hat{P_{\mathscr{T}}^i}, P_{\mathscr{T}}^i) \mid \mid  (K T_{a}^i \hat{P_{\mathscr{T}}^i} - K P_{\mathscr{T}}^i) - F_{{\mathscr{T}}} \mid \mid_1
\end{aligned}
\end{equation}

where $\sigma({P_{\mathscr{S}}^i}, P_{\mathscr{T}}^i) = e^{- \mid\mid {P_{\mathscr{S}}^i} - P_{\mathscr{T}}^i \mid \mid_2}$ denotes the reconstruction error between the transformed source pointcloud and the target pointcloud and serves as a measure of the quality of $T_{{\mathscr{S}} \rightarrow {\mathscr{T}}}^i$.
The constant factor $\gamma$ is set to $0.1$ in all experiments.
We present a schematic of our flow occlusion filling loss on Fig. \ref{fig:foccscheme}.

\subsubsection{Total Loss}
Our final loss function is composed by a weighted combination of all previously presented losses:

\begin{equation}
\begin{aligned}
\mathcal{L} = \lambda_1 \mathcal{L}_{p}^{M_1} + \lambda_2 \mathcal{L}_{s}^{M_1} + \lambda_3 \mathcal{L}_{\mathrm{WSPS}}^{M_1}+ \lambda_4 \mathcal{L}_{\mathrm{sky}}^{M_1} + \lambda_5 \mathcal{L}_{\mathrm{D_{occ}}}^{M_2}\\
  + \lambda_6 (\mathcal{L}_{3D}^{t \rightarrow (t+1)} + \mathcal{L}_{3D}^{(t+1) \rightarrow t}) + \\
  + \lambda_7 (\mathcal{L}_{\mathrm{Focc}}^{t \rightarrow (t+1)} + \mathcal{L}_{\mathrm{F_{occ}}}^{(t+1) \rightarrow t}) + \lambda_8 (\mathcal{L}_{\mathrm{occ}}^{M_1})
\end{aligned}
\end{equation}

Here, $M_1$ means that the loss is applied with $F_{l_t}$, $F_{l_{(t+1)}}$, $F_{l_t}^R$, $F_{l_{(t+1)}}^R$, $D_{l_t}$ and $D_{r_t}$, while $M_2$ means that the loss applied on $D_{l_t}$ and $D_{r_t}$. We apply $\mathcal{L}_{p}$, $\mathcal{L}_{s}$ and $\mathcal{L}_{\mathrm{sky}}$ on all available levels. For computational efficiency the smoothing loss $\mathcal{L}_{WSPS}$ and depth occlusion guidance loss $\mathcal{L}_{\mathrm{Docc}}$ are only applied to the last two levels 4 and 5, while the 3D reconstruction loss $\mathcal{L}_{3D}$ as well as the flow occlusion filling loss $\mathcal{L}_{\mathrm{Focc}}$ are applied only on the last level 5. We weight the individual loss terms by the hyperparameters $\lambda_1 ... \lambda_8$.

\section{Experimental Results}
\label{sec:reults}

\begin{figure*}
\scriptsize 
\centering 
\setlength{\tabcolsep}{0.3em}
\renewcommand{\arraystretch}{1}
\begin{tabular}{P{0.3cm} P{3cm} P{3cm} P{3cm} P{3cm} P{3cm}}
& Left/Right Image & Ours & UnOS \cite{wang2019unos} & DispSegNet \cite{zhang2019dispsegnet} & DispNetC \cite{mayer2016large} \\
\parbox[t]{2mm}{\multirow{2}{*}{\rotatebox[origin=c]{90}{Disparity}}}
& \includegraphics[width=.97\linewidth]{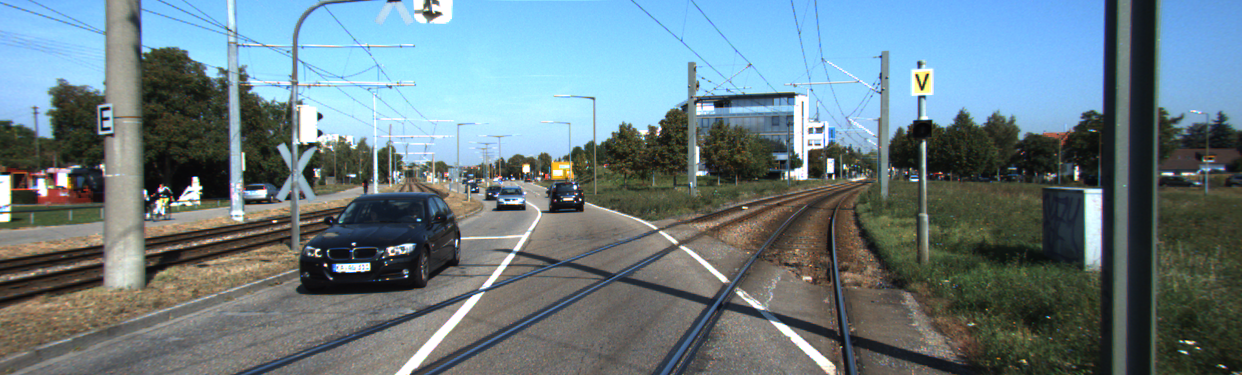}
& \includegraphics[width=.97\linewidth]{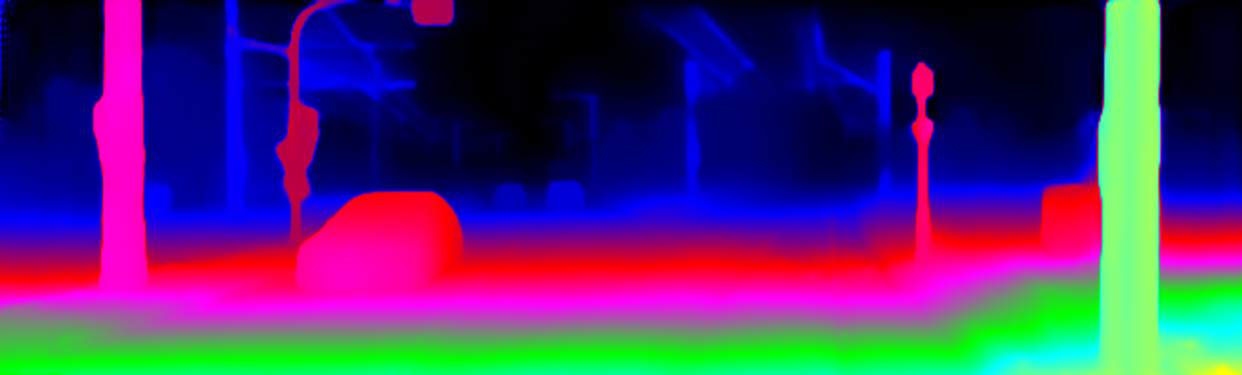}
& \includegraphics[width=.97\linewidth]{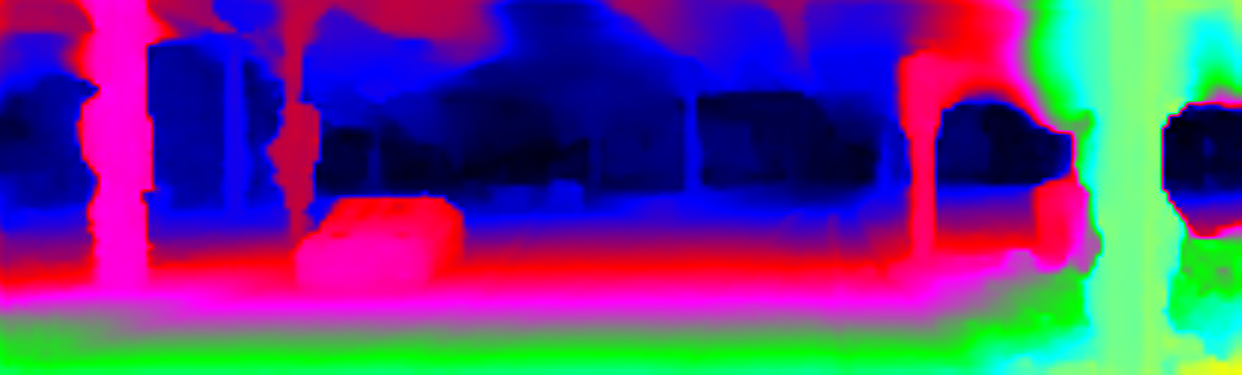}
& \includegraphics[width=.97\linewidth]{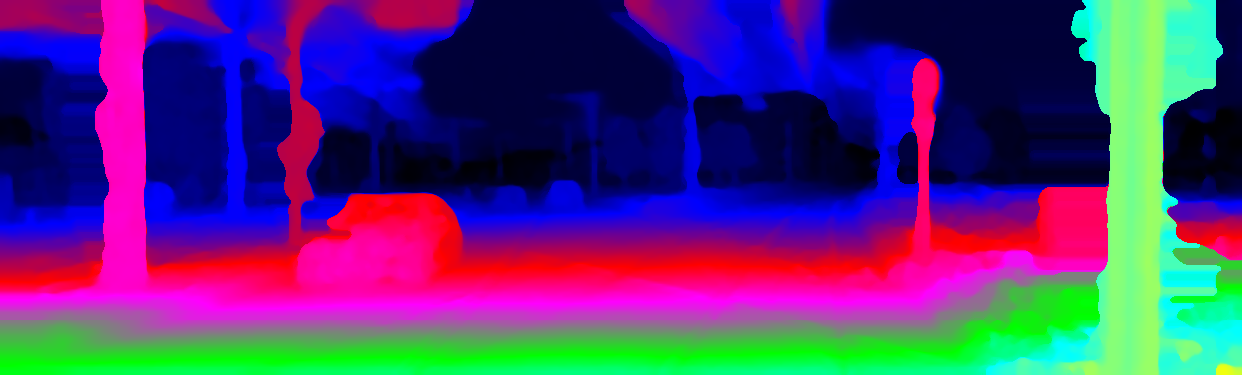}
& \includegraphics[width=.97\linewidth]{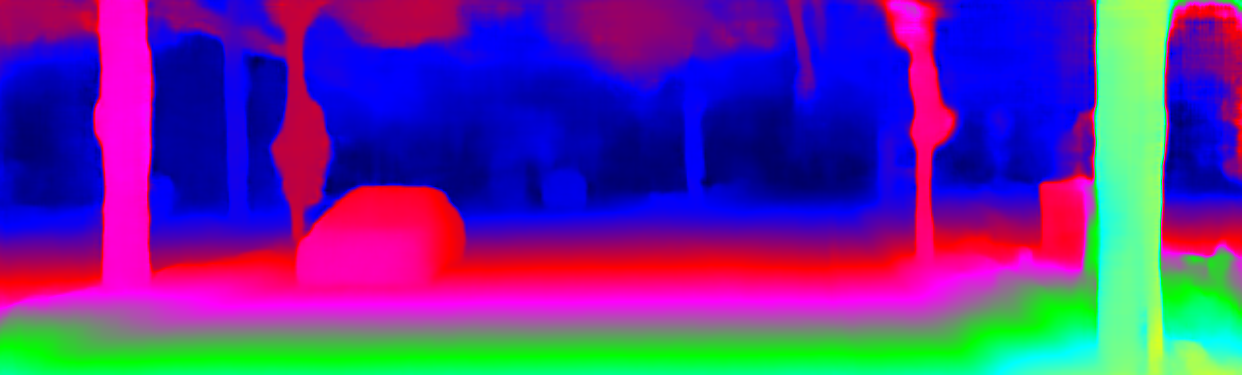} \\
& \includegraphics[width=.97\linewidth]{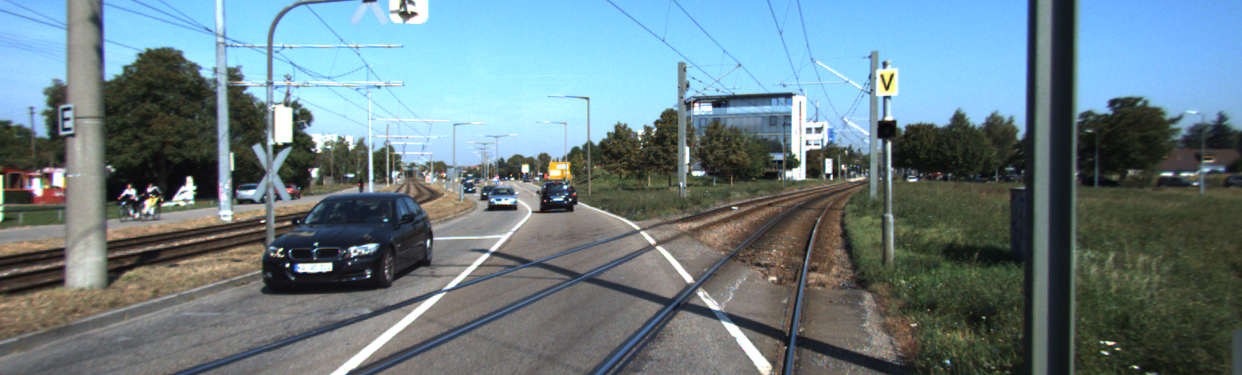}
& \begin{overpic}[width=0.97\linewidth,tics=10]{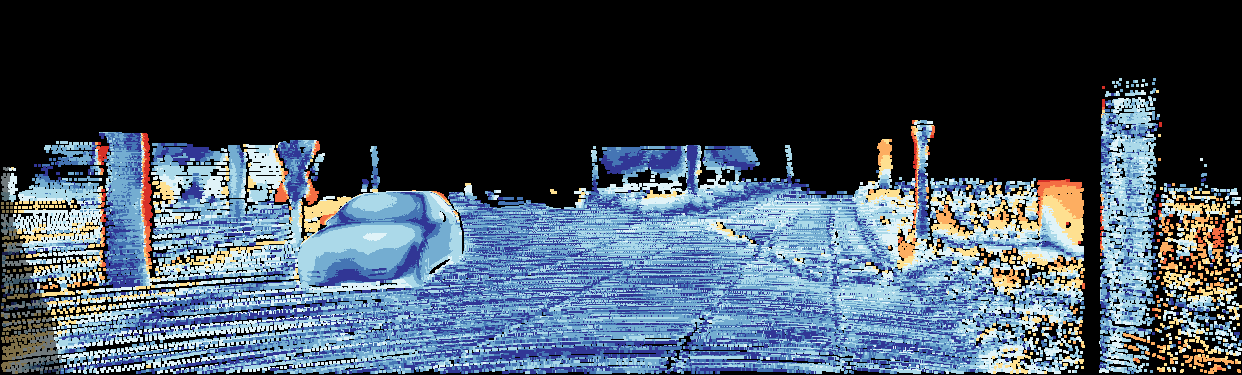}
 \put (60,25) {\textcolor{red}{\tiny $D1_{\mathrm{all}}=5.81$}}
\end{overpic}
& \begin{overpic}[width=0.97\linewidth,tics=10]{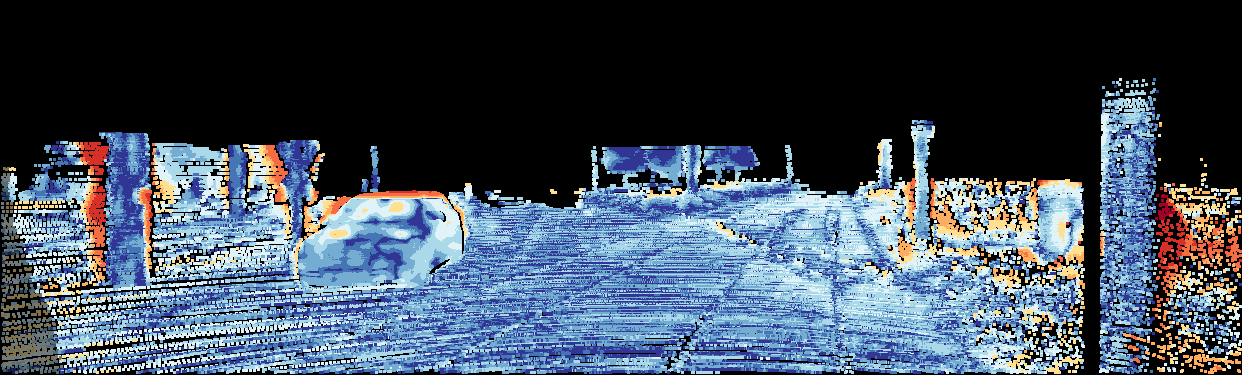}
 \put (60,25) {\textcolor{red}{\tiny $D1_{\mathrm{all}}=6.01$}}
\end{overpic}
& \begin{overpic}[width=0.97\linewidth,tics=10]{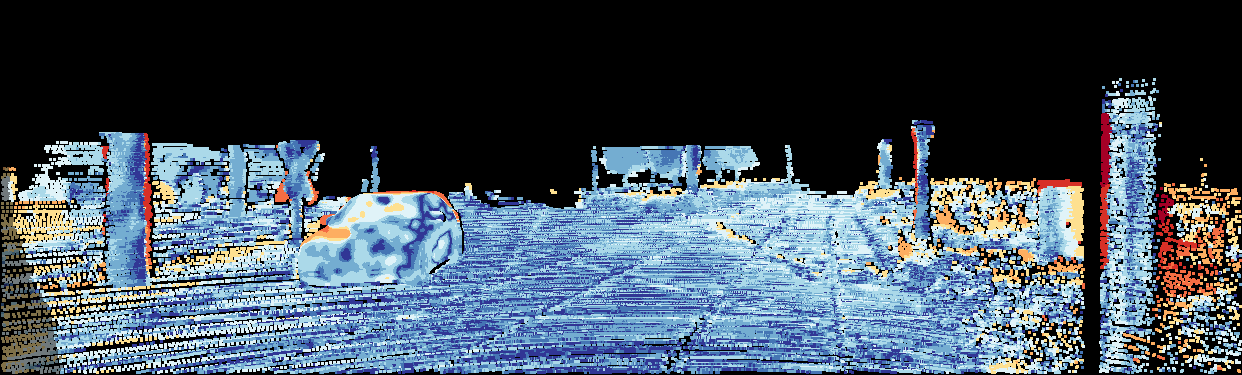}
 \put (60,25) {\textcolor{red}{\tiny $D1_{\mathrm{all}}=6.38$}}
\end{overpic}
& \begin{overpic}[width=0.97\linewidth,tics=10]{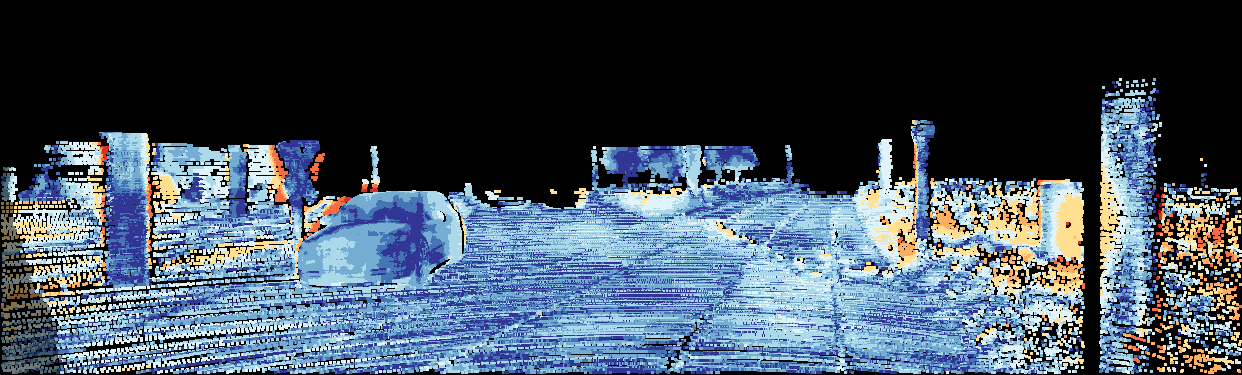}
 \put (60,25) {\textcolor{red}{\tiny $D1_{\mathrm{all}}=5.62$}}
\end{overpic}\\
\noalign{\smallskip}\hline\noalign{\smallskip}
& Left/Right Image & Ours & UnOS \cite{wang2019unos} & EPC++ \cite{epc++} & FlowNet2 \cite{IMKDB16} \\
\parbox[t]{2mm}{\multirow{2}{*}{\rotatebox[origin=c]{90}{Flow}}}
& \includegraphics[width=.97\linewidth]{publications/USegScene/figures/comparison_1/left_rgb.png}
& \includegraphics[width=.97\linewidth]{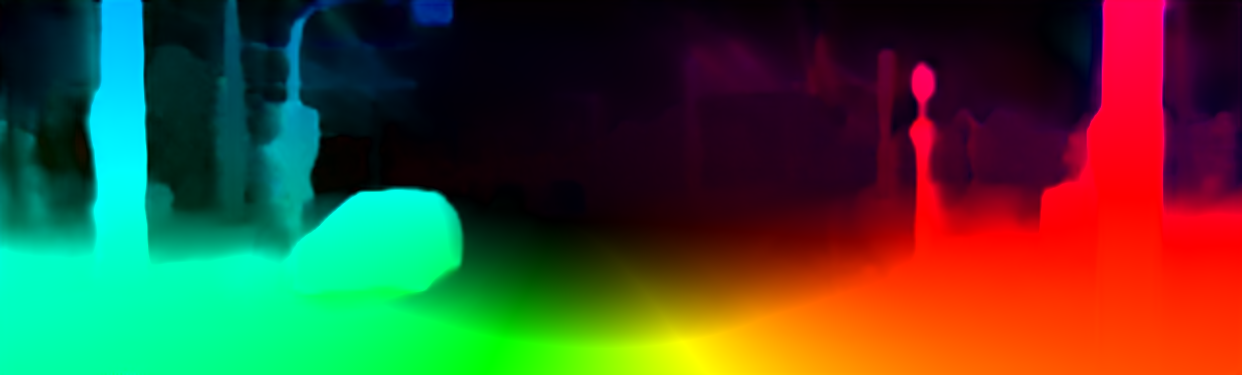}
& \includegraphics[width=.97\linewidth]{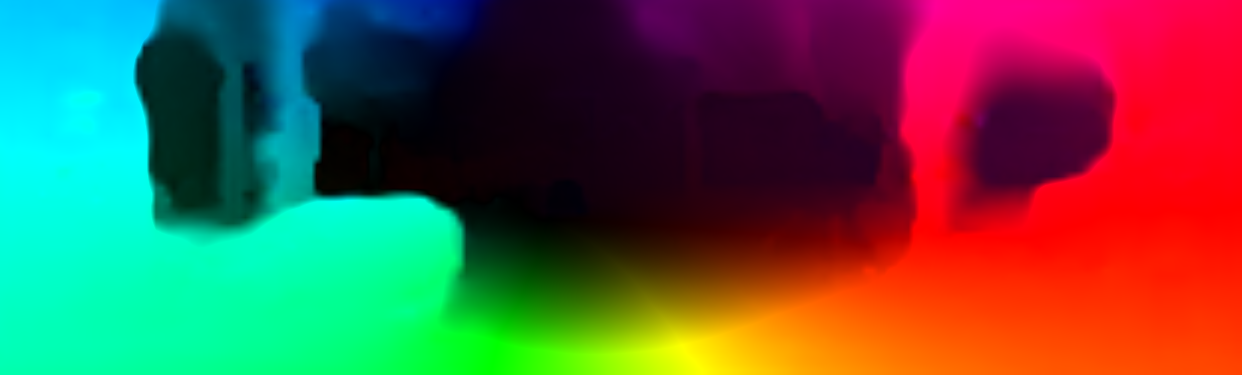}
& \includegraphics[width=.97\linewidth]{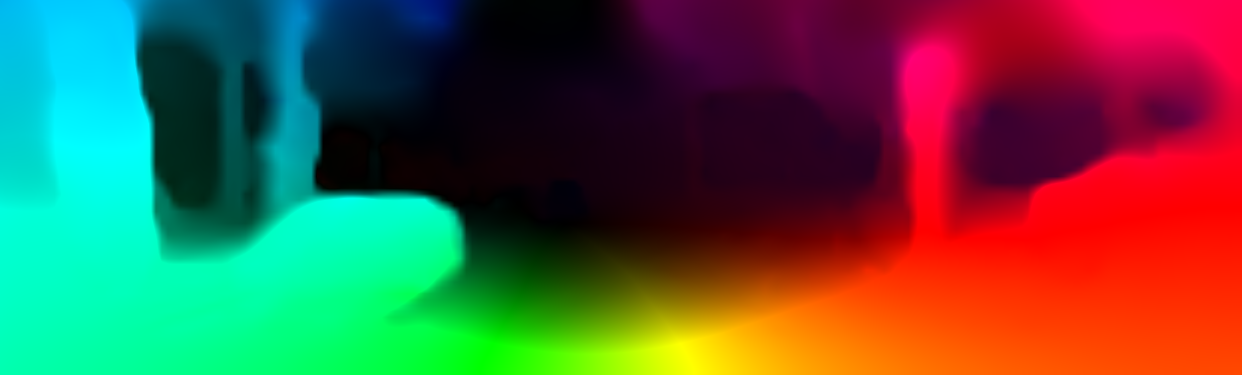}
& \includegraphics[width=.97\linewidth]{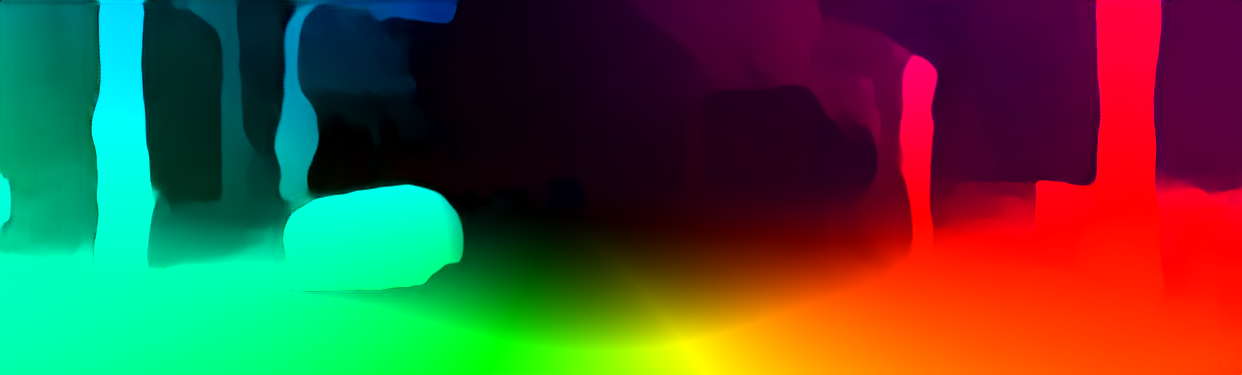} \\
& \includegraphics[width=.97\linewidth]{publications/USegScene/figures/comparison_1/right_rgb.png}
& \begin{overpic}[width=0.97\linewidth,tics=10]{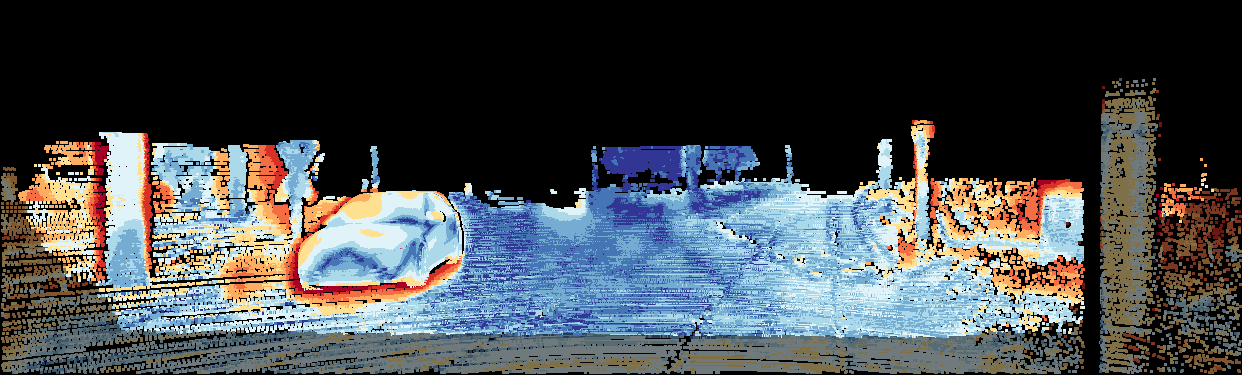}
 \put (60,25) {\textcolor{red}{\tiny $F1_{\mathrm{all}}=16.98$}}
\end{overpic}
& \begin{overpic}[width=0.97\linewidth,tics=10]{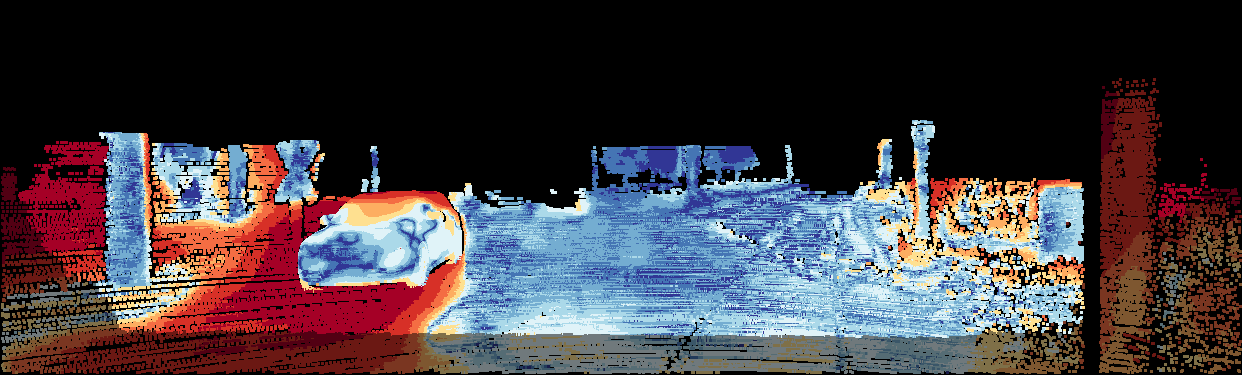}
 \put (60,25) {\textcolor{red}{\tiny $F1_{\mathrm{all}}=27.42$}}
\end{overpic}
& \begin{overpic}[width=0.97\linewidth,tics=10]{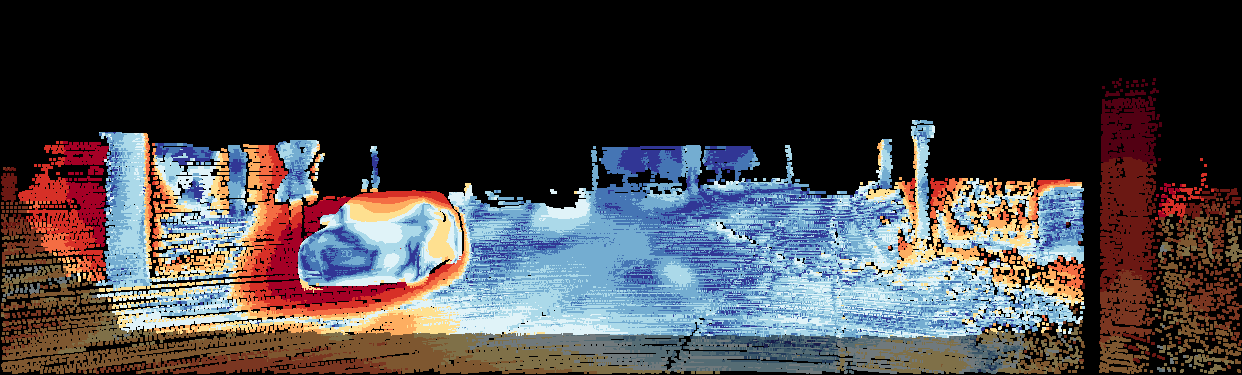}
 \put (60,25) {\textcolor{red}{\tiny $F1_{\mathrm{all}}=27.42$}}
\end{overpic}
& \begin{overpic}[width=0.97\linewidth,tics=10]{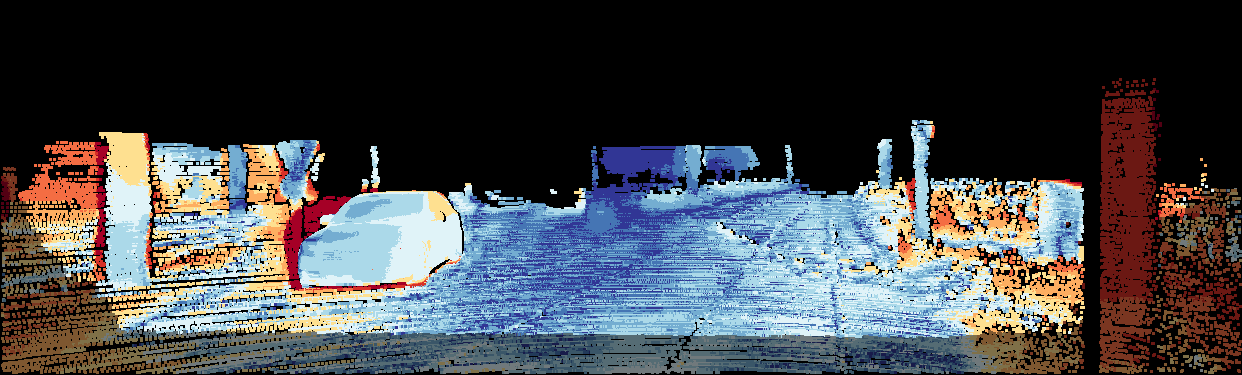}
 \put (60,25) {\textcolor{red}{\tiny $F1_{\mathrm{all}}=18.36$}}
\end{overpic}\\

\noalign{\smallskip}\hline\noalign{\smallskip}

& Left/Right Image & Ours & UnOS \cite{wang2019unos} & DispSegNet \cite{zhang2019dispsegnet} & DispNetC \cite{mayer2016large} \\
\parbox[t]{2mm}{\multirow{2}{*}{\rotatebox[origin=c]{90}{Disparity}}}
& \includegraphics[width=.97\linewidth]{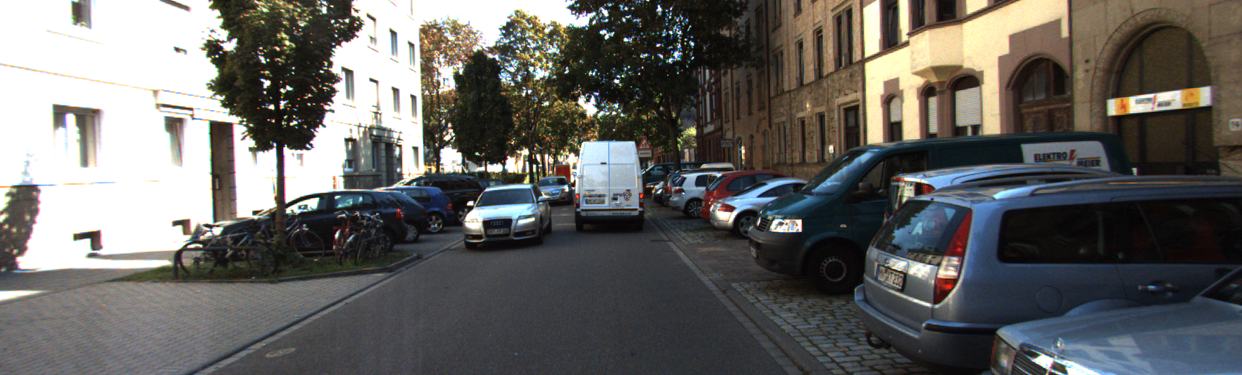}
& \includegraphics[width=.97\linewidth]{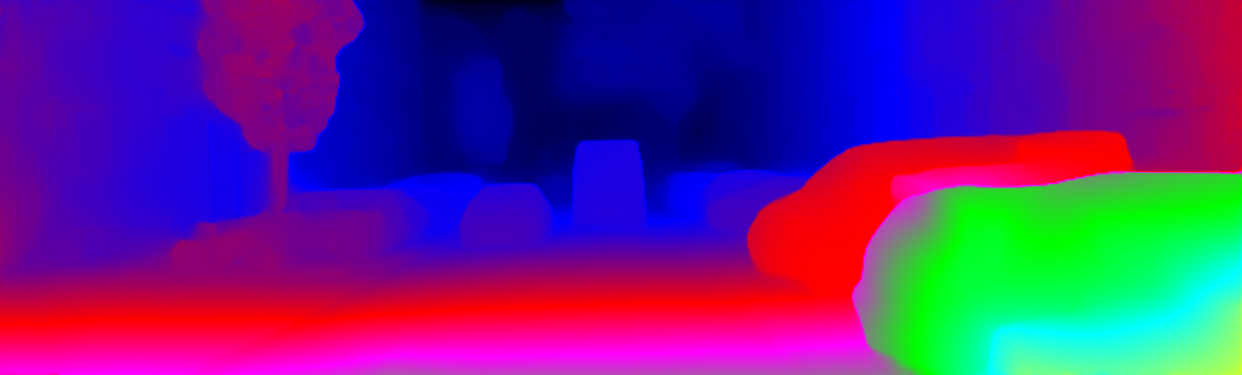}
& \includegraphics[width=.97\linewidth]{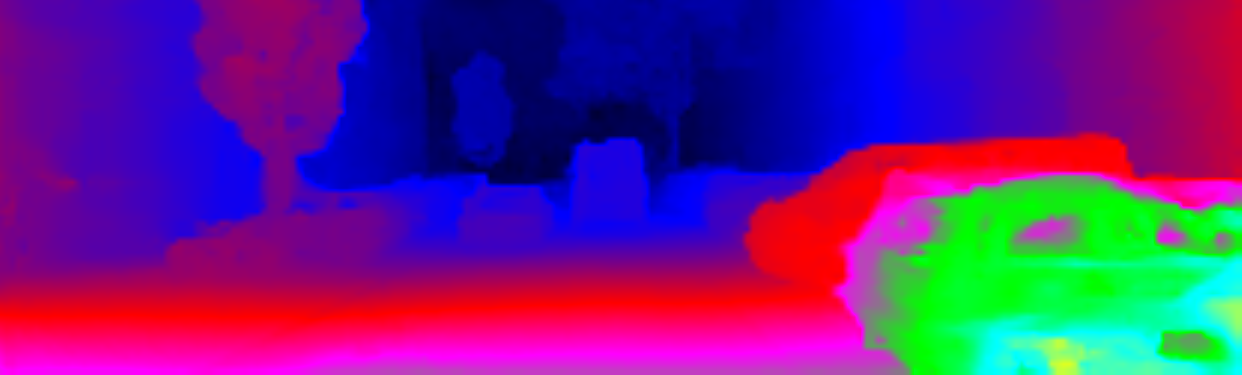}
& \includegraphics[width=.97\linewidth]{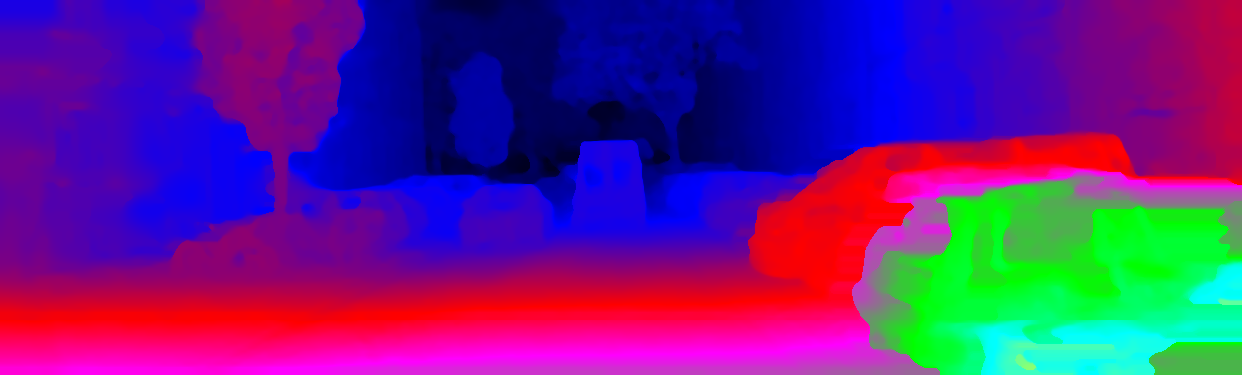}
& \includegraphics[width=.97\linewidth]{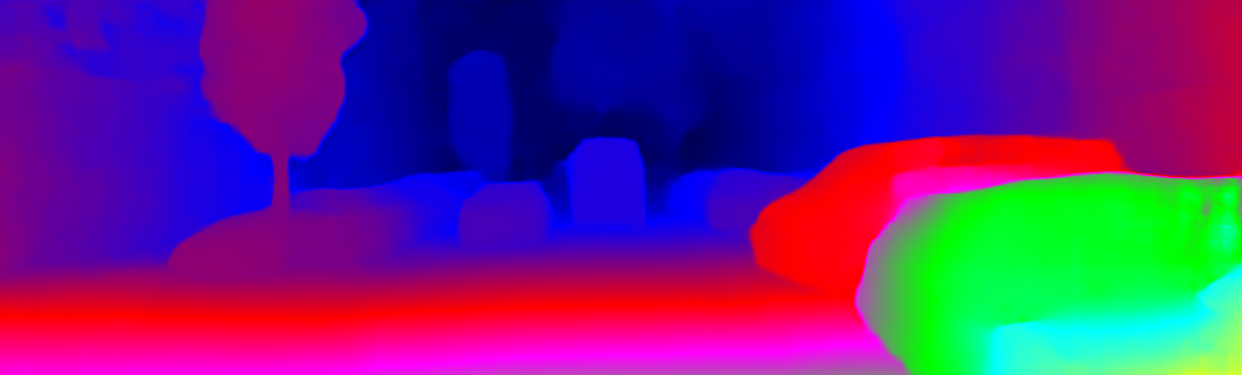} \\
& \includegraphics[width=.97\linewidth]{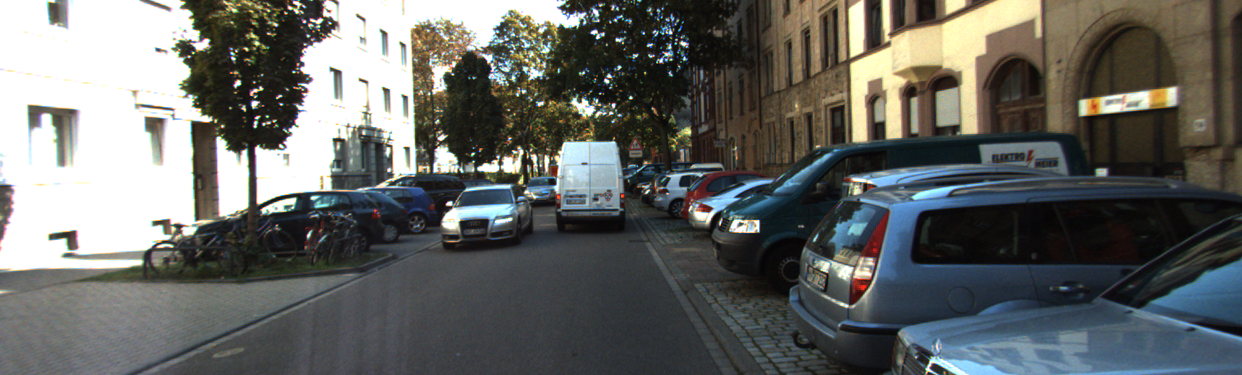}
& \begin{overpic}[width=0.97\linewidth,tics=10]{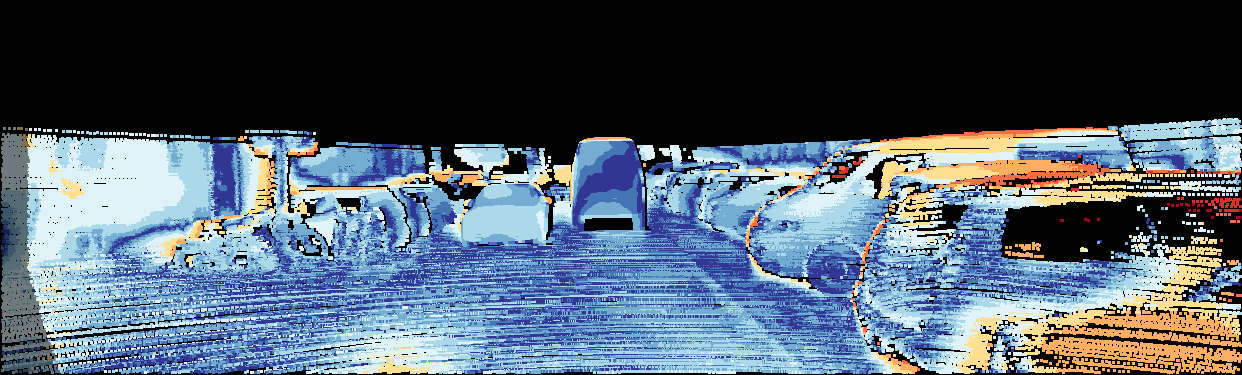}
 \put (60,25) {\textcolor{red}{\tiny $D1_{\mathrm{all}}=7.94$}}
\end{overpic}
& \begin{overpic}[width=0.97\linewidth,tics=10]{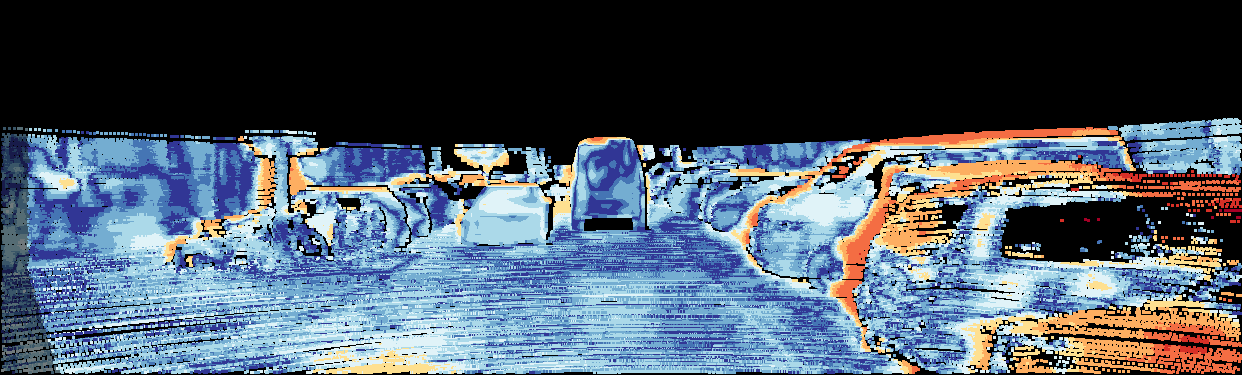}
 \put (60,25) {\textcolor{red}{\tiny $D1_{\mathrm{all}}=9.80$}}
\end{overpic}
& \begin{overpic}[width=0.97\linewidth,tics=10]{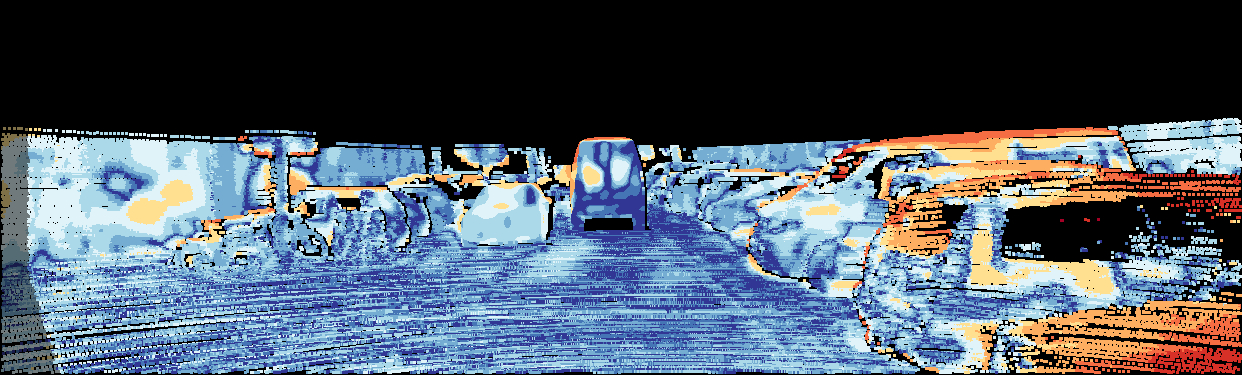}
 \put (60,25) {\textcolor{red}{\tiny $D1_{\mathrm{all}}=10.58$}}
\end{overpic}
& \begin{overpic}[width=0.97\linewidth,tics=10]{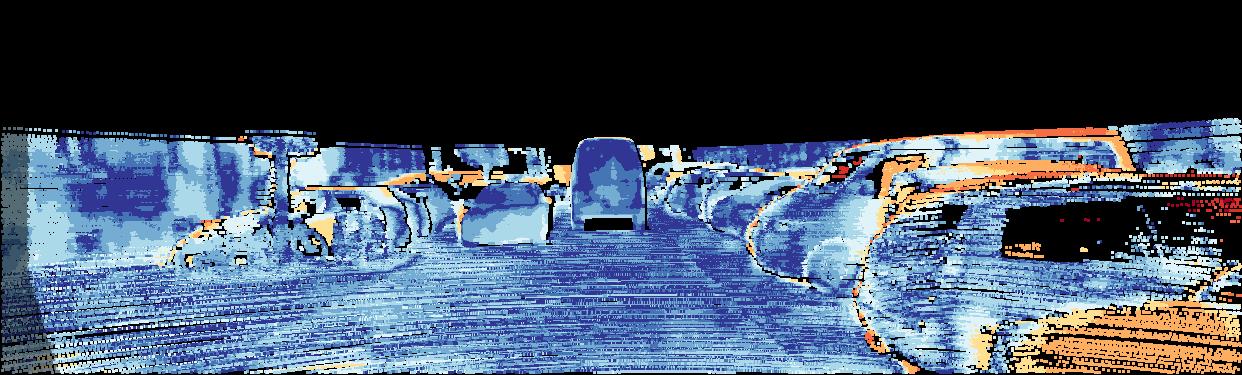}
 \put (60,25) {\textcolor{red}{\tiny $D1_{\mathrm{all}}=5.86$}}
\end{overpic}\\
\noalign{\smallskip}\hline\noalign{\smallskip}
& Left/Right Image & Ours & UnOS \cite{wang2019unos} & EPC++ \cite{epc++} & FlowNet2 \cite{IMKDB16} \\
\parbox[t]{2mm}{\multirow{2}{*}{\rotatebox[origin=c]{90}{Flow}}}
& \includegraphics[width=.97\linewidth]{publications/USegScene/figures/comparison_2/left_rgb.png}
& \includegraphics[width=.97\linewidth]{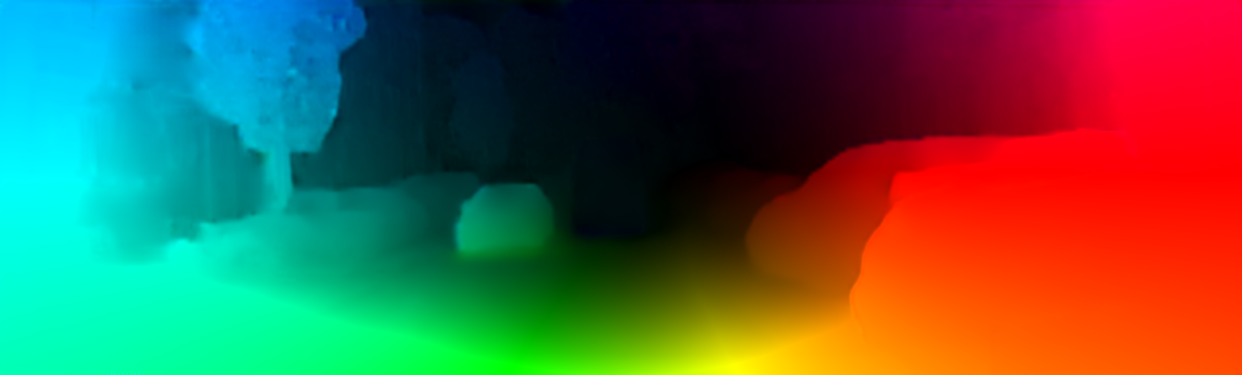}
& \includegraphics[width=.97\linewidth]{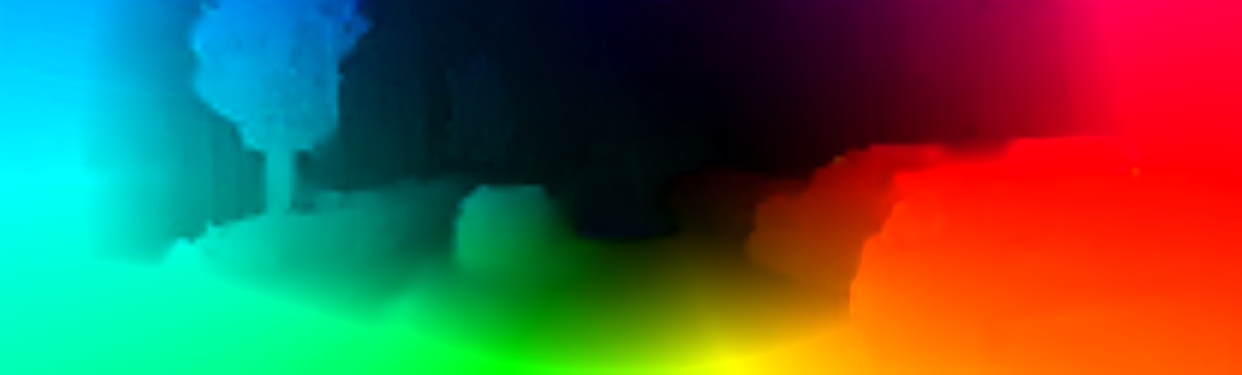}
& \includegraphics[width=.97\linewidth]{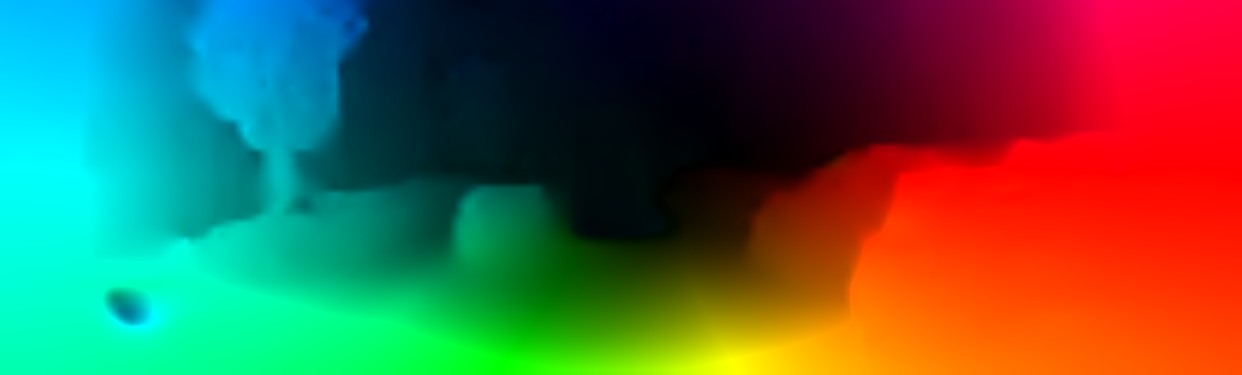}
& \includegraphics[width=.97\linewidth]{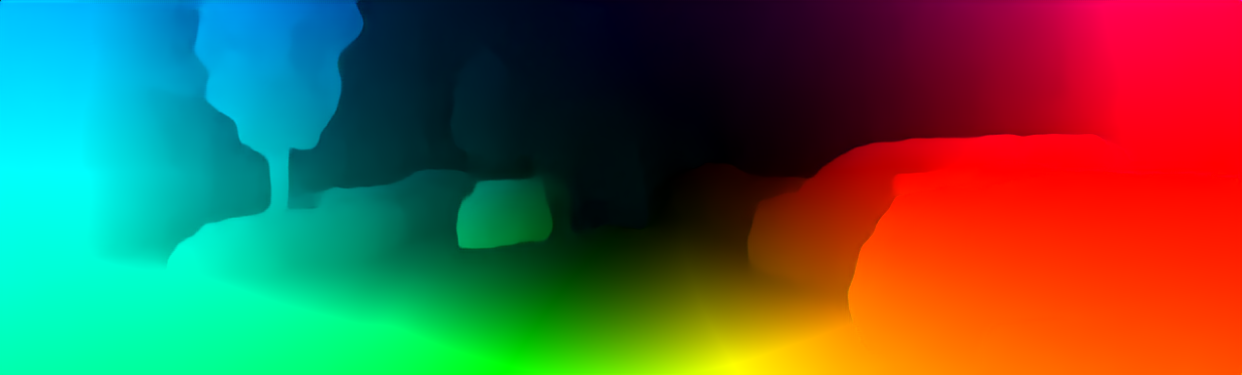} \\
& \includegraphics[width=.97\linewidth]{publications/USegScene/figures/comparison_2/right_rgb.png}
& \begin{overpic}[width=0.97\linewidth,tics=10]{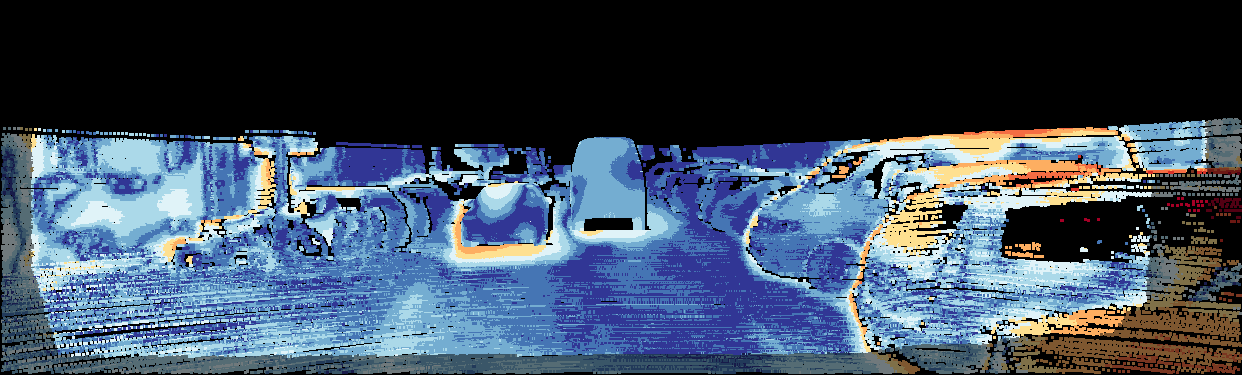}
 \put (60,25) {\textcolor{red}{\tiny $F1_{\mathrm{all}}=6.79$}}
\end{overpic}
& \begin{overpic}[width=0.97\linewidth,tics=10]{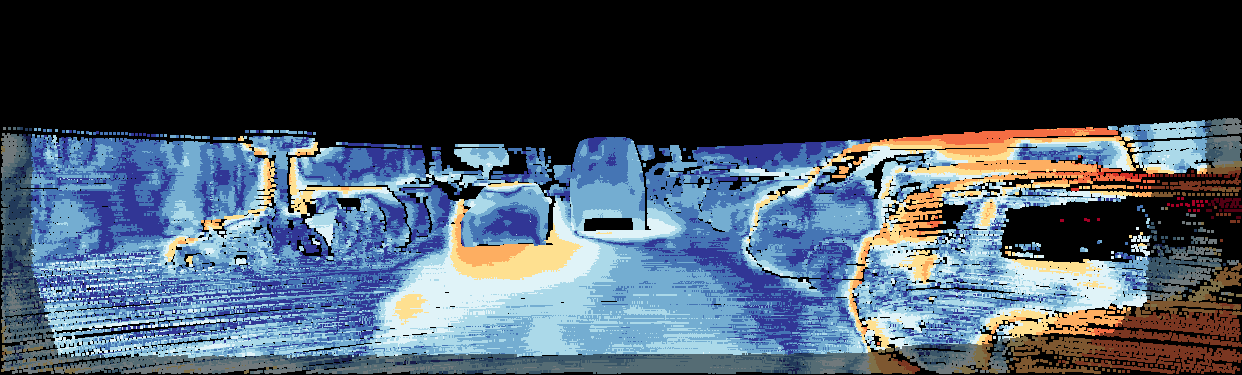}
 \put (60,25) {\textcolor{red}{\tiny $F1_{\mathrm{all}}=10.45$}}
\end{overpic}
& \begin{overpic}[width=0.97\linewidth,tics=10]{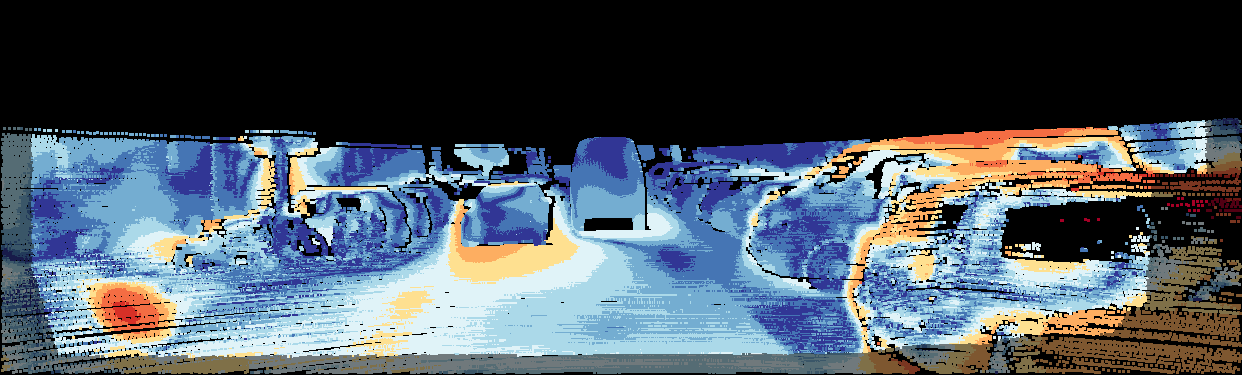}
 \put (60,25) {\textcolor{red}{\tiny $F1_{\mathrm{all}}=13.63$}}
\end{overpic}
& \begin{overpic}[width=0.97\linewidth,tics=10]{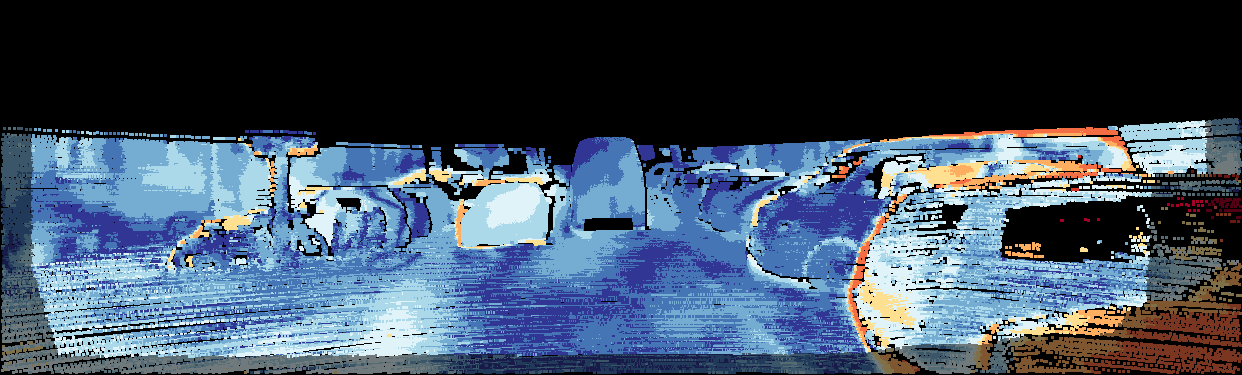}
 \put (60,25) {\textcolor{red}{\tiny $F1_{\mathrm{all}}=5.67$}}
\end{overpic}\\

\noalign{\smallskip}\hline\noalign{\smallskip}

& Left/Right Image & Ours & UnOS \cite{wang2019unos} & DispSegNet \cite{zhang2019dispsegnet} & DispNetC \cite{mayer2016large} \\
\parbox[t]{2mm}{\multirow{2}{*}{\rotatebox[origin=c]{90}{Disparity}}}
& \includegraphics[width=.97\linewidth]{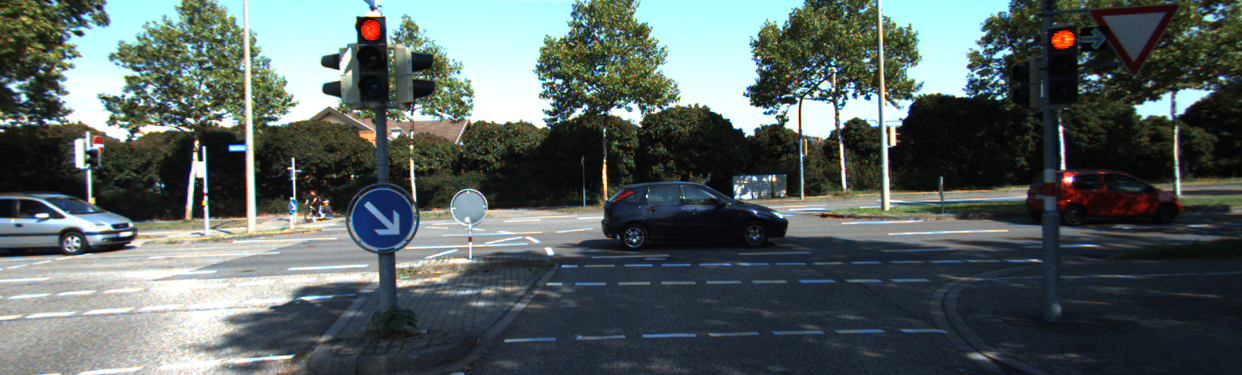}
& \includegraphics[width=.97\linewidth]{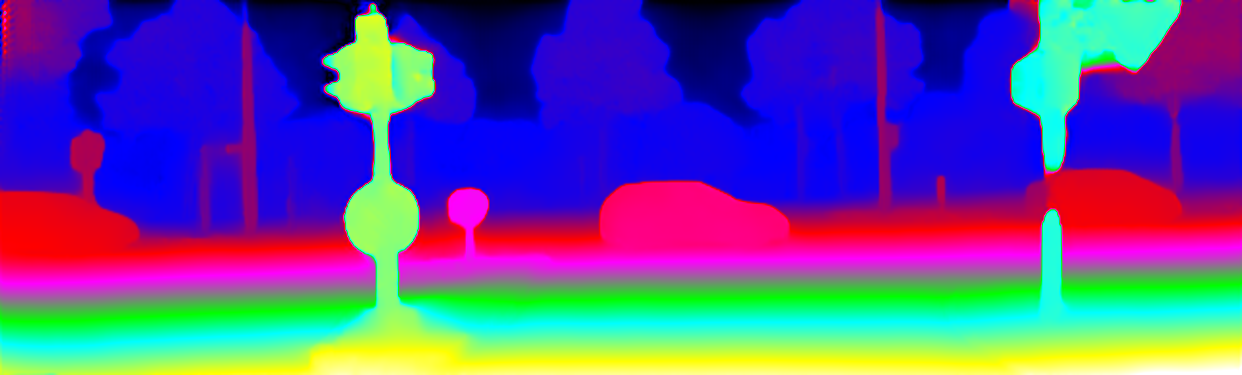}
& \includegraphics[width=.97\linewidth]{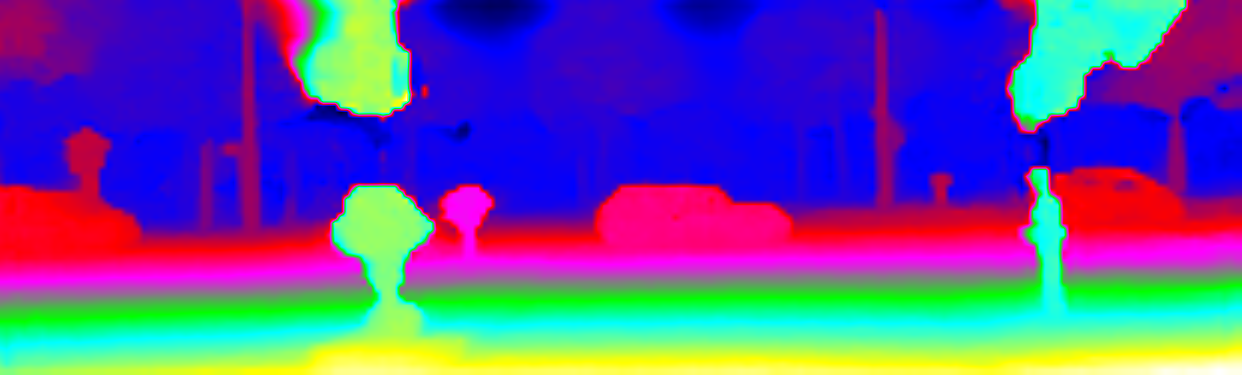}
& \includegraphics[width=.97\linewidth]{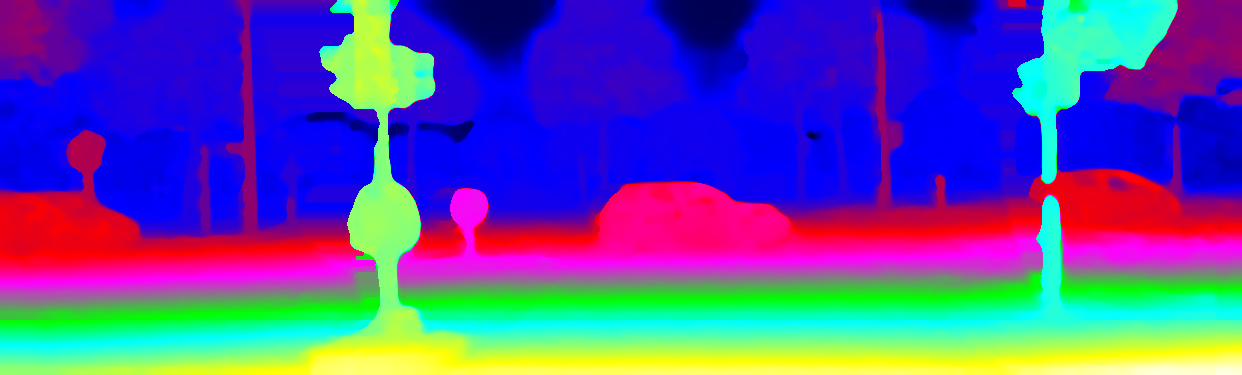}
& \includegraphics[width=.97\linewidth]{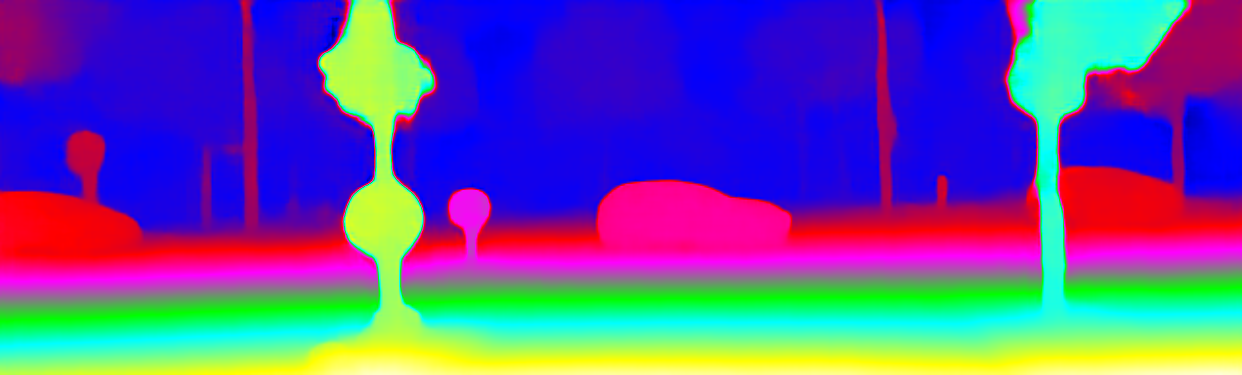} \\
& \includegraphics[width=.97\linewidth]{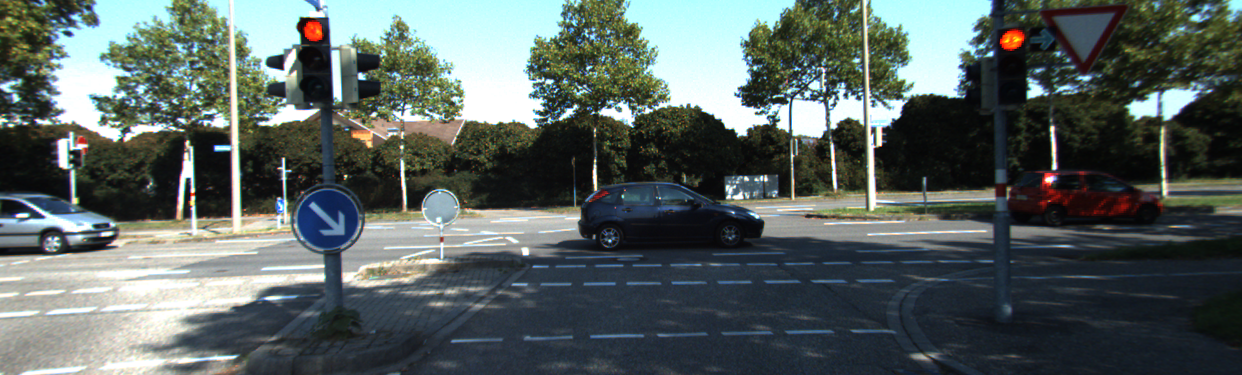}
& \begin{overpic}[width=0.97\linewidth,tics=10]{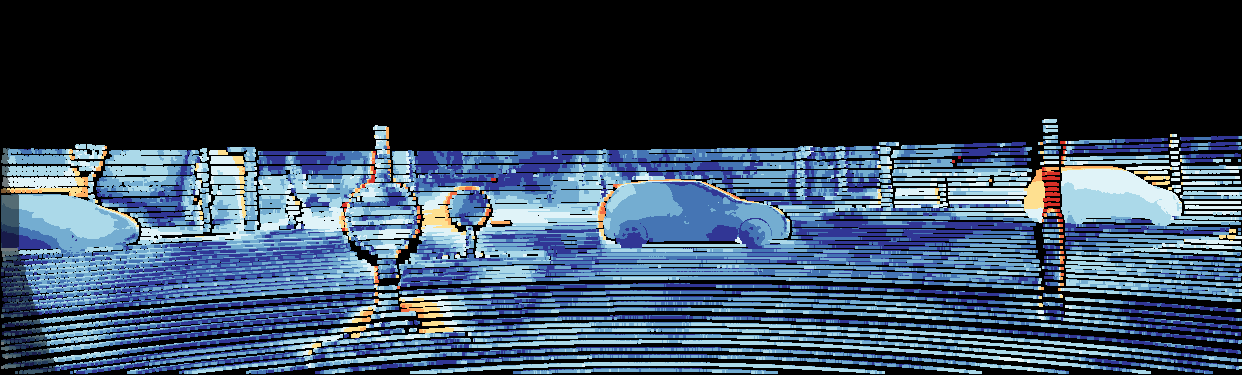}
 \put (60,25) {\textcolor{red}{\tiny $D1_{\mathrm{all}}=2.84$}}
\end{overpic}
& \begin{overpic}[width=0.97\linewidth,tics=10]{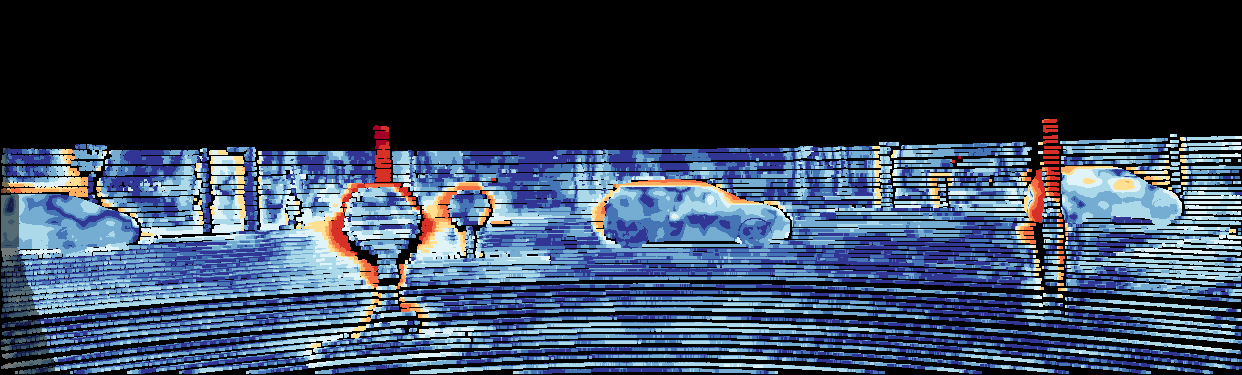}
 \put (60,25) {\textcolor{red}{\tiny $D1_{\mathrm{all}}=5.74$}}
\end{overpic}
& \begin{overpic}[width=0.97\linewidth,tics=10]{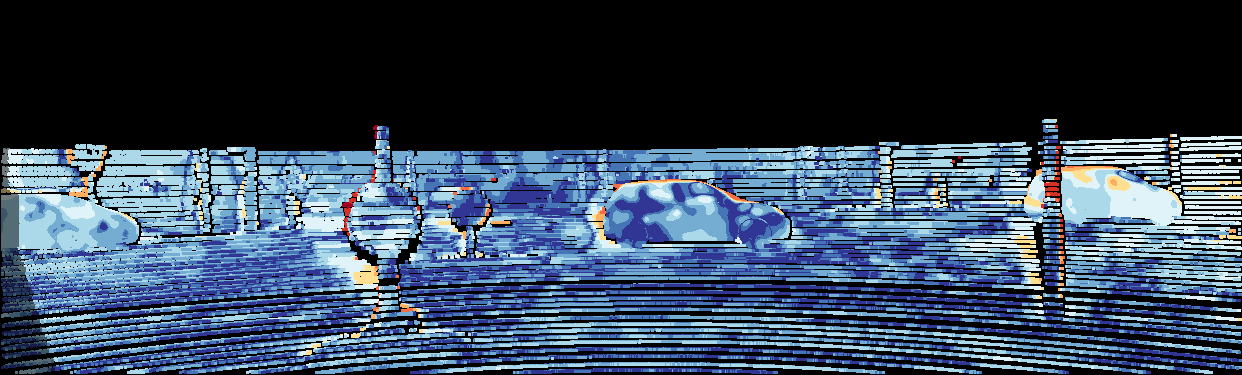}
 \put (60,25) {\textcolor{red}{\tiny $D1_{\mathrm{all}}=2.52$}}
\end{overpic}
& \begin{overpic}[width=0.97\linewidth,tics=10]{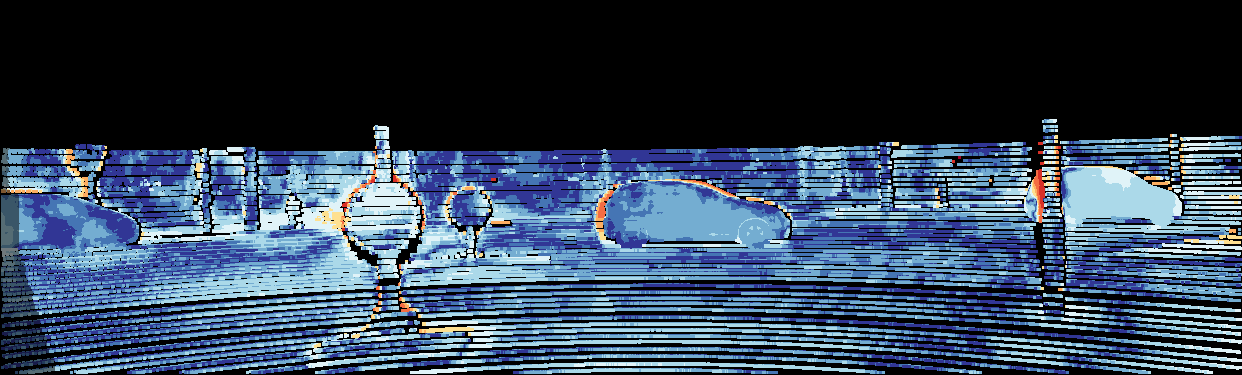}
 \put (60,25) {\textcolor{red}{\tiny $D1_{\mathrm{all}}=1.83$}}
\end{overpic}\\
\noalign{\smallskip}\hline\noalign{\smallskip}
& Left/Right Image & Ours & UnOS \cite{wang2019unos} & EPC++ \cite{epc++} & FlowNet2 \cite{IMKDB16} \\
\parbox[t]{2mm}{\multirow{2}{*}{\rotatebox[origin=c]{90}{Flow}}}
& \includegraphics[width=.97\linewidth]{publications/USegScene/figures/comparison_3/left_rgb.png}
& \includegraphics[width=.97\linewidth]{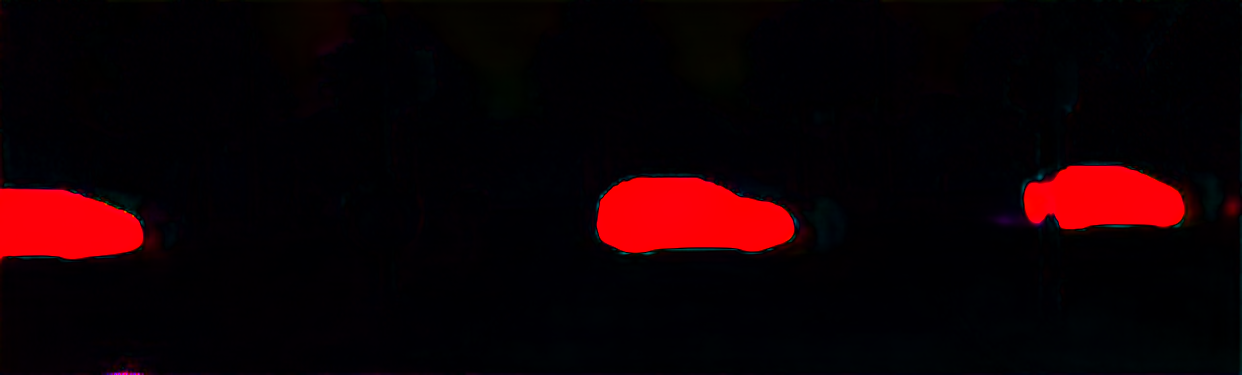}
& \includegraphics[width=.97\linewidth]{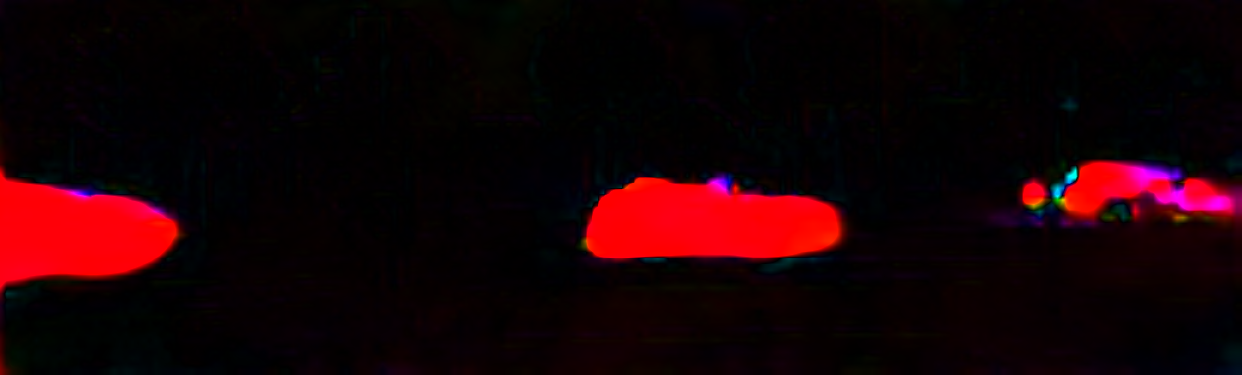}
& \includegraphics[width=.97\linewidth]{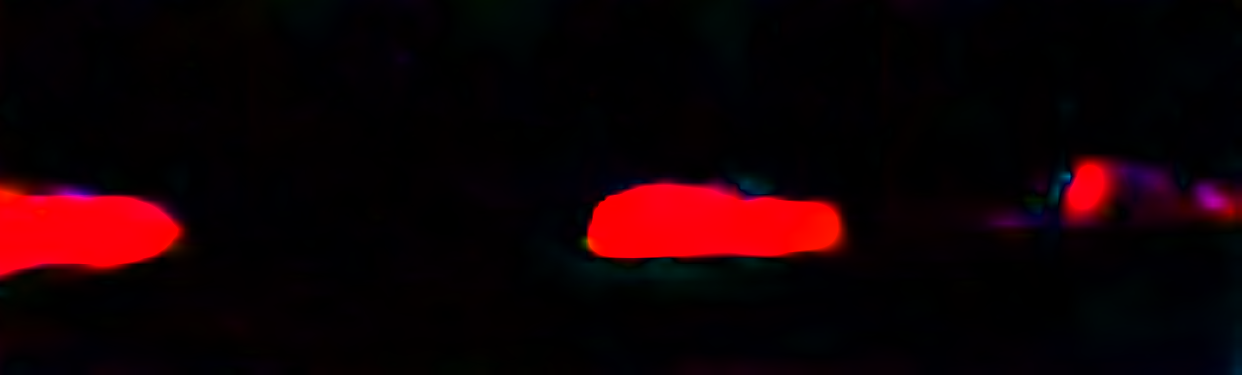}
& \includegraphics[width=.97\linewidth]{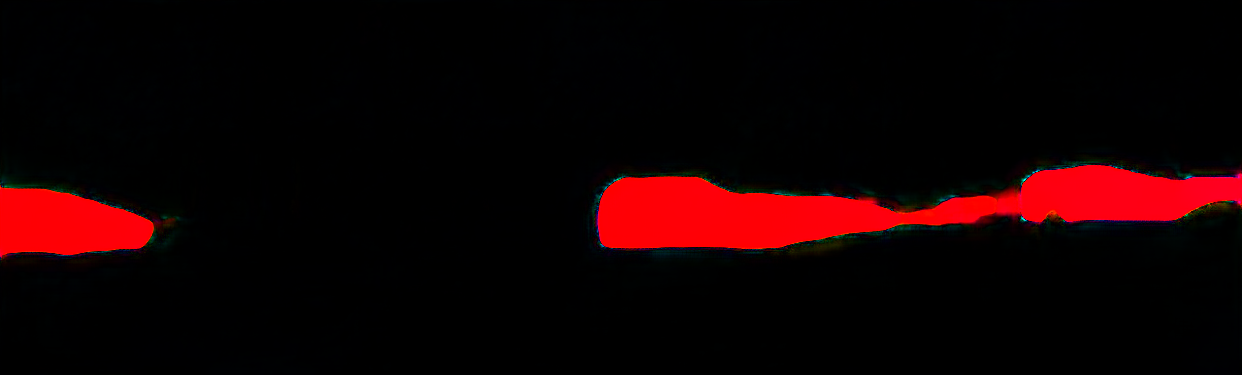} \\
& \includegraphics[width=.97\linewidth]{publications/USegScene/figures/comparison_3/right_rgb.png}
& \begin{overpic}[width=0.97\linewidth,tics=10]{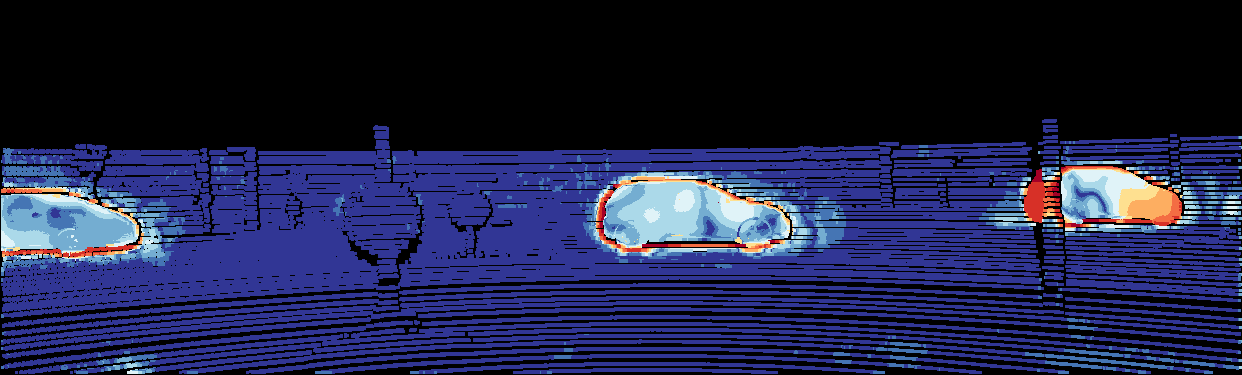}
 \put (60,25) {\textcolor{red}{\tiny $F1_{\mathrm{all}}=5.19$}}
\end{overpic}
& \begin{overpic}[width=0.97\linewidth,tics=10]{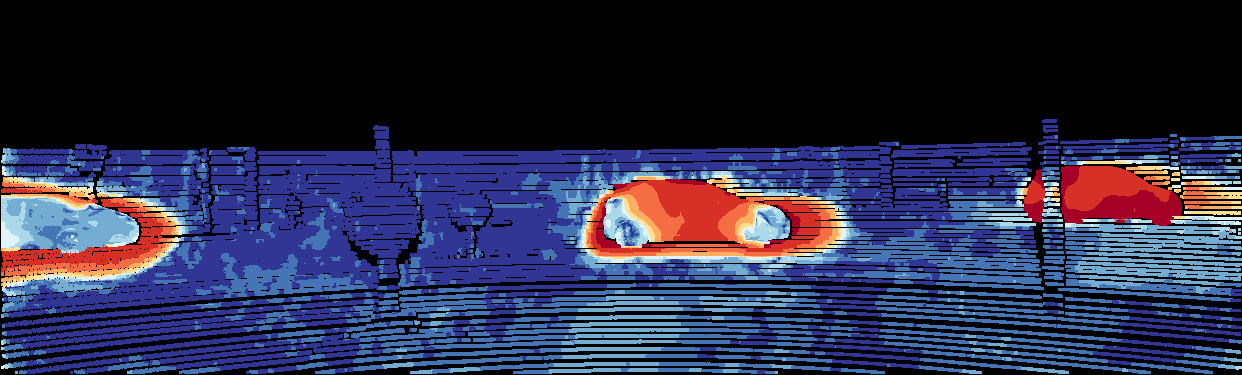}
 \put (60,25) {\textcolor{red}{\tiny $F1_{\mathrm{all}}=20.56$}}
\end{overpic}
& \begin{overpic}[width=0.97\linewidth,tics=10]{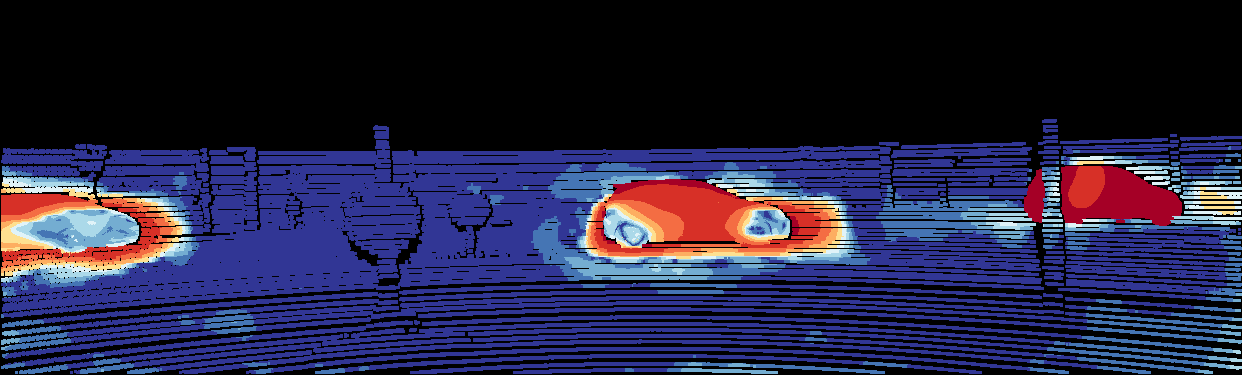}
 \put (60,25) {\textcolor{red}{\tiny $F1_{\mathrm{all}}=22.31$}}
\end{overpic}
& \begin{overpic}[width=0.97\linewidth,tics=10]{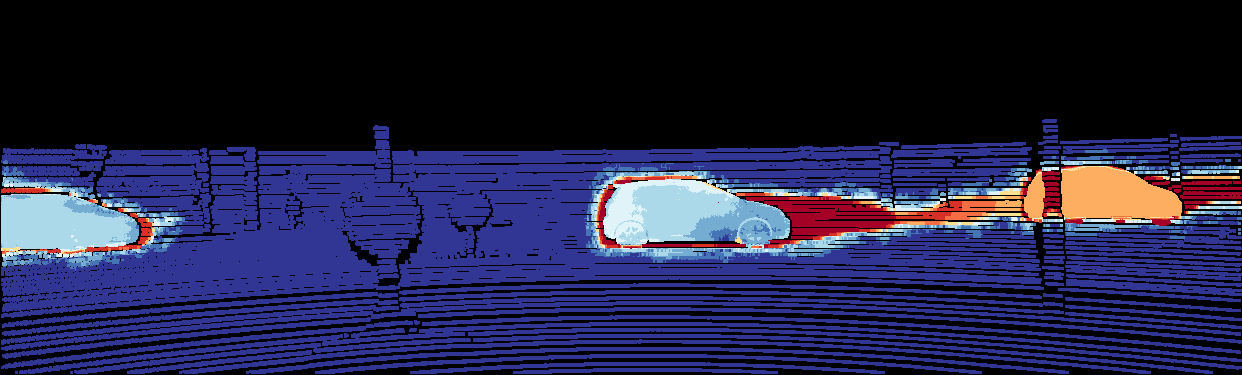}
 \put (60,25) {\textcolor{red}{\tiny $F1_{\mathrm{all}}=10.53$}}
\end{overpic}\\

\noalign{\smallskip}\hline\noalign{\smallskip}

\end{tabular} 
\caption{Qualitative comparison to other state-of-the-art methods for optical flow and disparity prediction. We compare our method to previous unsupervised methods UnOS, DispSegNet and EPC++, and to the supervised methods DispNetC and FlowNet2.}
\label{fig:qualComp}
\vspace{-0.3cm}
\end{figure*}

\subsection{Implementation Details}
We implement our method using the pytorch \cite{paszke2017automatic} framework. We train our network in two stages with two distinct sets of hyperparameters $\lambda_1 ... \lambda_8$. First, we train our network for 50k iterations with $[10,1, 0.001, 0.1, 0, 4, 0, 10]$ and a initial learning rate $5e^{-05}$. In the second stage we increase the weight of the WSPS, the depth occlusion filling and the flow occlusion filling terms, which yield a total parameter set as $[10,1, 0.1, 0.1, 0.5, 4, 2, 10]$. We train the second stage for 150k iterations with a initial learning rate of $1e^{-5}$. During all trainings we set the batch size to 2 and half the learning rate every 25k iterations. Adam \cite{kingma2014adam} is used as the optimizer.

We apply random cropping as data augmentation and set the crop size to $1024x256$. Note that our odometry module needs the full and non-cropped image to work. We thus pass the cropped and the full input in two copies of the network, resulting in two different outputs. We compute all losses on the outputs of the network with cropped input, but take the ego-motion estimate of the network with full input. We fix all parameters except the ones of the odometry estimator modules in the network with full-input. Thus, all training of the feature encoder, depth estimator and optical flow estimator is done on with cropped input, while optimizing the parameters of the ego-motion estimator is done with full image input.

\subsection{Datasets}
We use the KITTI raw dataset \cite{Geiger2013IJRR} to train our network for the disparity, optical flow and visual odometry tasks and adopt the corresponding training and testing sets for evaluation. In order to be comparable to other works \cite{wang2019unos, epc++, zhang2019dispsegnet, liu2019selflow, zou2018df} we train our network on different splits which are described in the following task-dependent sections. The semantic segmentation network is trained on the mapillary vistas dataset \cite{MVD2017} which comprises of 25k labelled images that are taken from various types of cameras and perspectives. While training all other parts of our network, the weights of the semantic segmentation network stay fixed.
Other than for the semantic network we do not apply any form of network pretraining.

\subsection{Disparity Evaluation}

\begin{table}[h]
\footnotesize 
\centering
\caption{Comparison of disparity estimation performance with state-of-the-art approaches on the KITTI 2015 dataset.}
\label{tab:dispEval}
\begin{tabular}{p{0.5cm} p{2.3cm}  p{0.5cm} p{0.5cm} p{0.5cm} V{3} cp{0.5cm} }
\hline\noalign{\smallskip}
 & &  \multicolumn{3}{c}{\textbf{Test split}} &  \multicolumn{1}{c}{\textbf{Train split}} \\
Type & Approach & D1-bg & D1-fg & D1-all &  D1-all\\
\noalign{\smallskip}\hline\hline\noalign{\smallskip}
\parbox[t]{2mm}{\multirow{1}{*}{\rotatebox[origin=c]{90}{S}}} 
& DispNetC \cite{mayer2016large} 
& $4.32$ & $4.41$ & $4.34$ & -  \\
\hline
\parbox[t]{2mm}{\multirow{5}{*}{\rotatebox[origin=c]{90}{U}}} 
& Godard et al. \cite{godard2017unsupervised}
& - & - & - & $9.19$\\
& UnOS \cite{wang2019unos}
& $5.10$ & $14.551$ & $6.67$ & $5.94$\\
& DispSegNet \cite{zhang2019dispsegnet}
& $4.20$ & $16.97$ & $6.33$ & $6.32$\\
& SegStereo \cite{yang2018segstereo}
& - & - & - & $10.03$\\
\cmidrule{2-6}
& USegScene (ours) 
& $\mathbf{4.12}$ & $\mathbf{6.58}$ & $\mathbf{4.53}$ & $\mathbf{4.50}$\\

\noalign{\smallskip}\hline\noalign{\smallskip}
\end{tabular}
\vspace{-0.4cm}
\end{table}

\begin{table*}

\footnotesize 
\centering
\caption{Model ablation study for optical flow and depth on the Kitti 2015 dataset. All models are trained on the multi-view and evaluated on the training split.}
\label{tab:ablationStudy}
\begin{tabular}{m{0.5cm} | p{1.2cm} p{1.2cm} p{1.2cm} p{1.2cm} p{1.2cm} p{1.2cm} | p{0.5cm} p{0.5cm} | p{0.5cm} p{0.5cm} | p{0.5cm} p{0.5cm} | p{0.5cm} p{0.5cm}} 

\hline\noalign{\smallskip}

 & & & & & & & \multicolumn{4}{c}{\textbf{Flow}} & \multicolumn{2}{c}{\textbf{Depth}}\\
Nr. & \centering{Depth} & \centering{Semantic} &\centering{Semantic} & \centering{WSPS} & \centering{3D} & \centering{Rigid}  & \multicolumn{2}{c}{All} & \multicolumn{2}{c}{Noc} & \multicolumn{1}{c}{All}\\

& \centering{Input}& \centering{Input}& \centering{Matching} &  & \centering{Coupling} & \centering{Occlusion Filling}  & F1 & EPE & F1 & EPE & D1\\

\noalign{\smallskip}\hline\hline\noalign{\smallskip}
1 & \centering{\cmark} & \centering{\cmark} & \centering{\xmark} & \centering{\xmark} & \centering{\xmark} & \centering{\xmark} & $85.81$ & $4.10$ & $74.99$ & $10.69$  & $9.66$\\
2 &\centering{\cmark} & \centering{\cmark} & \centering{\cmark} & \centering{\xmark} & \centering{\xmark} & \centering{\xmark} & $86.56$ & $3.85$ & $75.55$ & $10.54$  & $9.50$\\
3 &\centering{\cmark} & \centering{\cmark} & \centering{\cmark} & \centering{\cmark} & \centering{\xmark} & \centering{\xmark} & $88.28$ & $3.26$ & $77.49$ & $9.52$  & $6.71$\\
4 & \centering{\cmark} & \centering{\cmark} & \centering{\cmark} & \centering{\cmark} & \centering{\cmark} & \centering{\xmark} & $88.37$ & $3.26$ & $77.54$ & $9.41$  & $5.84$\\
5 & \centering{\xmark} &\centering{\cmark} & \centering{\cmark} & \centering{\cmark} & \centering{\cmark} & \centering{\cmark} & $88.93$ & $2.29$ & $82.29$ & $3.80$  & $6.00$\\
6 &\centering{\cmark} &\centering{\xmark} & \centering{\cmark} & \centering{\cmark} & \centering{\cmark} & \centering{\cmark} & $89.30$ & $2.34$ & $86.96$ & $3.15$  & $6.30$\\
7 & \centering{\cmark} & \centering{\cmark} & \centering{\cmark} & \centering{\cmark} & \centering{\cmark} & \centering{\cmark} & $\mathbf{90.62}$ & $\mathbf{2.03}$ & $\mathbf{88.56}$ & $\mathbf{2.79}$ & $\mathbf{5.73}$ \\

\noalign{\smallskip}\hline\noalign{\smallskip}

\end{tabular}

\begin{tabular}{m{0.5cm} | m{4cm} m{4cm} } 

\hline\noalign{\smallskip}
Nr. & Qualitative Flow & Qualitative Depth \\

1 & \includegraphics[width=.97\linewidth]{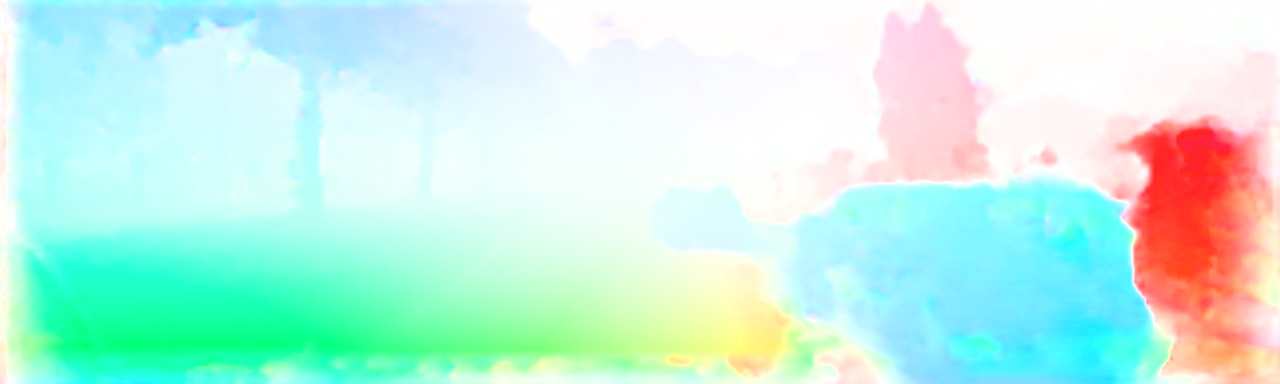} & \includegraphics[width=.97\linewidth]{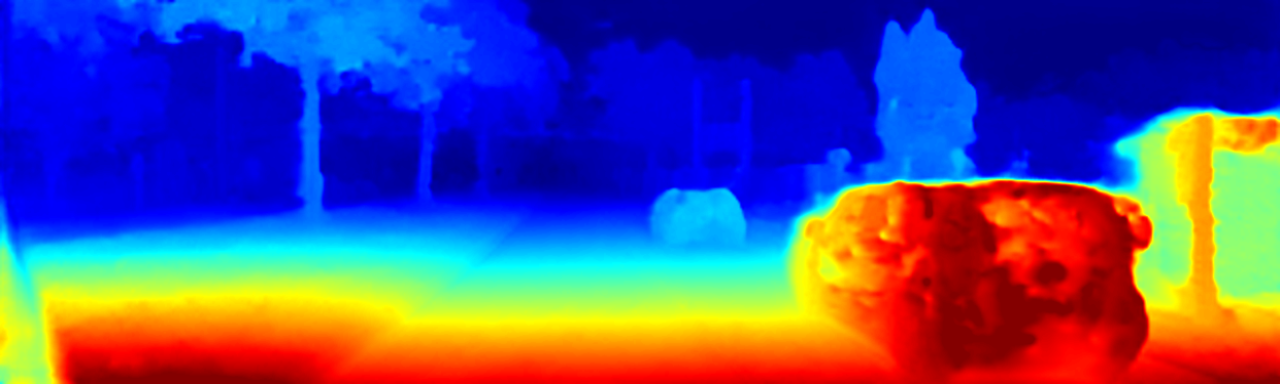} \\

2 & \includegraphics[width=.97\linewidth]{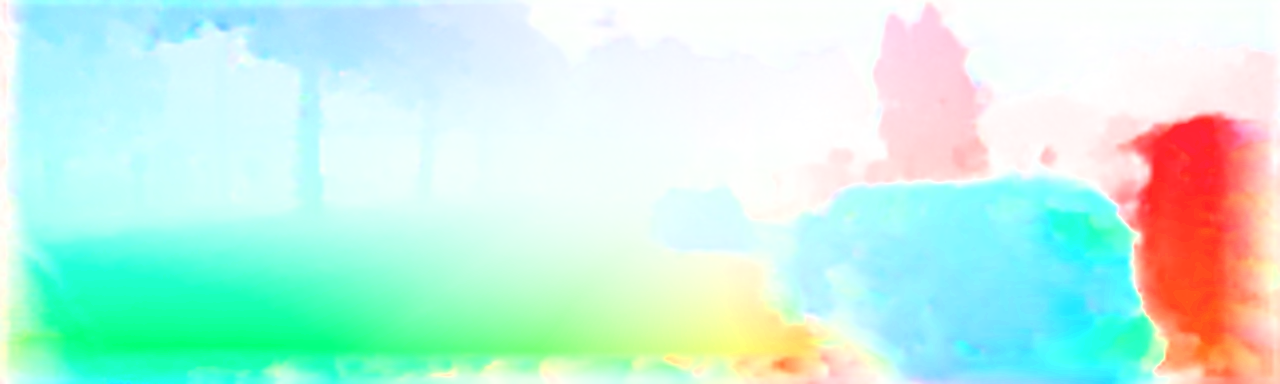} & \includegraphics[width=.97\linewidth]{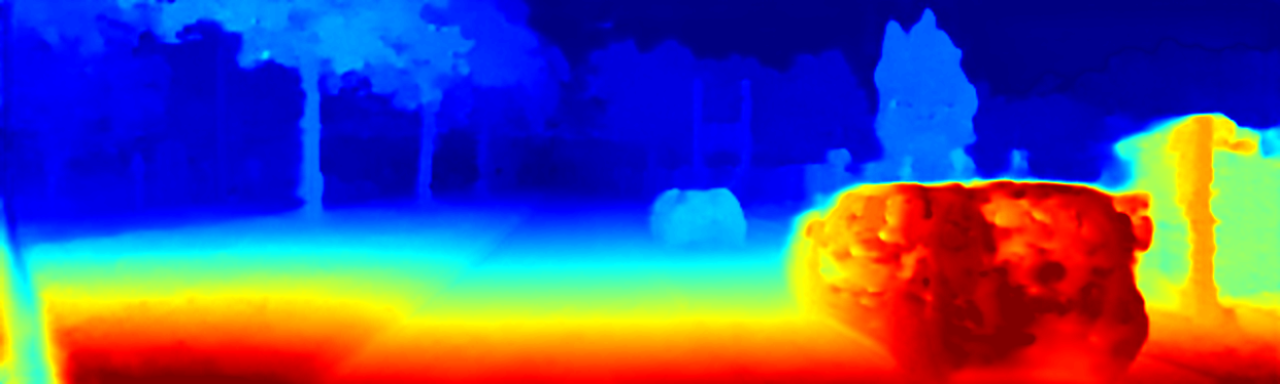} \\

3 & \includegraphics[width=.97\linewidth]{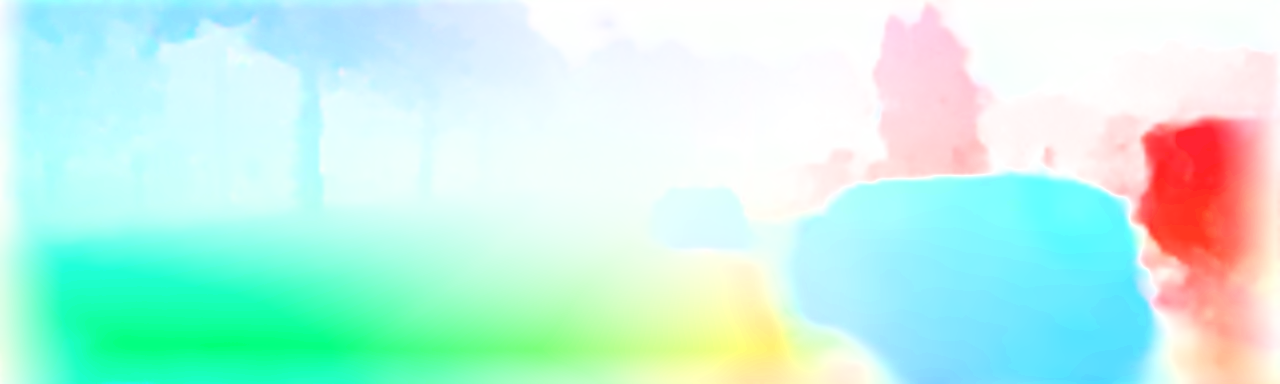} & \includegraphics[width=.97\linewidth]{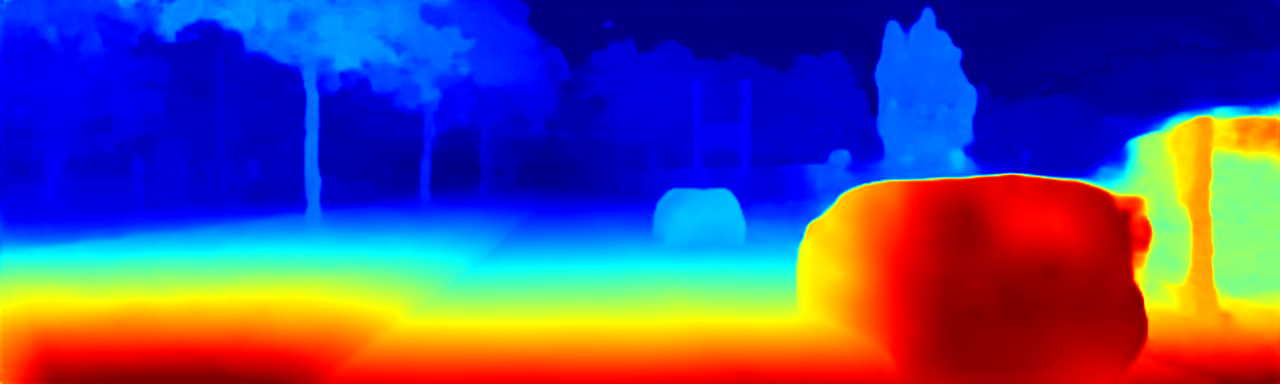} \\

4 & \includegraphics[width=.97\linewidth]{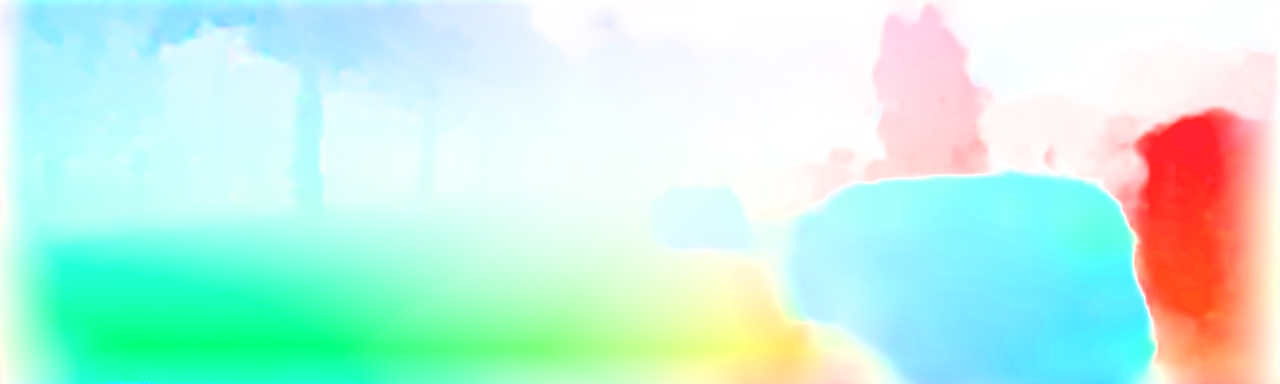} & \includegraphics[width=.97\linewidth]{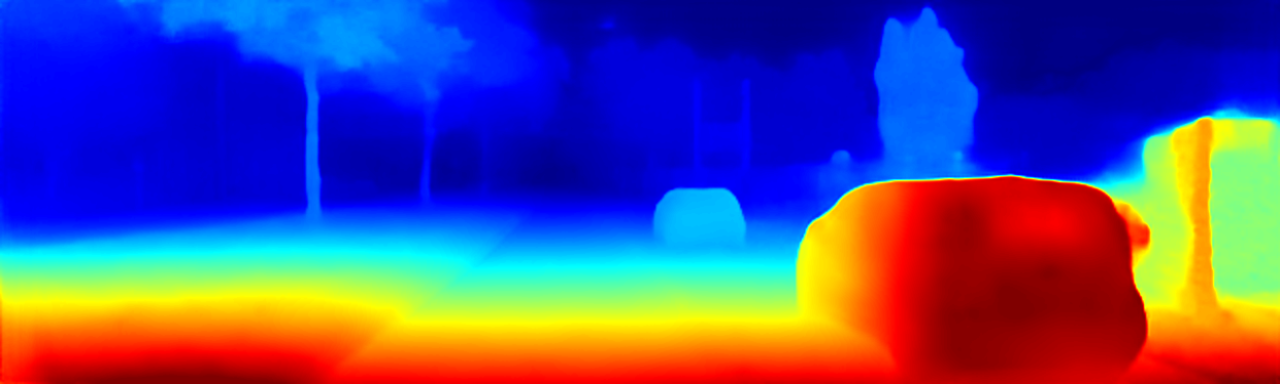} \\

5 & \includegraphics[width=.97\linewidth]{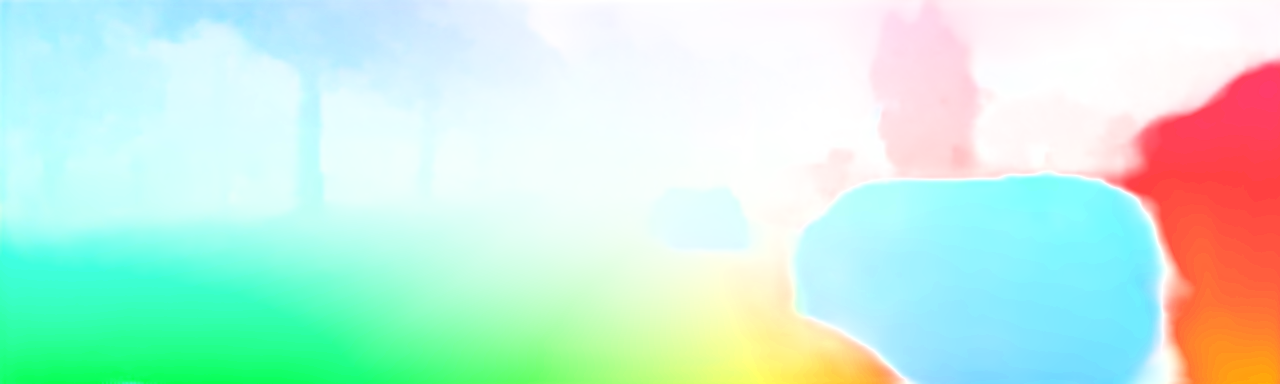} & \includegraphics[width=.97\linewidth]{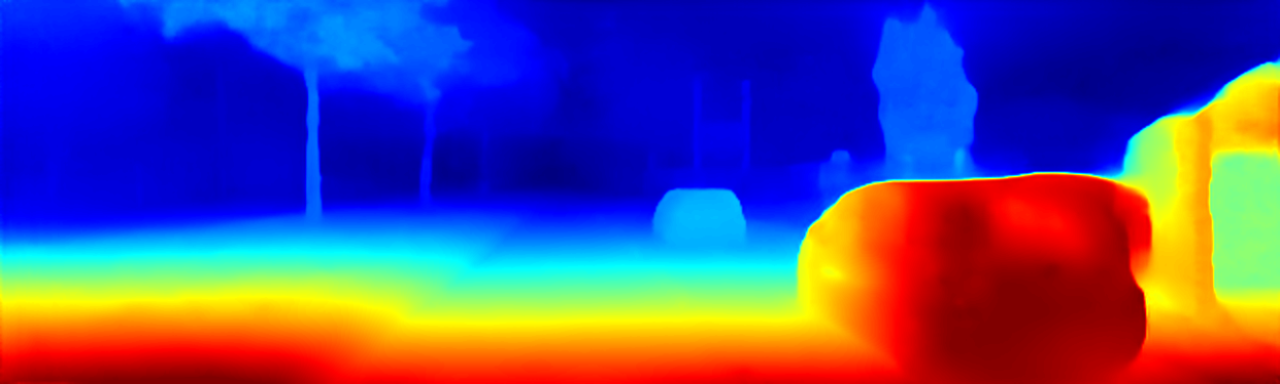} \\

6 & \includegraphics[width=.97\linewidth]{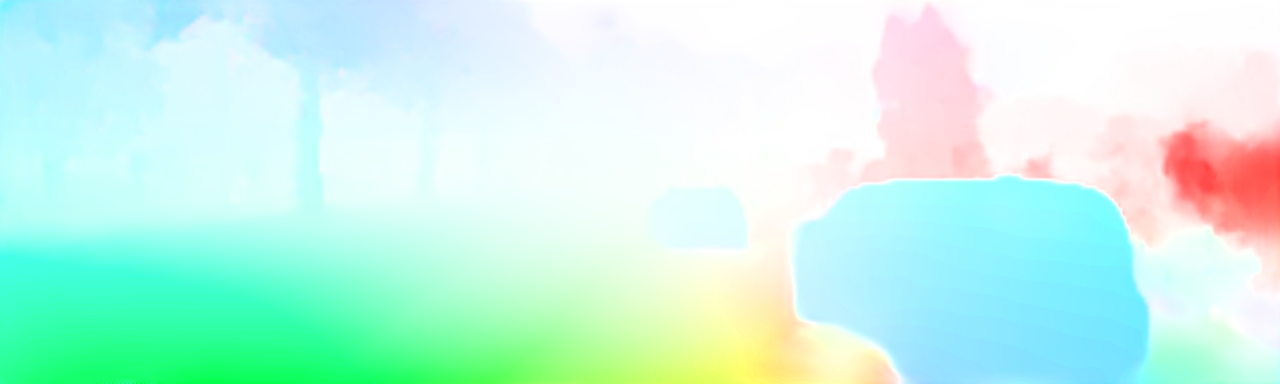} & \includegraphics[width=.97\linewidth]{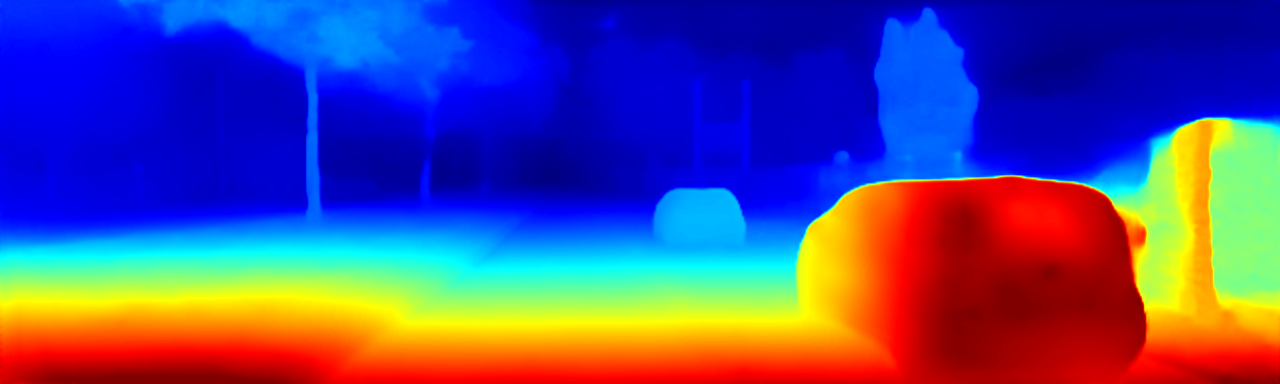} \\

7 & \includegraphics[width=.97\linewidth]{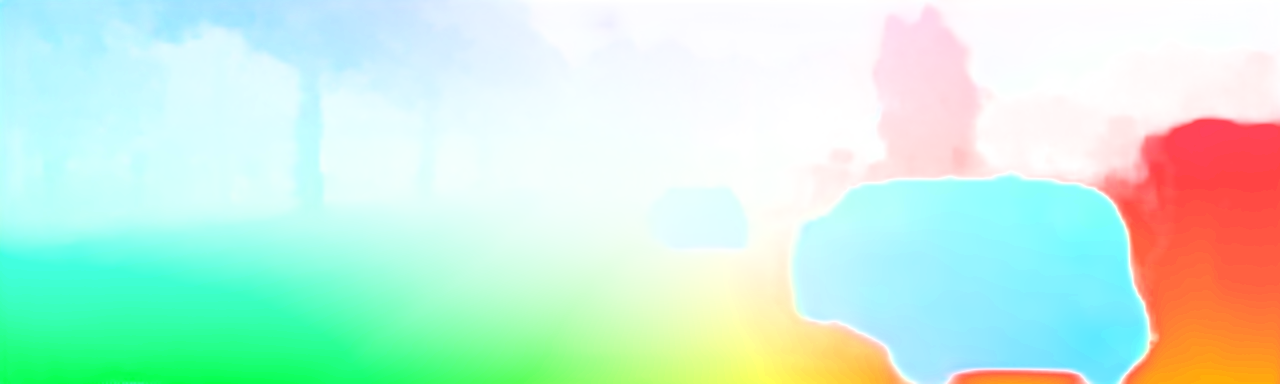} & \includegraphics[width=.97\linewidth]{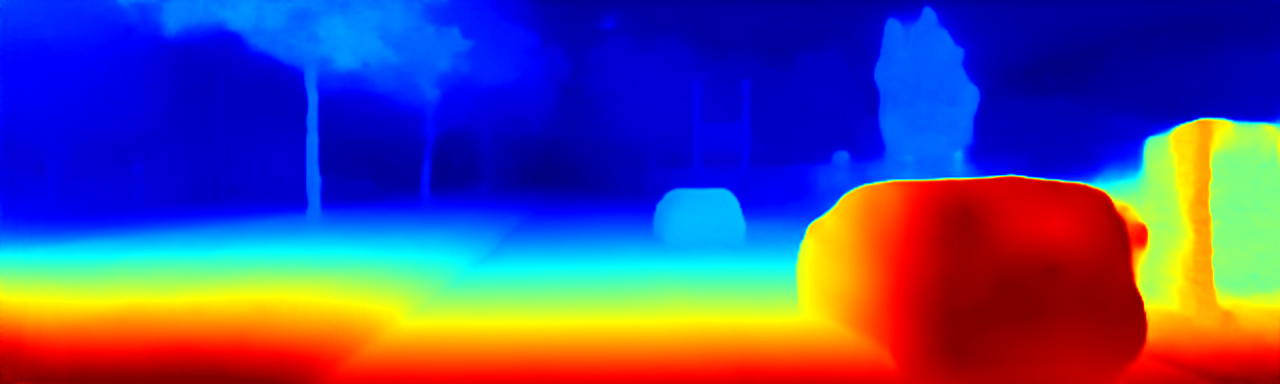} \\

\noalign{\smallskip}\hline\noalign{\smallskip}
\end{tabular}
\vspace{-0.4cm}
\end{table*}

In order to evaluate the disparity estimations we train our network on the KITTI raw dataset and remove all scenes that contain images of the KITTI 2012 or 2015 optical flow/ disparity sets, following \cite{wang2019unos, zhang2019dispsegnet, epc++}. In Table \ref{tab:dispEval} we show the D1 error \cite{Geiger2012CVPR} on the training and testing splits of KITTI and compare it to various state-of-the-art methods. It can be seen that our method outperforms other unsupervised approaches by a large margin. Our approach
 performs on par with the fully supervised method DispNetC \cite{mayer2016large}, which is trained on the ground-truth data from KITTI 2015.

Fig. \ref{fig:qualComp} illustrates multiple comparisons to other unsupervised and unsupervised methods and reveals that our network predicts significantly smoother disparity maps while being sharp at depth boundaries.
Additionally, our disparity predictions are significantly less blurred out in occluded regions. Since we regularize the disparity predictions in regions that belong to sky, we do not get artifacts that occur in previous works. In contrast to other works, our network directly predicts the occlusions. An example of a occlusion prediction can be seen in Fig. \ref{fig:covergirl}.

In Table. \ref{tab:ablationStudy} we further present different configurations of our approach. We observe that the all presented losses consistently improve the results, while the WSPS and 3D coupling losses yield the highest gains of $41.6\%$  and $14.9\%$ respectively.
If the semantic inputs are neglected in the disparity estimator modules the performs drops by $9.9\%$, showing the benefit of complementary semantic inputs.
 
\subsection{Optical Flow Evaluation}
\label{sec::opticalFLowEval}

\begin{table*}[h]
\footnotesize 
\centering
\caption{Comparison of optical flow estimation performance with state-of-the-art approaches on the KITTI 2015 dataset.}
\label{tab:flowComp}
\begin{tabular}{p{4cm} p{1.0cm} p{0.6cm} p{0.6cm} p{0.6cm} V{3} p{0.6cm} | p{0.6cm} |}
\hline\noalign{\smallskip}
 & & \multicolumn{3}{c}{\textbf{Test split}} &  \multicolumn{2}{c}{\textbf{Train split}} \\
 &  & \multicolumn{3}{c}{F1} & \multicolumn{1}{c}{EPE} & \multicolumn{1}{c}{F1} \\
Approach & Split & bg-all & fg-all & all & all & all\\
\noalign{\smallskip}\hline\hline\noalign{\smallskip}
UnFlow-CSS \cite{meister2018unflow} & raw
& - & - & - & $8.10$ & - \\
\noalign{\smallskip}\hline\noalign{\smallskip}
GeoNet \cite{yin2018geonet} & raw
& - & - & -  & $10.81$  & -\\
DF-Net \cite{zou2018df} & raw
& - & - & $25.70$ & $8.98$  & -\\
Competitive-Collaboration-uft \cite{ranjan2019competitive} & raw
& - & - & $25.27$  & $5.66$  & -\\
EPC++ (stereo) \cite{epc++} & raw
& $19.24$ & $26.93$ & $20.52$ & $5.43$  & -\\
UnOS \cite{wang2019unos} & raw
& $16.93$ & $23.34$ & $18.00$ & $5.58$ & -\\

\noalign{\smallskip}\hline\noalign{\smallskip}
Multi-frame \cite{janai2018unsupervised} & mv
& $22.67$ & $24.27$  & $22.94$ & $6.59$ & -\\
DSTFlow \cite{ren2017unsupervised} & mv
& - & - & $39.00$ & $16.79$ & -\\
OccAwareFlow \cite{wang2018occlusion} & mv
& - & -  & $31.20$ & $8.88$ & -\\
DDFlow \cite{liu2019ddflow} & mv
& $13.08$ & $20.40$ & $14.29$ & $5.72$  & -\\
SelFlow \cite{liu2019selflow} & mv
& $12.68$ & $21.74$ & $14.19$  & $4.84$ & - \\

\cmidrule{1-7}
USegScene (ours) & raw
& - & - & - & $4.78$ & $14.47$\\
USegScene (ours) & mv
& $\mathbf{11.69}$ & $\mathbf{20.23}$ & $\mathbf{13.11}$ & $\mathbf{2.79}$  & $\mathbf{11.97}$\\
\noalign{\smallskip}\hline\noalign{\smallskip}
\end{tabular}
\vspace{-0.4cm}
\end{table*}

As in the disparity evaluation we train our network on the raw dataset without the scenes that hold images of the 2015/2012 optical flow training/testing sets that we use for evaluation. This enables us to be comparable to UnOS \cite{wang2019unos}, EPC++ \cite{epc++}, DF-Net\cite{zou2018df}, Competitive-Collaboration \cite{ranjan2019competitive}, GeoNet \cite{yin2018geonet} and UnFlow \cite{meister2018unflow}.

Since most of the dynamic object instances of the KITTI dataset are contained in the corresponding scenes of the optical flow/disparity training/testing sets, many recent works \cite{liu2019selflow, liu2019ddflow, ren2017unsupervised, wang2018occlusion, janai2018unsupervised} leverage the multi-view extensions of those sets for training. Here, all images with ground-truth maps, and their neighbors, are removed for training.

We train our model on both sets and name the sets 'raw' and 'mv' respectively. In table \ref{tab:flowComp}, we compare our optical flow predictions to all other methods using the average end-point error as well as the F1 error \cite{Geiger2013IJRR}.

We found that training on the multi-view dataset yields consistently better results, which may be due to higher occurrences of dynamic objects in the dataset.

Our method outperforms other methods by a significant margin. 

The qualitative results, shown in Fig. \ref{fig:qualComp} reveal that our method shows especially great performance in occluded regions and regions with high displacement vectors.

As in the disparity evaluation section, we also present an ablation study for the optical flow estimation in Fig. \ref{tab:ablationStudy}. We observe that semantic feature matching improves the EPE in the non-occluded benchmark by $6.5\%$, WSPS by another $18.1\%$ and rigid occlusion filling by further $60.5\%$. In contrast to the disparity case we see only little improvements when utilizing the 3D coupling loss. When looking at the EPE errors in occluded and non-occluded regions one can note that our rigid occlusion filling loss improves the EPE from $9.41$ to $2.79$.

In order to better understand the influence of our novel network architecture we additionally conduct experiments where we neglect the semantic inputs $S$ and $B$ to all modules. We can note that removing the semantic inputs yields to $1.8\%$ performance decrease. 
Furthermore, we note a $7.6\%$ decrease in the performance of the optical flow prediction when neglecting the depth and depth-dependent inputs $D_{l_t}$ and $F_{l_t}^R$ at the input of the optical flow estimator modules. This confirms that geometric inputs ease the optical flow estimation tasks.

\begin{figure}
\scriptsize 
\centering 
\setlength{\tabcolsep}{0.3em}
\renewcommand{\arraystretch}{1}
\begin{tabular}{P{4cm} P{4cm}}
\includegraphics[width=.97\linewidth]{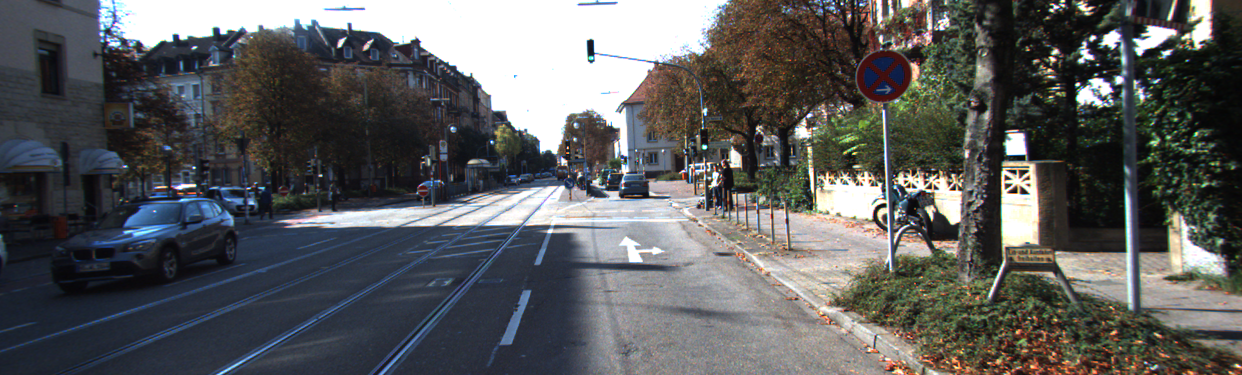} &
\includegraphics[width=.97\linewidth]{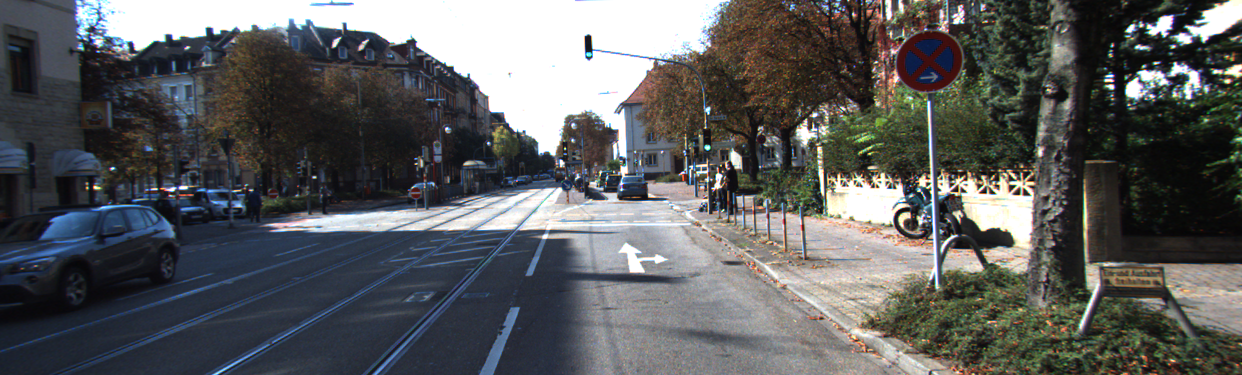} \\
\includegraphics[width=.97\linewidth]{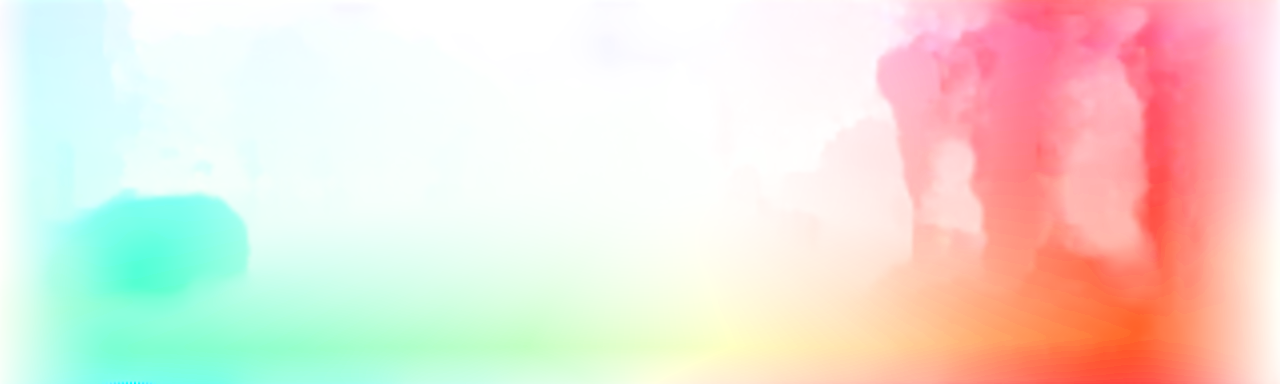} &
\includegraphics[width=.97\linewidth]{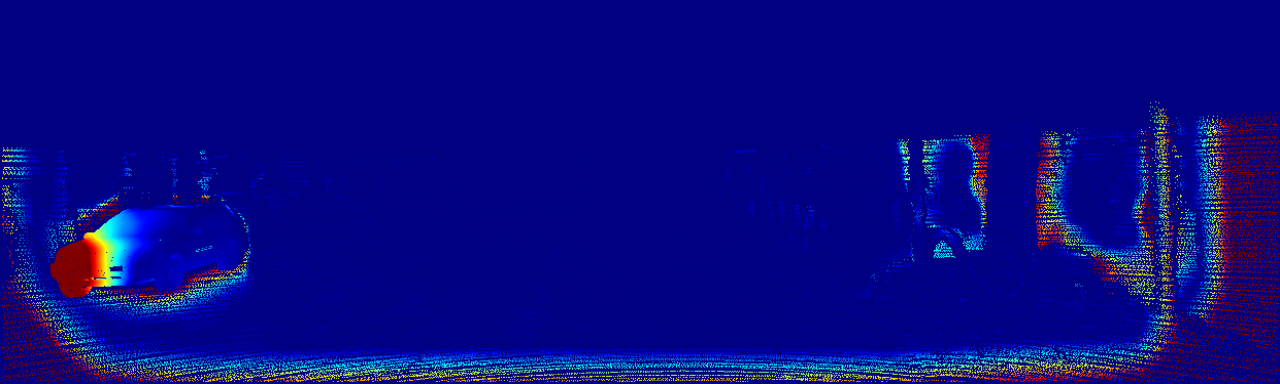} \\
\includegraphics[width=.97\linewidth]{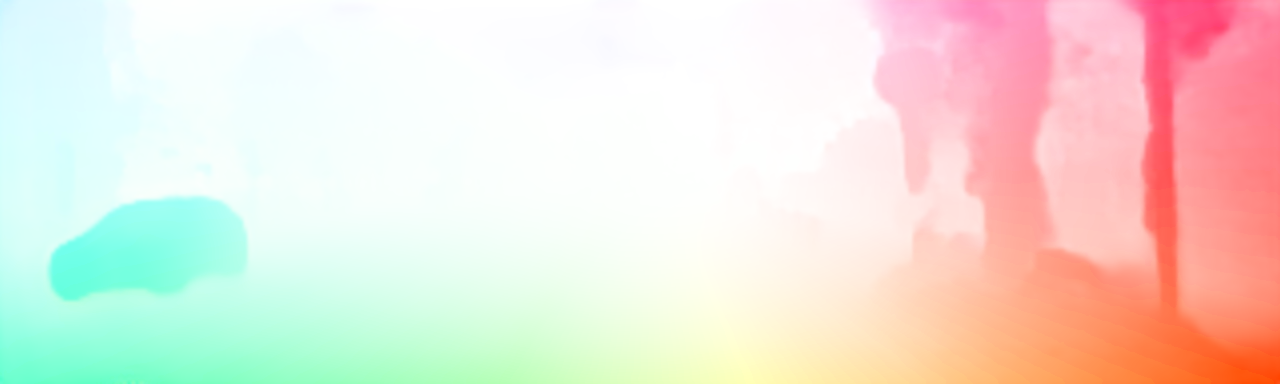} &
\includegraphics[width=.97\linewidth]{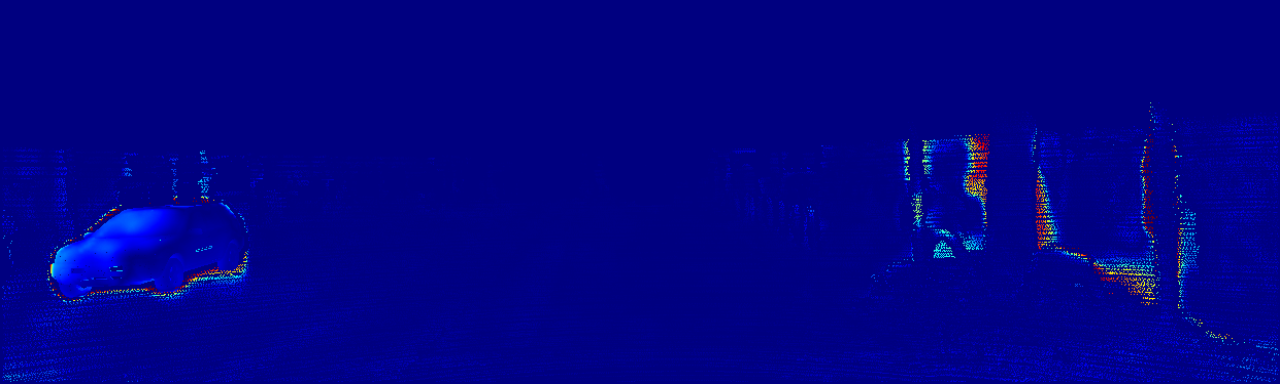}
\end{tabular} 
\caption{
This figure shows the utility of the optical flow rigid occlusion filling loss. It can be observed that the model, which is trained with rigid occlusion filling predicts much more accurate optical flow in occluded static as well as dynamic regions.
Top: consecutive rgb images. Middle: optical flow prediction and corresponding error-map without rigid flow occlusion filling. Bottom: optical flow prediction and corresponding error-map with rigid flow occlusion filling.}
\label{fig:occComparison}
\vspace{-0.3cm}
\end{figure}

In Fig. \ref{fig:occComparison} we show a example optical flow prediction with and without rigid optical flow occlusion filling. Here, rigid optical flow occlusion filling not only drastically reduces the error in occluded static regions but also in occluded rigid dynamic regions, as we guide the prediction by individual SE(3) transformations of the dynamic rigid object.

\subsection{Ego-Motion Estimation}

We follow \cite{epc++, wang2019unos} and use the KITTI odometry sequences 00-08 as training and sequences 09, 10 as validation datasets. Table \ref{tab:odometryEvaluation} shows a comparison to other unsupervised  state-of-the-art methods using the metrics described in \cite{Geiger2012CVPR}. Note that we only use 2-frames to predict the ego-motion, while others \cite{sfmlearner, yin2018geonet, mahjourian2018unsupervised} require more. The numbers suggest that our method outperforms other approaches by a large margin, showing the effectiveness of our method.

\begin{table}
\footnotesize 
\centering
\caption{Odometry evaluation on the KITTI odometry dataset. We compare our approach to other state-of-the-art methods on sequence 9 and 10.}
\label{tab:odometryEvaluation}
\begin{tabular}{p{2cm} | P{1.0cm}P{1.0cm} | P{1.0cm} P{1.0cm}}
\hline\noalign{\smallskip}
Approach & \multicolumn{2}{c}{\textbf{Seq. 09}} &  \multicolumn{2}{c}{\textbf{Seq. 10}}\\
 & $t_{err}\% $ & $ r_{err}(\frac{\degree}{100m})$  & $t_{err}\% $ & $r_{err}(\frac{\degree}{100m})$\\
\noalign{\smallskip}\hline\hline\noalign{\smallskip}
Zhou et al. \cite{sfmlearner} & $30.75$ & $11.41$ & $44.22$ & $12.42$\\
GeoNet \cite{yin2018geonet} & $39.43$ & $14.30$ & $28.99$ & $8.85$\\
EPC++ \cite{epc++} & $8.84$ & $3.34$ & $8.86$ & $3.18$\\
UnOS \cite{wang2019unos} & $5.21$ & $1.80$ & $5.20$ & $2.18$\\

\noalign{\smallskip}\hline\noalign{\smallskip}
Uscene (ours) & $\mathbf{2.71}$ & $\mathbf{0.52}$ & $\mathbf{3.43}$ & $\mathbf{0.72}$ \\

\noalign{\smallskip}\hline\noalign{\smallskip}
\end{tabular}
\vspace{-0.5cm}
\end{table}

\section{Conclusion}
In this paper we propose a novel unsupervised framework for learning depth, optical flow and ego-motion. We propose a new architecture that solves all individual tasks jointly, using shared network components and leveraging semantic maps as complementary information. We present novel loss functions for complementary semantic feature matching, semantic regularization, 3D-constraints and occlusion handling. To quantify our work, we present state-of-the-art quantitative and qualitative results on the KITTI benchmark \cite{Geiger2012CVPR}. 

\bibliographystyle{unsrt}
\bibliography{publications/USegScene/sections/references}






\end{document}